\documentclass[runningheads]{llncs}

\usepackage[mobile]{eccv}

\usepackage{eccvabbrv}

\usepackage{graphicx}
\usepackage{booktabs}

\usepackage{amssymb}
\usepackage{enumitem}
\usepackage{mathtools}
\usepackage{wrapfig}
\usepackage{xcolor}
\usepackage{subcaption}
\usepackage{algorithm}
\usepackage{algpseudocode}

\usepackage[accsupp]{axessibility}  

\usepackage[hidelinks]{hyperref}

\usepackage{orcidlink}

\begin{document}

\title{Vector Scaffolding: Inter-Scale Orchestration for Differentiable Image Vectorization}

\titlerunning{Vector Scaffolding}

\author{%
Jaerin Lee\inst{1}\orcidlink{0009-0005-5271-8270} \and
Kanggeon Lee\inst{1}\orcidlink{0000-0002-2258-1059} \and
Kyoung Mu Lee\inst{1,2}\orcidlink{0000-0001-7210-1036}
}

\authorrunning{J.~Lee et al.} 
\institute{%
Dept. of ECE\&ASRI, Seoul National University, Korea
\and IPAI, Seoul National University, Korea\\
\email{\{ironjr,dlrkdrjs97,kyoungmu\}@snu.ac.kr}
}

\maketitle

\begin{abstract}
  Differentiable vector graphics have enabled powerful gradient-based optimization of vector primitives directly from raster images.
However, existing frameworks formulate this as a flat optimization problem, forcing hundreds to thousands of randomly initialized curves to blindly compete for pixel-level error reduction.
This disordered optimization leads to \emph{topology collapse}, where macroscopic structures are distorted by internal high-frequency noise, resulting in a redundant and uneditable ``polygon soup'' that limits practical editability.
To address this limitation, we propose \textbf{Vector Scaffolding}, a novel hierarchical optimization framework that shifts from flat pixel-matching to structured topological construction tailored for vector graphics.
By identifying a key cause of topology collapse as the mathematical imbalance between area and boundary gradients, we introduce \emph{Interior Gradient Aggregation} to stabilize the learning dynamics of multi-scale curve mixtures.
Upon this stabilized landscape, we employ \emph{Progressive Stratification} and \emph{Rapid Inflation Scheduling} to progressively densify vector primitives with extremely high learning rates ($\times 50$).
Experiments demonstrate that our approach accelerates optimization by $2.5\times$ while simultaneously improving PSNR by up to 1.4\,dB over the previous state of the art.

\end{abstract}

\section{Introduction}
\label{sec:1_intro}
\begin{figure}[t]
\centering
\begin{subfigure}[t]{0.70\linewidth}
\centering
\includegraphics[width=\linewidth]{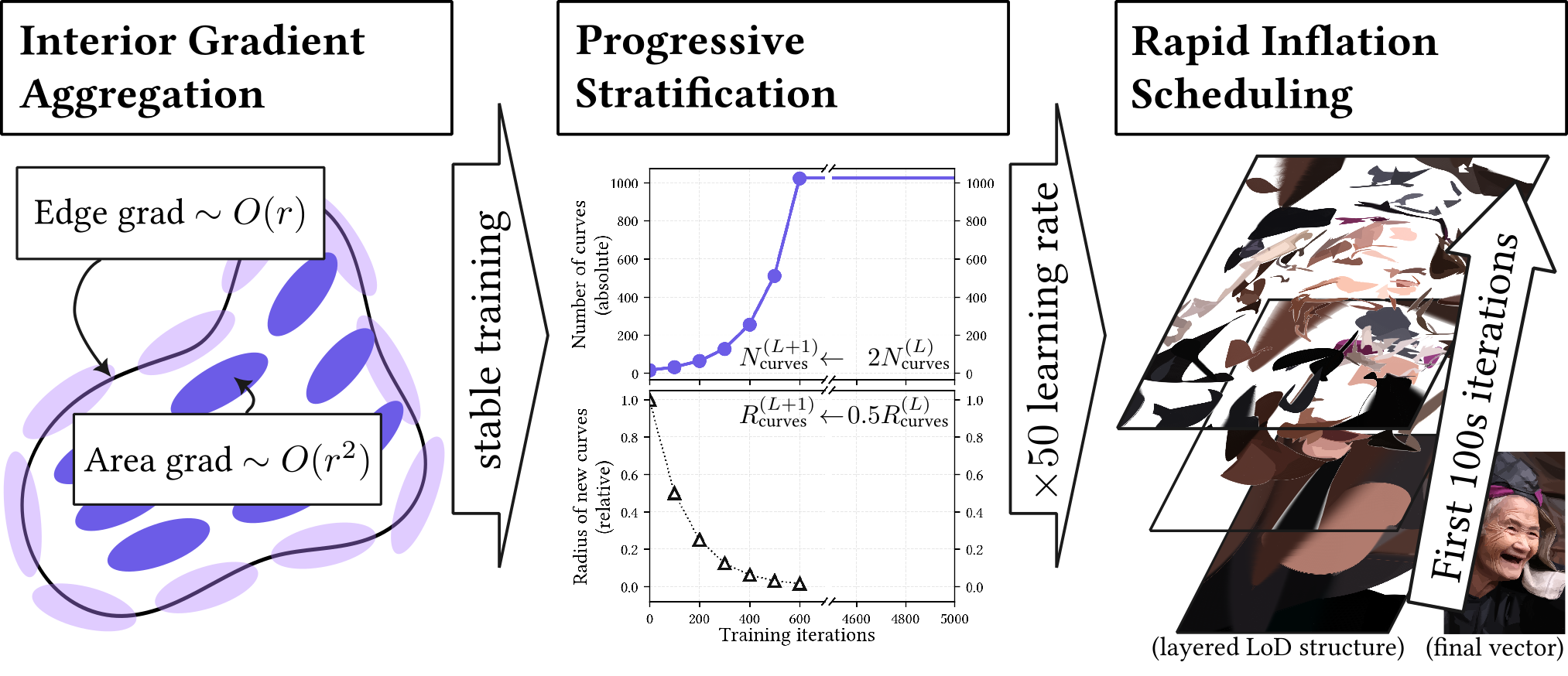}
\caption{High-level concept of Vector Scaffolding.}
\label{fig:figure_one_concept}
\end{subfigure}
\hfill
\begin{subfigure}[t]{0.29\linewidth}
\centering
\includegraphics[width=\linewidth]{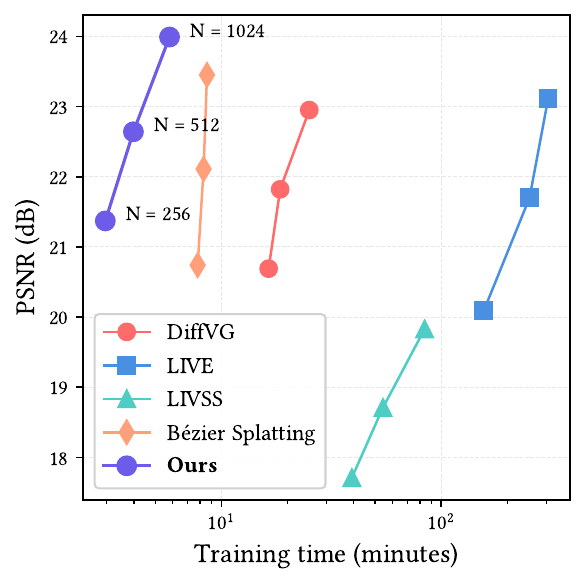}
\caption{DIV2K benchmark}
\label{fig:figure_one_benchmark}
\end{subfigure}
\caption{
\textbf{Overview.}
We introduce \textbf{Vector Scaffolding}, a hierarchical optimization framework for fast and stable differentiable image vectorization.
(a) Our framework consists of three ideas:
\textit{Interior Gradient Aggregation} stabilizes training vectors with large scale variance.
\textit{Progressive Stratification} further stabilizes training to enable huge learning rate.
This allows \textit{Rapid Inflation Scheduling} to accelerate training and outperform baselines by a large margin.
(b) We achieve higher rendering fidelity in a fraction of wall clock time required by existing methods.
}
\label{fig:figure_one}
\end{figure}

Raster images capture \textit{what we see}; vector graphics encode \textit{how we understand it}.
Because they are resolution-independent and inherently editable through professional creative software relying on path-based tools, vector graphics have become indispensable for digital art, animation, and modern design workflows.
Recent advancements in differentiable vector graphics~\cite{li2020diffvg,ma2022live,wang2025livss,liu2025bezier} bridge the gap between these two distinct domains.
Their central idea is to allow gradients to flow through the rasterization process, enabling gradient-based optimization algorithms to directly synthesize or reconstruct vectors from pixel supervision.
While pioneering works like DiffVG~\cite{li2020diffvg}, and later LIVE~\cite{ma2022live}, established the mathematical foundation for this task, their prohibitive computational costs make them impractical for real-world applications.

The slow speed of these early works is due to the sequential reconstruction of vectors, curve by curve.
B\'ezier Splatting~\cite{liu2025bezier} introduced a parallelized optimization scheme for vectorization.
A highly optimized differentiable rasterization pipeline powered by 2D Gaussian splatting~\cite{kerbl2023gaussian,huang2024twodgs} along B\'ezier curves significantly accelerates the optimization process.
However, it still inherits the \emph{flat optimization problem} from the early works, where all curves are treated uniformly.
Initializing hundreds of curves simultaneously on a single plane and letting them compete for per-pixel local loss reduction leads to disordered optimization dynamics.
At best, this leads to a superpixel-like representation with low editability and representational fidelity.
At worst, it results in an unstructured, uneditable ``polygon soup,'' blocking creators from further editing and polishing the artwork.
We name this phenomenon \emph{topology collapse}.
This undermines the key advantage of vector representation: editability.

Natural \emph{and} artistic images, on the other hand, have an inherent hierarchy of scales and details.
Later works like LIVSS~\cite{wang2025livss} and Optimize \& Reduce~\cite{hirschorn2024optimize} were the first to leverage this hierarchical structure.
Nevertheless, these methods depend heavily on external image priors such as diffusion models~\cite{ho2020ddpm} to build the image hierarchy.
The computational cost introduced by these priors fundamentally limits the scalability of the frameworks.

In our \textbf{Vector Scaffolding}, we instead propose a purely on-the-fly optimization-based framework for hierarchical vectorization, without relying on learned external priors.
Following the wisdom of how artists create artwork from simple sketches and gradually add details, we construct vector graphics by progressively spawning smaller curves above base constructions.
We observe that the commonly used splatting formulation~\cite{liu2025bezier} emphasizes boundary samples and omits positional gradients from interior samples.
Our work starts with \textbf{Interior Gradient Aggregation} to add the missing gradients.
This allows large base curves to be effectively pulled from their interior, greatly stabilizing the optimization process.
We then establish a structural hierarchy through \textbf{Progressive Stratification}, inspired by the natural power law of image frequency.
Once lower-frequency base layers stabilize, we stack a new layer with a larger number of smaller curves in places where the residual error concentrates.
Empirically, our spawning rule can support multiple levels of detail without suffering from curve redundancy.
The \textbf{Progressive Stratification} rule also prevents upper-layer high-frequency errors from distorting the lower layers, further stabilizing the overall optimization trajectory.
This allows us to increase the learning rate by a factor of 50 compared to the previous state of the art~\cite{liu2025bezier}, without causing instability.
Finally, based on this fast optimization trajectory, we redesign the scheduling to further accelerate convergence.
Our \textbf{Rapid Inflation Scheduling} strategy allocates new curves in the first few hundred iterations safely, while the rest of the iterations finalize the optimization process.

By restructuring the optimization dynamics, our \emph{Vector Scaffolding} accelerates the entire optimization process by $2.5\times$ in wall-clock time compared to the state-of-the-art B\'ezier Splatting~\cite{liu2025bezier}, with improved reconstruction fidelity by up to 1.4\,dB in PSNR.
The results provide strong empirical evidence that optimization structure is crucial in achieving good performance in vectorization.

In summary, our main contributions are as follows:

\begin{itemize}
\item We propose \textbf{Vector Scaffolding}, a novel hierarchical optimization framework for differentiable image vectorization that shifts from flat pixel-matching to structured construction for fast and better vectorization.
\item We introduce \textbf{Interior Gradient Aggregation} to match the optimization target to the Reynolds transport theorem, greatly stabilizing the optimization process.
\item We formulate \textbf{Progressive Stratification} with \textbf{Rapid Inflation Scheduling}, establishing a structured hierarchy that allows aggressive learning rates without causing instability.
\item Experiments demonstrate that our method achieves a $2.5\times$ speedup in optimization measured in wall-clock time and up to 1.4\,dB PSNR improvement at identical curve budgets compared to the state-of-the-art B\'ezier Splatting~\cite{liu2025bezier}.
\end{itemize}

\section{Related Work}
\label{sec:2_survey}

\subsection{Differentiable Vector Graphics and Optimization}

Image vectorization has been significantly advanced by differentiable rendering techniques.
DiffVG~\cite{li2020diffvg} introduced the first differentiable vector graphics (VG) rasterizer, enabling gradient-based optimization of vector primitives through the rasterization process.
Building upon this foundation, subsequent works such as LIVE~\cite{ma2022live}, Optimize \& Reduce (O\&R)~\cite{hirschorn2024optimize}, and layered vectorization approaches~\cite{wang2025livss} explored layer-wise path initialization and hierarchical optimization strategies to better preserve image topology during vectorization.
Other works focused on improving color representation within vector graphics, for example by modeling regional color distributions using linear gradients~\cite{du2023lineargradient} or implicit neural representations~\cite{chen2024highfidelity}.
Despite their effectiveness, these approaches often incur substantial computational overhead due to the complexity of differentiable rasterization~\cite{li2020diffvg,ma2022live} or introduction of external priors~\cite{wang2025livss}.
As a result, optimizing vector representations for high-resolution images can remain computationally expensive.

Another line of work learns neural models to directly synthesize vector graphics from raster inputs.
Early approaches such as Lopes et al.~\cite{lopes2019learned} adopt encoder–decoder architectures for vector generation.
Subsequent works including Im2Vec~\cite{reddy2021im2vec} and SVGFormer~\cite{cao2023svgformer} further improve structured vector synthesis by introducing sequential modeling and transformer-based architectures.
More recently, diffusion-based generative models have been explored for vector graphics synthesis and text-to-VG generation~\cite{ho2020ddpm, zhang2024texttovector, jain2023vectorfusion, xing2024svgdreamer, xing2023diffsketcher}.
While these learning-based approaches show promising results for stylized graphics such as icons or sketches, they often face challenges when applied to complex images in professional domains.

Recently, Bézier Splatting~\cite{liu2025bezier} was proposed to accelerate differentiable vectorization by combining Bézier curves with the efficient Gaussian splatting rendering pipeline~\cite{kerbl2023gaussian,huang2024twodgs}.
Although this substantially improves rendering efficiency, the optimization process remains largely flat, with all curves optimized simultaneously under a single pixel-level objective.
Such dynamics may lead to unstable geometric structures, as curves can adapt to high-frequency textures instead of capturing semantically meaningful boundaries, or compete with each other for the same pixel-level residuals.
Our work addresses this limitation by introducing a hierarchical optimization strategy~\cite{hirschorn2024optimize,wang2025livss} that separates coarse structural geometry from fine-grained details, but without external priors.
Unlike prior layered methods that sequentially add paths~\cite{ma2022live} or rely on external diffusion priors~\cite{wang2025livss}, we directly stabilize parallel optimization of closed Bézier curves without any external model.
This simple framework yields an order-of-magnitude speedup over previous hierarchy-based methods~\cite{wang2025livss} with $+1\,\mathrm{dB}$ PSNR improvements as shown in Figure~\ref{fig:figure_one_benchmark} and Table~\ref{tab:div2k_quantitative}.

\subsection{Gaussian Splatting and Representation Learning}

3D Gaussian Splatting (3DGS)~\cite{kerbl2023gaussian,zwicker2001ewa} has recently emerged as an efficient explicit representation for neural rendering and novel view synthesis.
Its differentiable tile-based rasterization pipeline enables high-fidelity rendering with real-time performance.
The efficiency and flexibility of this representation have inspired numerous extensions, including dynamic scene modeling~\cite{luiten2024dynamic3dgaussians,wu2024fourdfs}, generative 3D content creation~\cite{tang2024dreamgaussian,yi2024gaussiandreamer}, and geometric reconstruction using alternative primitives such as 2D Gaussians~\cite{guedon2024sugar,huang2024twodgs} or tetrahedral meshes~\cite{guo2025tetsphere}.

Beyond 3D scene representation, Gaussian-based primitives have also been explored for 2D image modeling.
Works such as GaussianImage~\cite{zhang2024gaussianimage} and Image-GS~\cite{zhang2025imagegs} demonstrate that collections of 2D Gaussians can serve as compact and expressive alternatives to raster images or implicit neural representations~\cite{mueller2022instant,sitzmann2020siren}.

However, directly transferring splatting-based optimization to vector graphics introduces additional challenges due to the geometric constraints of vector primitives such as closed Bézier curves.
Naively densifying primitives to minimize pixel-wise reconstruction errors can produce redundant curve sets and reduce the editability of the resulting vector graphics.
In contrast, our method adopts a residual-guided adaptive densification strategy together with hierarchical scale control, enabling compact vector representations while maintaining structural coherence.

\begin{figure}[t]
\centering
\includegraphics[width=\linewidth]{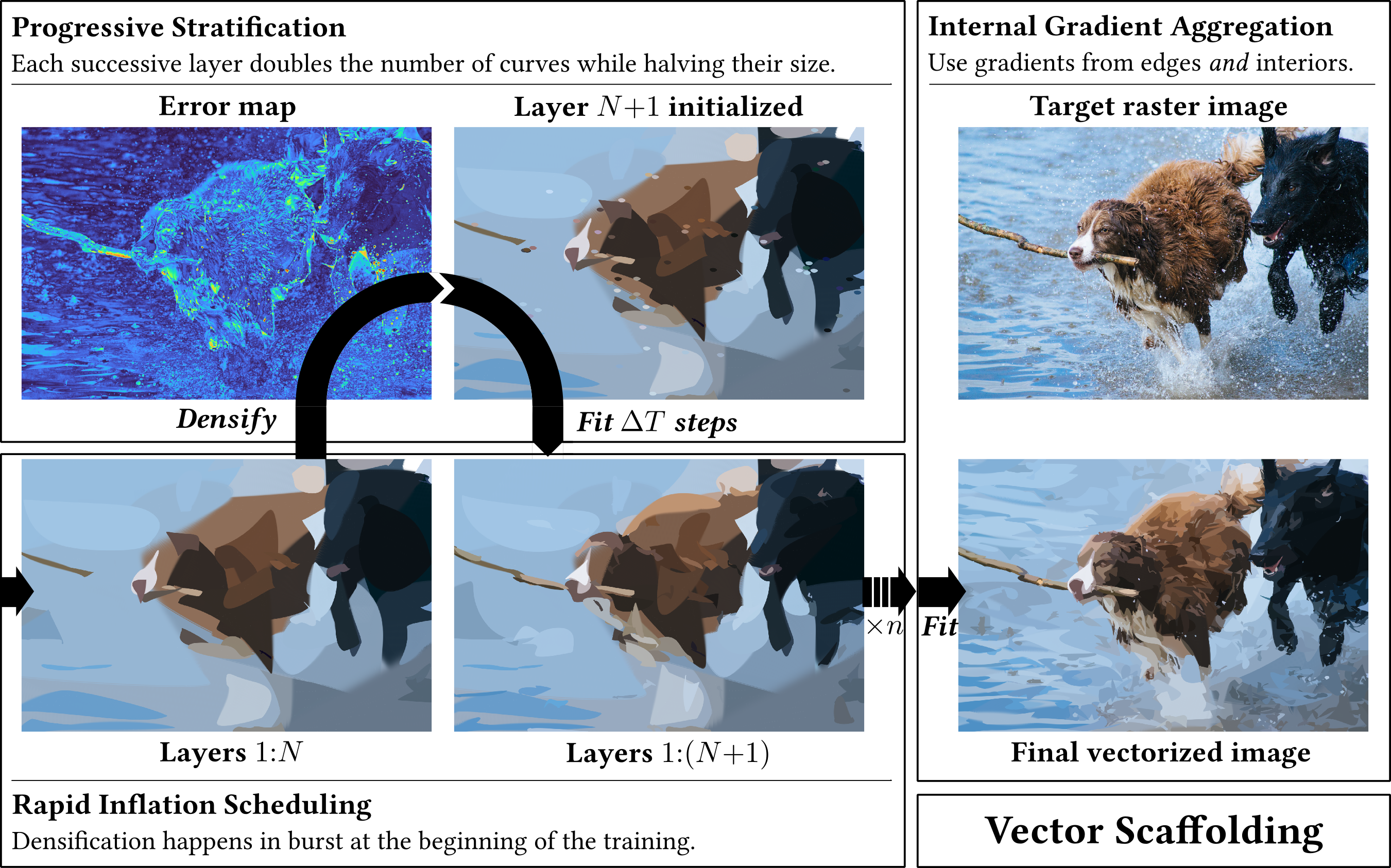}
\caption{\textbf{Algorithm visualized.} (tr) \textit{Interior Gradient Aggregation}: Optimization is stabilized by aggregating internal area gradients alongside boundary gradients via the Reynolds transport theorem.
(tl) \emph{Progressive Stratification} aligns vector representation with the natural power law of image frequency, enabling extremely high learning rates without instability.
(bl) \textit{Rapid Inflation Scheduling}: 
The vector representation \emph{inflates} in the first few hundred iterations.
(br) \textit{Vector Scaffolding Algorithm}: These components work together to achieve the final vector representation.}
\label{fig:pipeline}
\end{figure}

\section{Method}
\label{sec:3_method}
\subsection{Preliminaries}

The goal of image vectorization is to construct a vector-based representation $\mathcal{B} = \{\mathcal{B}_i\}_{i=1}^N \in V_N$ of $N$ vector primitives from a raster image $\mathcal{I} \in \mathbb{R}^{C \times H \times W}$.
We define a differentiable renderer $g: \mathcal{B} \mapsto \hat{\mathcal{I}}$ that produces the corresponding raster image $\hat{\mathcal{I}}$ from the vector representation $\mathcal{B}$.
The problem of image vectorization can then be formulated as a minimization problem:
\begin{equation} \label{eq:reconstruction_problem}
    \mathcal{B}^* \;=\; \arg\min_{\mathcal{B} \in V_{N}} \mathcal{L}_{2}(g(\mathcal{B}),\, \mathcal{I}) + \mathcal{L}_{\text{reg}}(\mathcal{B}),
\end{equation}
where $\mathcal{L}_{2}$ is the pixel-wise MSE loss in the raster domain and $\mathcal{L}_{\text{reg}}$ is a collection of shape regularization terms that prevent vector primitives from forming unnatural pathological shapes.
The regularization strategy is described in Section~\ref{subsec:regularization}.

We employ closed B\'ezier curves as our internal representation and adopt the splatting-based differentiable rasterizer backend of B\'ezier Splatting~\cite{liu2025bezier}.
Formally, a B\'ezier curve $\mathcal{B}(t)$ of degree $M$ is defined by a Bernstein polynomial:
\begin{equation}
    \mathcal{B}(t) \;=\; \sum_{j=0}^M P_j B_j^M(t), \quad t \in [0,1],
\end{equation}
where $B_j^M(t)$ is the $j$-th Bernstein polynomial of degree $M$ and $P_j$ is the $j$-th control point in the 2D image space.
We treat each curve as a closed loop by linearly interpolating between the first and last control points, and assign boundary samples the same color and opacity as the filled region.
That is, we additionally have a curve color $\mathbf{c}_i$ and an opacity $\alpha_i$ parameter for each curve $i \in \{1, 2, \dots, N\}$.
To rasterize these curves, we perform $\alpha$-blending of 2D Gaussians~\cite{kerbl2023gaussian} sampled along the boundary and interior of the curves, following the approach of B\'ezier Splatting~\cite{liu2025bezier}.
Each Gaussian is parameterized with an anisotropic covariance matrix $\Sigma=RSS^T R^T$, where the rotation matrix $R$ is aligned with the local tangent to the curve, and the scale matrix $S = \text{diag}(\sigma_x, \sigma_y)$ is analytically derived from the distances between adjacent sampled points.
The $\alpha$-blending requires an additional layer depth parameter $z_i$ for each curve for consistent occlusion.
That is, at each pixel $v$, its color $C[v]$ is computed as:
\begin{equation} \label{eq:alpha_blending}
    C[v] \;=\; \sum_{i \in V[v]} \mathbf{c}_i \alpha_i[v] \prod_{j \in V[v], z_j < z_i} (1 - \alpha_j[v]),
\end{equation}
where $V[v]$ is the set of curves that cover the pixel $v$, and $\alpha_i[v]$ denotes the rasterized opacity contribution of curve $i$ at pixel $v$.
This defines the differentiable rasterizer $g(\mathcal{B})$.

Unlike 3DGS~\cite{kerbl2023gaussian}, determining the depth parameter $z_i$ in this 2D scenario is not straightforward.
For example, B\'ezier Splatting~\cite{liu2025bezier} dynamically reassigns the depth parameter $z_i$ during optimization by sorting curves by their bounding box sizes.
With unstructured, random initialization of the primitives and maximally fair competition for per-pixel loss reduction, this leads to unstable, jittering optimization dynamics.
The curves become heavily redundant and unnaturally shrink, overlap, or twist to escape local minima, resulting in a highly uneditable, superpixel-like representation similar to \emph{polygon soup} in 3D graphics.
We refer to this failure mode as \emph{topology collapse}, where curves lose coherent structural roles and degenerate into redundant, locally optimized fragments.
Our Vector Scaffolding addresses this challenge by introducing a dedicated optimization methodology for 2D image vectorization, inspired by the artist's workflow and the structure of natural images.
Our method consists of three synergistic components: \emph{Interior Gradient Aggregation} to stabilize individual base curves, \emph{Progressive Stratification} to construct a structured hierarchy, and \emph{Rapid Inflation Scheduling} that rapidly establishes the hierarchical structure upon stabilized learning dynamics.

\subsection{Interior Gradient Aggregation}
\label{subsec:aggregation}

In our scenario, the Gaussians serve as a surrogate representation that carries local interaction information for each curve.
This differs from the 3D case~\cite{kerbl2023gaussian}, where the Gaussians are independent.
The actual gradients driving our optimization problem are a sum of the gradients of the Gaussians sampled from a single curve, not the individual Gaussians.
Therefore, the relative weights of gradient contributions within a curve significantly affect the optimization trajectory.
There are two classes of Gaussians: those sampled along the boundary and those sampled along the interior of the curve.
A key cause of \emph{topology collapse} stems from the inherent scale disparity between the interior area and the boundary of a closed geometric primitive.
In 2D space, the gradient magnitude derived from a curve's interior area scales quadratically $O(r^2)$, while the gradient from its boundary scales linearly $O(r)$, where $r$ is the radius of the curve.
Given a curve $\mathcal{B}_i$, let $\mathcal{A}_i$ denote the 2D interior region enclosed by it, and $\partial\mathcal{A}_i$ denote its boundary.
During the backward pass, the gradient of the photometric reconstruction loss $\mathcal{L}_2 = \int_\Omega \|g(\mathcal{B})(x) - \mathcal{I}(x)\|_2^2 dx$ with respect to the control points $P_j$ can be conceptually decomposed using the Reynolds transport theorem~\cite{li2020diffvg}:
\begin{equation}
\label{eq:gradient_decomposition}
\nabla_{P_j} \mathcal{L}_2 \; \approx \; \underbrace{\int_{\mathcal{A}_i} \nabla_{P_j} e(x) dx}_{\text{Area Gradient} \sim O(r^2)} + \underbrace{\int_{\partial\mathcal{A}_i} e(x) \left( \frac{\partial \mathbf{x}}{\partial P_j} \cdot \mathbf{n} \right) ds}_{\text{Boundary Gradient} \sim O(r)},
\end{equation}
where $e(x) = \|g(\mathcal{B})(x) - \mathcal{I}(x)\|_2^2$ is the per-pixel error, and $\mathbf{n}$ is the outward normal vector.
In practice, we approximate the area term as a consistent discrete quadrature over interior Gaussian samples.

Consequently, for large base curves representing macroscopic structures, the area gradient induced by fine-grained internal textures can dominate the optimization trajectory.
This makes the interior gradients crucial for stable optimization.
Nevertheless, previous parallel optimization~\cite{liu2025bezier} ignored these interior gradients, leading to insufficient supervision where curves struggle to obtain information from their enclosed regions.
We address this by aggregating the positional gradients from the interior of each curve with its boundary gradients.
Importantly, \emph{Interior Gradient Aggregation} does not by itself decide which frequency content a curve should explain; rather, it restores the missing positional support from the interior of a closed region.
The separation of coarse structures from fine textures is imposed by \emph{Progressive Stratification} in the next stage, which limits where and at what scale new curves should be introduced.

\subsection{Progressive Stratification}
\label{subsec:scale_decay}

\emph{Interior Gradient Aggregation} effectively aggregates the gradients from the interior of the curve, allowing curves to focus on their own scope and scale.
This enables us to employ the following \emph{Progressive Stratification} to establish a structured hierarchy.
Instead of unstructured random initialization and densification, we establish a structured hierarchy in an ordered manner, starting from the coarsest representation and gradually adding more details.
This is analogous to how human artists build vector artwork from simple sketches and gradually add details.

To encourage separation between coarse and fine structures, new curves $\mathcal{B}^{(k)} \setminus \mathcal{B}^{(k-1)}$ are spawned exclusively in regions with high residual errors, computed as $E(x) = \|g(\mathcal{B}^{(k-1)})(x) - \mathcal{I}(x)\|_2^2$.
More importantly, instead of balancing between pruning and densification operations, we monotonically increase the number of curves geometrically, starting from the coarsest representation, \eg, $N_0 = 16$ and $N_k = r_{\mathrm{num}}^k N_0$, where $r_{\mathrm{num}}$ is the ratio of the number of curves per generation.
Let $R_k$ denote the initialization radius of curves spawned at generation $k$.
We impose a monotonic scale-decaying constraint on newly spawned curves: $R_{k} = r_{\mathrm{rad}} R_{k-1}$.
We reinterpret the depth parameter $z_i$ used in ordering $\alpha$-blending as the temporal ordering of curves, ensuring that newer curves are always on top of older ones.
Newly spawned curves are strictly assigned a lower depth value $z_{new} < z_{base}$, placing them physically on top of older generations.
This explicit z-ordering removes the chaotic depth-sorting instability present in previous works~\cite{liu2025bezier}.

We find that setting $r_{\mathrm{num}} = 2.0$ and $r_{\mathrm{rad}} = 0.5$ yields the best results.
This inherently aligns with the power law of natural images, ensuring that newer generations are less likely to overwrite the low-frequency areas already resolved by their predecessors.
This spatial constraint prevents upper-layer high-frequency errors from distorting the established scaffold.
Our Vector Scaffolding stabilizes the optimization landscape: the depth $z_i$ is structurally fixed, the scale $R_k$ is implicitly band-limited, and the interior gradients are included.
Thanks to this structural stability, we can safely boost the learning rates by $\times 50$.
That is, we increase the LR of color and opacity from $0.01$ to $0.5$ and the LR of control points from $2 \times 10^{-4}$ to $1 \times 10^{-2}$.
This extremely high LR enforces colors to match the image almost immediately and control points to quickly converge to their optimal positions.

\subsection{Vector Scaffolding via Rapid Inflation}
\label{subsec:inflation}

The optimization starts with an extremely sparse set of curves, \eg, $N_0 = 16$, to exclusively capture the global illumination and macroscopic background colors.
Once the lowest-frequency base layer is anchored, we aggressively double the number of curves every $\Delta T$ iterations until the target budget $N_{max}$ is reached.
We find that our \emph{Progressive Stratification} strategy allows us to use a very high learning rate, which makes this \emph{inflation phase} extremely fast.
Throughout the paper, we fix $\Delta T = 100$.
This means we double the number of curves every 100 iterations of training, reaching $N = 1024$ curves after only $600$ iterations.
For the rest of the training process, we do \emph{not} reduce the learning rate but maintain the high learning rate of $0.5$ (color \& opacity) and $0.01$ (control points) until the end of training, where we observe monotonic improvement in reconstruction fidelity.
We achieve increased reconstruction fidelity with up to $+1.4\,\mathrm{dB}$ PSNR improvement without optimization collapse with a small number of iterations (5k), ultimately accelerating the overall vectorization process by $2.5\times$ compared to the best baseline~\cite{liu2025bezier} as shown in Figure~\ref{fig:figure_one_benchmark}.

\subsection{Regularization}
\label{subsec:regularization}

To ensure the final output $\mathcal{B}^*$ is an editable vector graphic rather than fuzzy Gaussians, we apply topology regularization with opacity-based pruning.
Since practical vector-art workflows favor solid, opaque elements, we introduce an opacity regularization loss pushing $\alpha_i$ towards $1.0$:
\begin{equation}
\label{eq:opacity_reg}
\mathcal{L}_{\alpha} \;=\; \frac{1}{N} \sum_{i=1}^N |1 - \sigma(\alpha_i)|,
\end{equation}
where $\sigma(\cdot)$ is the sigmoid activation.
We also adopt the Xing loss~\cite{ma2022live} $\mathcal{L}_{\text{Xing}}$ and B\'ezier curve regularization~\cite{liu2025bezier} $\mathcal{L}_{\text{BS}}$ to avoid self-intersection and maintain structural integrity:
\begin{equation}
\label{eq:regularization}
\mathcal{L}_{\text{reg}} \;=\; 0.01 \mathcal{L}_{\alpha} + 0.02 \mathcal{L}_{\text{Xing}}(\mathcal{B}) + \mathcal{L}_{\text{BS}}(\mathcal{B}).
\end{equation}
The total objective is given in Equation~\eqref{eq:reconstruction_problem}.

Coupled with this regularization, we periodically prune any curve $\mathcal{B}_i$ where $\sigma(\alpha_i) < 0.01$.
Curves that become occluded or drift from target structures naturally lose gradient contribution and opacity, enabling automated culling without complex heuristics.

\section{Experiments}
\label{sec:4_exp}
\subsection{Experimental Setup}
\label{subsec:exp_setup}

\subsubsection{Datasets and Metrics.} 
We evaluate our method on two distinct datasets. 
First, we use the Kodak~\cite{kodak1999lossless} dataset to compare general performance on natural images.
Second, to test reconstruction quality on high-frequency textures, we utilize the high-resolution DIV2K~\cite{agustsson2017ntire} training dataset.
We use standard image quality metrics, PSNR, SSIM, and LPIPS~\cite{zhang2018lpips}, to evaluate reconstruction and perceptual fidelity, and use qualitative visualizations to assess structural coherence.
We also track the wall clock time for the entire optimization process to evaluate the computational efficiency of our framework.
To demonstrate structural efficiency, we strictly enforce identical curve budgets ($N \in \{256, 512, 1024\}$) across all comparisons.

\subsubsection{Baselines.}
We benchmark our method against existing differentiable vectorization methods: DiffVG~\cite{li2020diffvg}, LIVE~\cite{ma2022live}, LIVSS~\cite{wang2025livss}, and B\'ezier Splatting~\cite{liu2025bezier}. 
Since we target practically useful vector graphics, we only consider \emph{closed curves} for evaluation, where areas are filled by the interior of the curves rather than strokes with arbitrary thickness.

\begin{table*}[t]
\centering
\caption{\textbf{Quantitative Evaluation on the Kodak dataset~\cite{kodak1999lossless}}. We outperform the evaluated closed-curve baselines with available Kodak results across all curve budgets.}
\resizebox{\textwidth}{!}{%
\begin{tabular}{l|ccc|ccc|ccc}
\toprule
& \multicolumn{3}{c|}{256 Curves} & \multicolumn{3}{c|}{512 Curves} & \multicolumn{3}{c}{1024 Curves} \\
& PSNR $\uparrow$ & SSIM $\uparrow$ & LPIPS $\downarrow$ & PSNR $\uparrow$ & SSIM $\uparrow$ & LPIPS $\downarrow$ & PSNR $\uparrow$ & SSIM $\uparrow$ & LPIPS $\downarrow$ \\
\midrule
DiffVG~\cite{li2020diffvg} & 24.11 & 0.622 & 0.513 & 25.34 & 0.666 & 0.475 & 26.66 & 0.719 & 0.420 \\
B\'ezier Splatting~\cite{liu2025bezier} & 24.19 & 0.621 & 0.519 & 25.61 & 0.664 & 0.485 & 26.91 & 0.708 & 0.448 \\
\midrule
\textbf{Ours} & \textbf{25.18} & \textbf{0.631} & \textbf{0.427} & \textbf{26.68} & \textbf{0.687} & \textbf{0.353} & \textbf{28.30} & \textbf{0.749} & \textbf{0.275} \\
\bottomrule
\end{tabular}
}
\label{tab:kodak_quantitative}
\end{table*}

\subsubsection{Implementation Details.} 
We utilize the B\'ezier Splatting~\cite{liu2025bezier} rasterizer as the differentiable rasterization backend.
For all experiments, we also aggregate the interior gradient from the interior of the curve.
Leveraging the structurally stabilized loss landscape provided by our exponential scale decaying ($r_{\mathrm{num}}=2.0, r_{\mathrm{rad}}=0.5$), we employ an aggressive spatial learning rate of $0.01$ for the geometric parameters, $50\times$ higher than the standard configuration of B\'ezier Splatting~\cite{liu2025bezier}.
The curve population inflates monotonically, doubling every $\Delta T = 100$ iterations until reaching the target budget $N$. 
That is, starting from $N_0 = 16$ curves, we double the number of curves every 100 iterations of training immediately after initialization.
This means we densify $4$ times for $N =256$ curves, $5$ times for $N = 512$ curves, and $6$ times for $N = 1024$ curves, reaching $N = 1024$ curves at the $600$-th iteration.
For the optimizer, we use the Adan optimizer~\cite{xie2024adan} with $\beta_1 = 0.98$, $\beta_2 = 0.92$, and $\beta_3 = 0.99$ by default.
Learning rates for the scaling and rotation (Cholesky) parameters are set to $\lambda = 0.5$, which is $50\times$ higher than B\'ezier Splatting's learning rate of $0.01$.
Learning rates for the position (control points), color, and opacity parameters are set to $\lambda_{\mathrm{position}} = 0.01$, $\lambda_{\mathrm{color}} = 0.5$, and $\lambda_{\mathrm{opacity}} = 0.5$, respectively.
For our Vector Scaffolding framework, we train each model for 5k iterations.
All reported timings are measured on a single A100 GPU.

\subsection{Main Results}

\subsubsection{Kodak Experiment.}
Table~\ref{tab:kodak_quantitative} summarizes the performance of our method on the Kodak dataset~\cite{kodak1999lossless}.
On Kodak, our hierarchical optimization method outperforms all reproduced baselines under matched closed-curve budgets.
We improve PSNR by 1.0--1.4$\,\mathrm{dB}$ over B\'ezier Splatting across curve budgets, with substantial gains in perceptual quality (LPIPS).
Notably, the performance gap between our method and the state-of-the-art B\'ezier Splatting~\cite{liu2025bezier} exceeds that between DiffVG~\cite{li2020diffvg} and B\'ezier Splatting~\cite{liu2025bezier}.
Since our method and B\'ezier Splatting share the same differentiable rasterization backend, the observed gains isolate the effect of optimization restructuring rather than renderer replacement.
This strongly suggests that our hierarchical optimization framework is as significant as the rasterization backend, which has been the sole focus of recent literature.

\begin{table*}[t]
\centering
\caption{\textbf{Quantitative Evaluation on the DIV2K Dataset.} We compare our Vector Scaffolding with state-of-the-art differentiable vectorization methods (closed curves only). Our method consistently achieves the best PSNR across all sparsity levels, with a margin of roughly 0.5--0.6$\,\mathrm{dB}$ over B\'ezier Splatting, while achieving competitive or better perceptual quality depending on the curve budget.}
\resizebox{\textwidth}{!}{%
\begin{tabular}{l|ccc|ccc|ccc}
\toprule
& \multicolumn{3}{c|}{256 Curves} & \multicolumn{3}{c|}{512 Curves} & \multicolumn{3}{c}{1024 Curves} \\
& PSNR $\uparrow$ & SSIM $\uparrow$ & LPIPS $\downarrow$ & PSNR $\uparrow$ & SSIM $\uparrow$ & LPIPS $\downarrow$ & PSNR $\uparrow$ & SSIM $\uparrow$ & LPIPS $\downarrow$ \\
\midrule
DiffVG~\cite{li2020diffvg} & 20.69 & 0.578 & 0.548 & 21.82 & 0.601 & 0.531 & 22.95 & 0.631 & 0.509 \\
LIVE~\cite{ma2022live} & 20.09 & 0.576 & 0.543 & 21.70 & 0.611 & 0.521 & 23.11 & 0.648 & 0.495 \\
LIVSS~\cite{wang2025livss} & 17.71 & \textbf{0.586} & \textbf{0.542} & 18.71 & \textbf{0.630} & 0.530 & 19.83 & \textbf{0.678} & 0.517 \\
B\'ezier Splatting~\cite{liu2025bezier} & 20.74 & 0.580 & 0.546 & 22.11 & 0.607 & 0.528 & 23.45 & 0.639 & 0.507 \\
\midrule
\textbf{Ours} & \textbf{21.37} & 0.564 & 0.548 & \textbf{22.64} & 0.598 & \textbf{0.503} & \textbf{23.99} & 0.640 & \textbf{0.450} \\
\bottomrule
\end{tabular}%
}
\label{tab:div2k_quantitative}
\end{table*}

\begin{figure}[h!]
\centering
\begin{subfigure}[b]{0.325\linewidth}
\centering
\includegraphics[width=\linewidth]{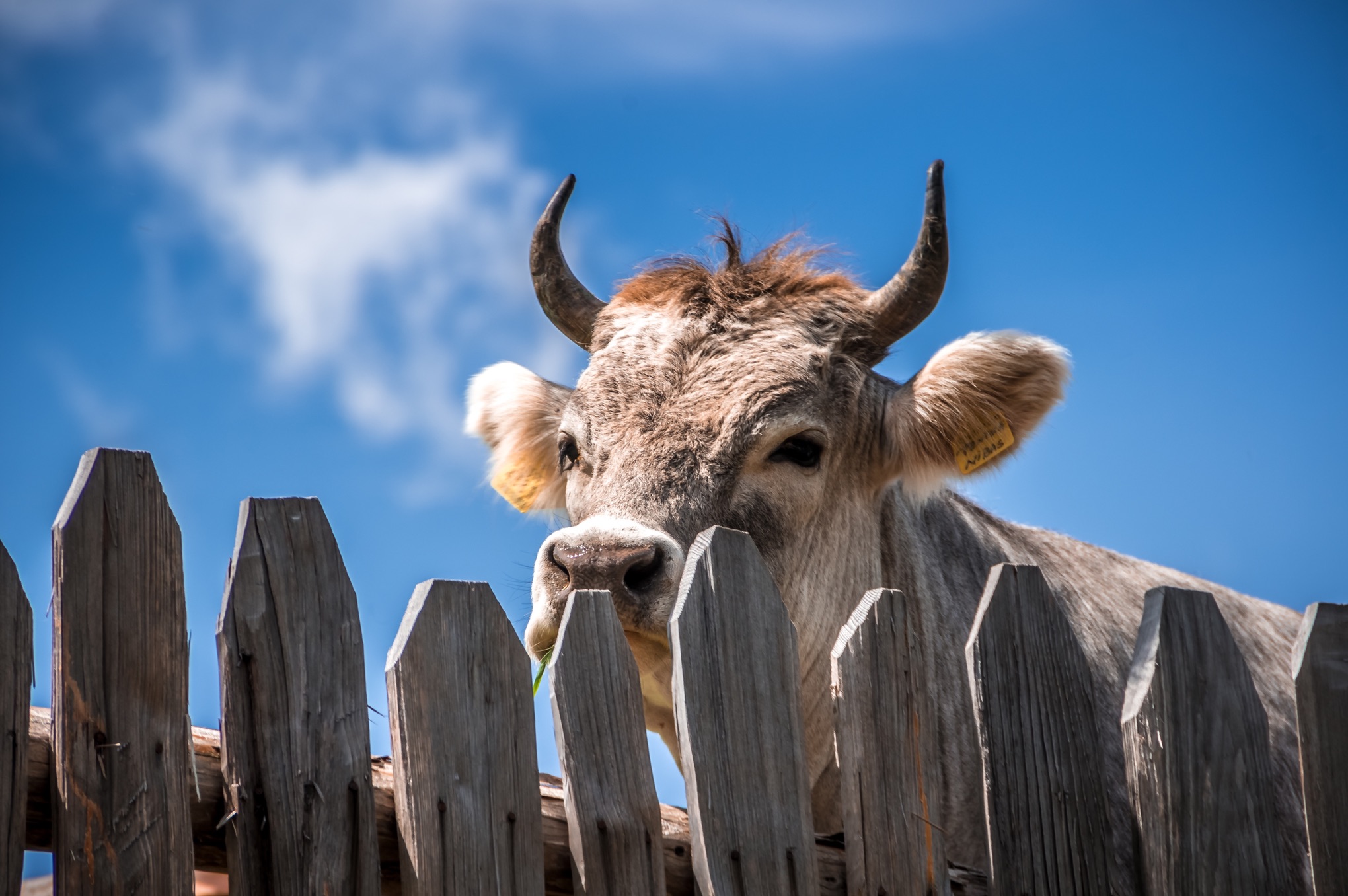}
\caption*{GT, DIV2K 0368}
\end{subfigure}
\hfill
\begin{subfigure}[b]{0.325\linewidth}
\centering
\includegraphics[width=\linewidth]{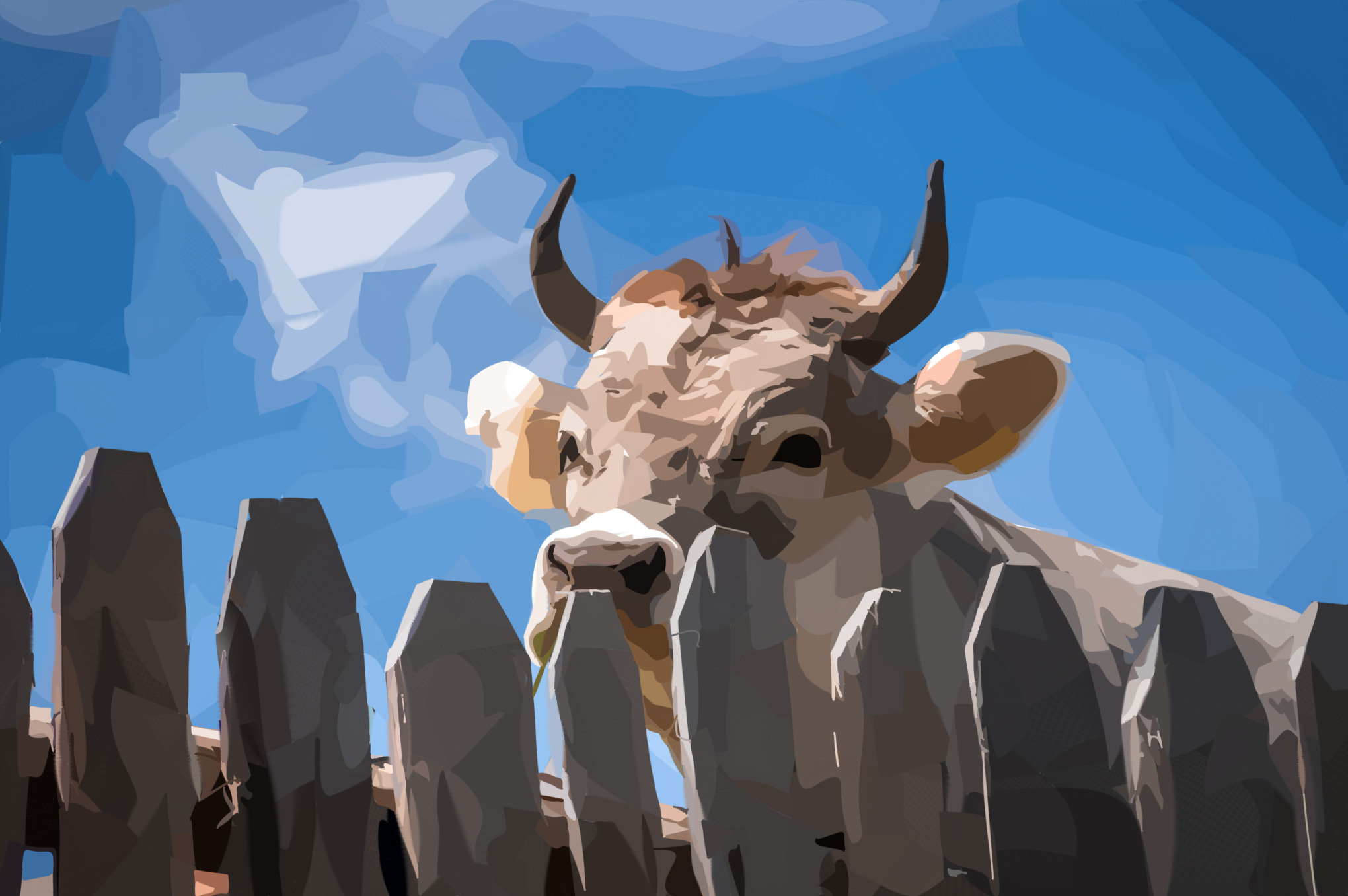}
\caption*{B\'ezier Splatting, 26.4344 dB}
\end{subfigure}
\hfill
\begin{subfigure}[b]{0.325\linewidth}
\centering
\includegraphics[width=\linewidth]{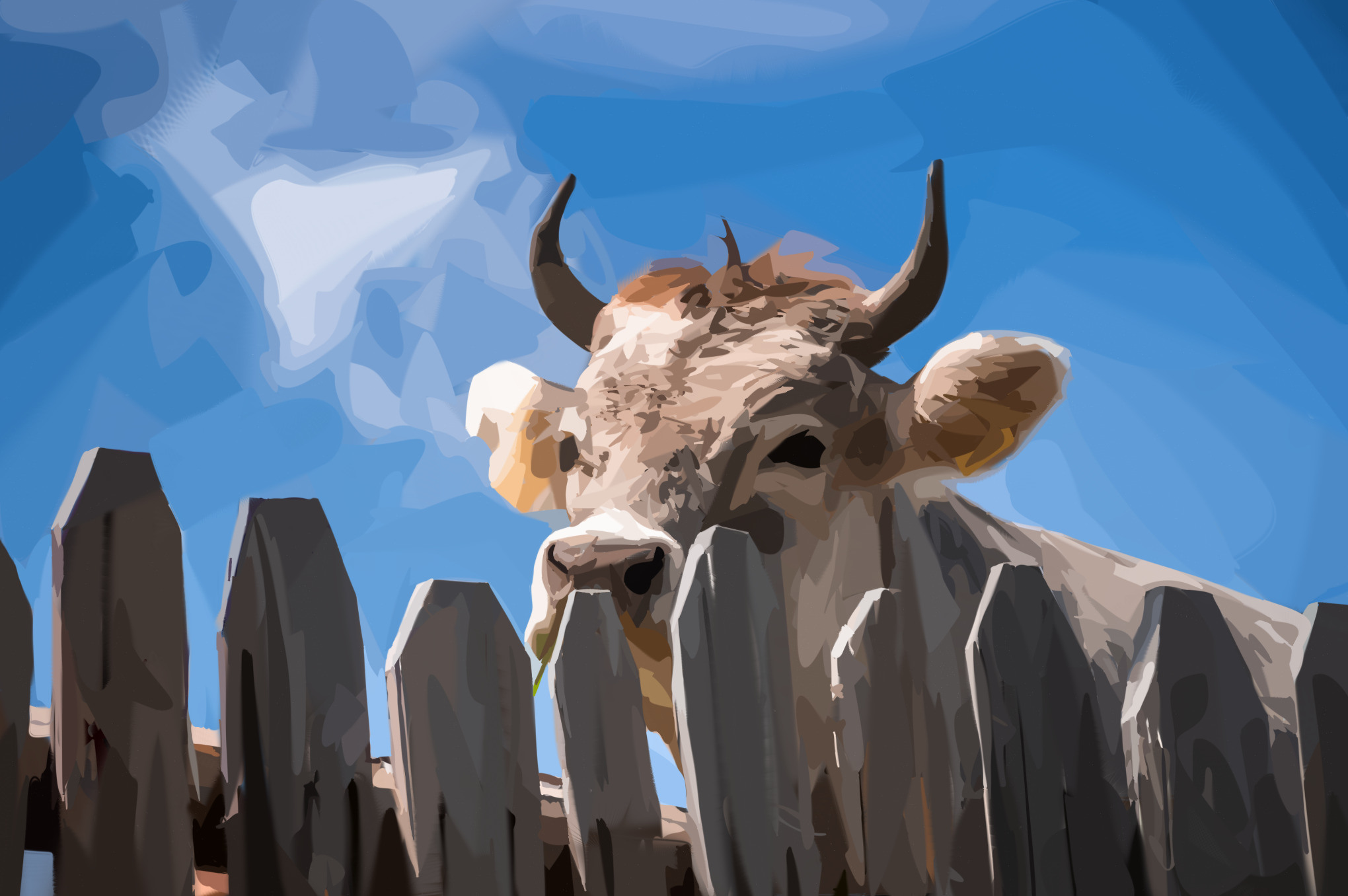}
\caption*{\textbf{Ours}, 27.3865 dB}
\end{subfigure}


\begin{subfigure}[b]{0.325\linewidth}
\centering
\includegraphics[width=\linewidth]{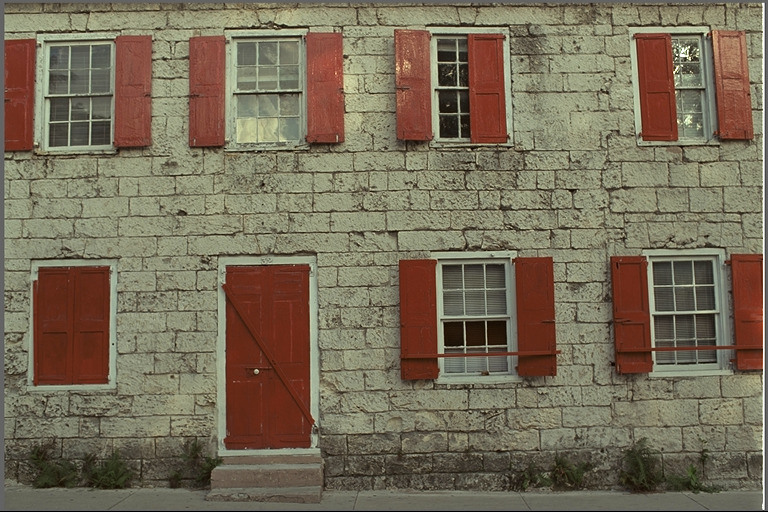}
\caption*{GT, Kodim 01}
\end{subfigure}
\hfill
\begin{subfigure}[b]{0.325\linewidth}
\centering
\includegraphics[width=\linewidth]{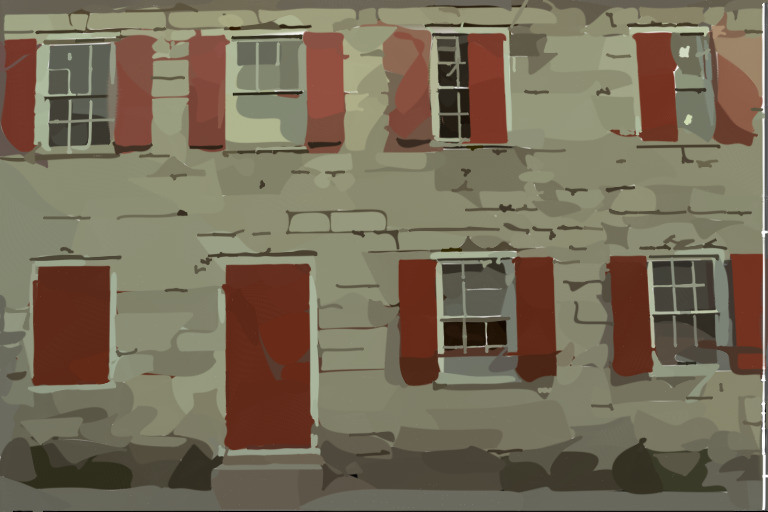}
\caption*{B\'ezier Splatting, 22.4223 dB}
\end{subfigure}
\hfill
\begin{subfigure}[b]{0.325\linewidth}
\centering
\includegraphics[width=\linewidth]{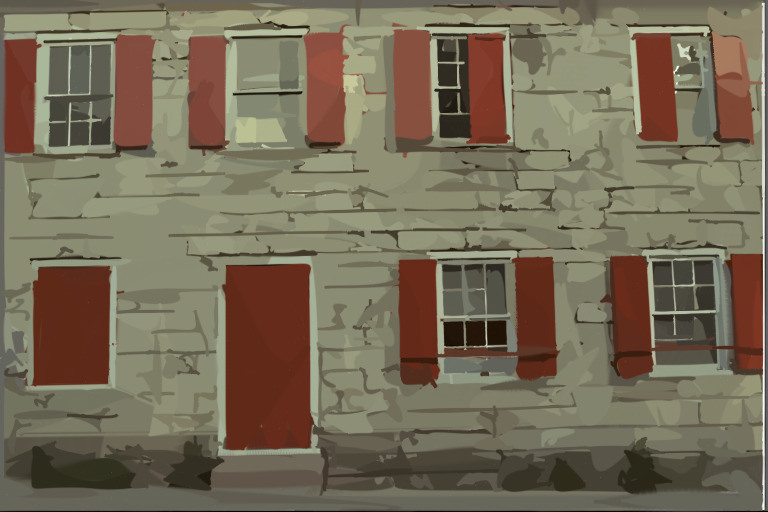}
\caption*{\textbf{Ours}, 23.1342 dB}
\end{subfigure}


\begin{subfigure}[b]{0.32\linewidth}
\centering
\includegraphics[width=\linewidth]{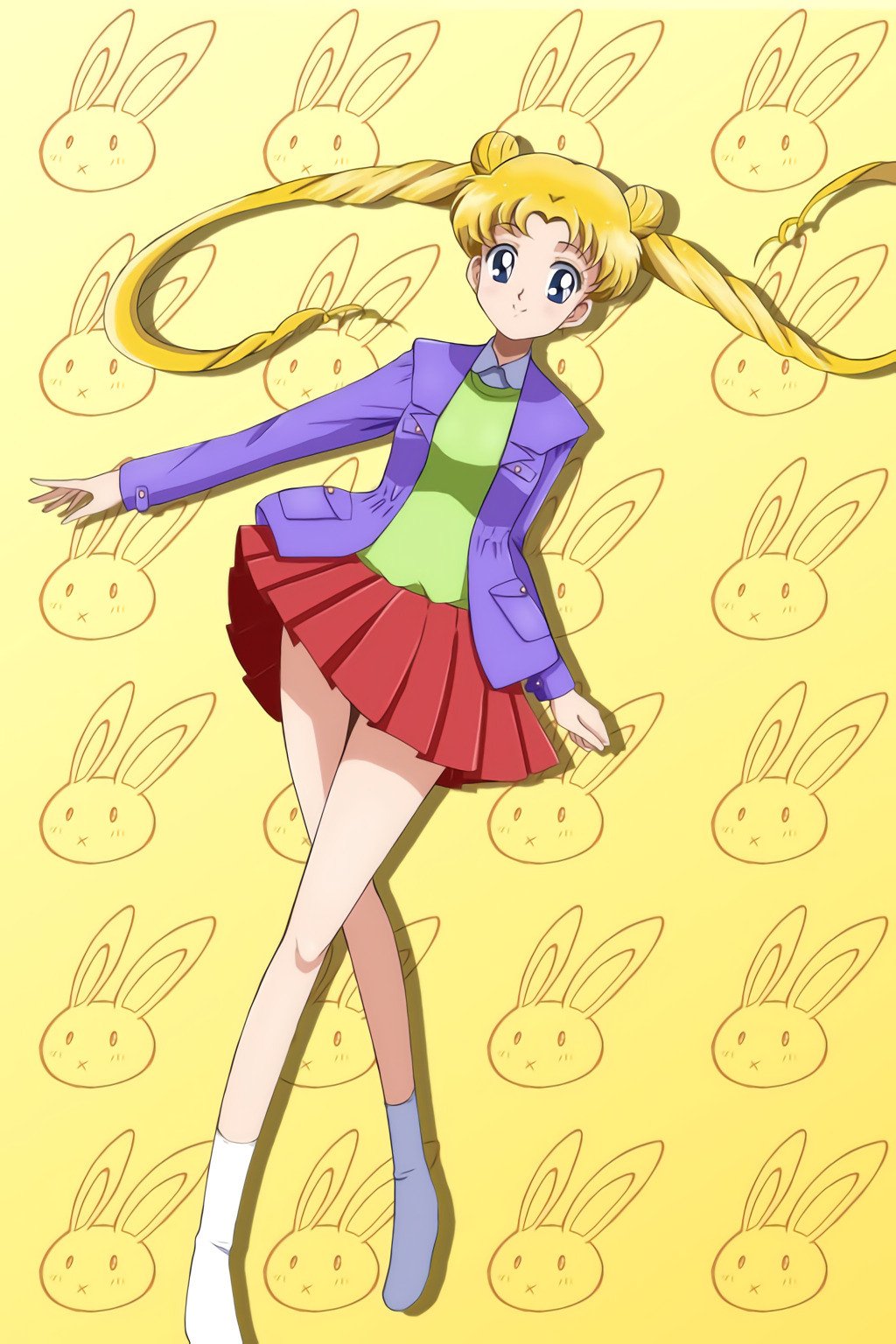}
\caption*{GT, DanbooRegion 20}
\end{subfigure}
\hfill
\begin{subfigure}[b]{0.32\linewidth}
\centering
\includegraphics[width=\linewidth]{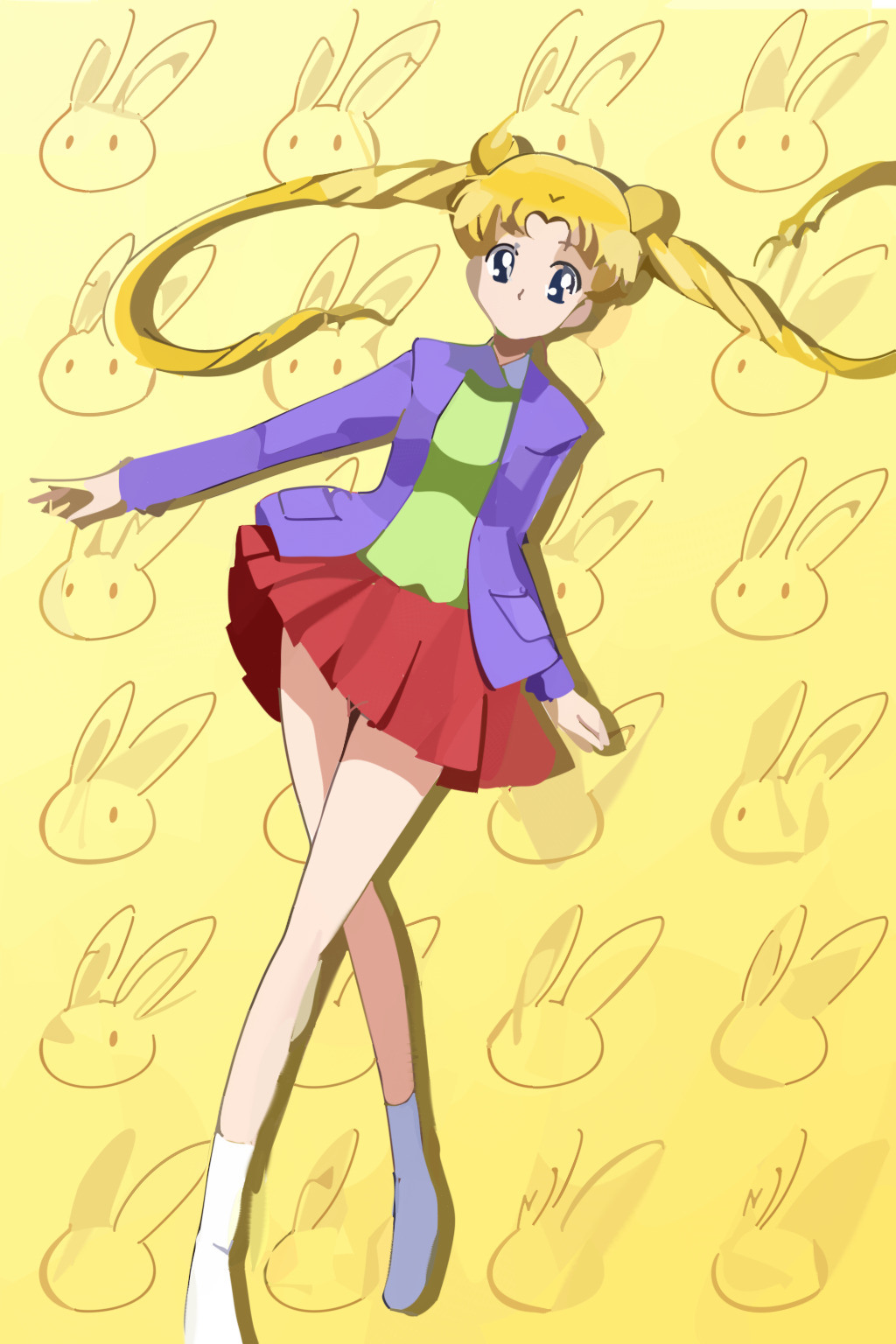}
\caption*{B\'ezier Splatting, 25.6231 dB}
\end{subfigure}
\hfill
\begin{subfigure}[b]{0.32\linewidth}
\centering
\includegraphics[width=\linewidth]{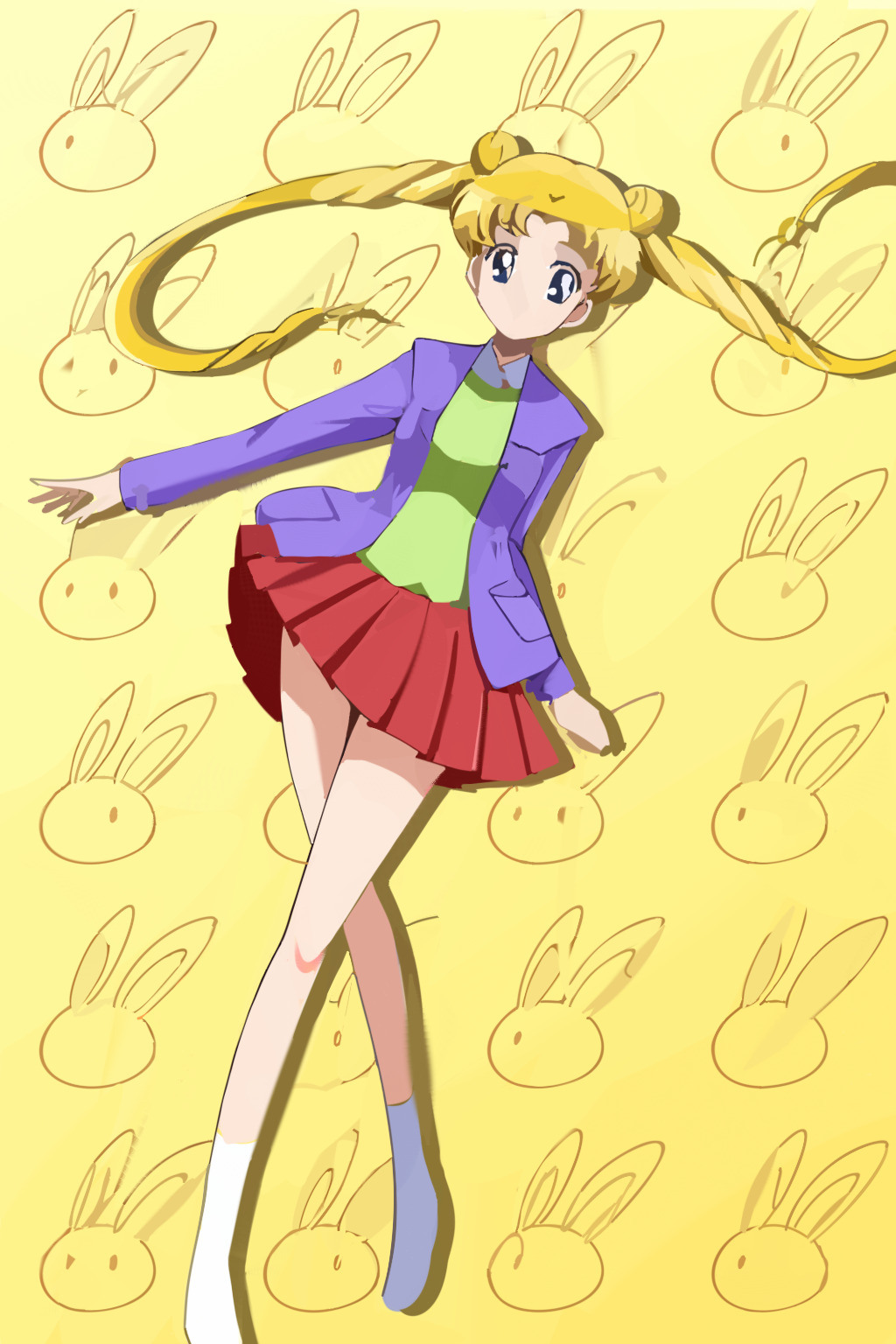}
\caption*{\textbf{Ours}, 27.4212 dB}
\end{subfigure}

\caption{\textbf{Qualitative Comparison.} Compared with the state-of-the-art differentiable vectorization method~\cite{liu2025bezier}, our method preserves fine structural details and coherent object boundaries under the same curve budget ($N=512$).}
\label{fig:qualitative}
\end{figure}

\subsubsection{DIV2K Experiment.}
To evaluate the robustness and generalizability of our framework, we compare our method with previous vectorization methods on the high-resolution DIV2K training dataset containing 800 images~\cite{agustsson2017ntire}.
This dataset is known for its extreme high-frequency textures and complex structures, originally designed for image super-resolution benchmarks.
Table~\ref{tab:div2k_quantitative} summarizes the results.
Even under this challenging evaluation, our \emph{Vector Scaffolding} consistently outperforms the state-of-the-art by a significant margin of roughly 0.5--0.6$\,\mathrm{dB}$ in PSNR.
Regarding perceptual quality measured by LPIPS, our method achieves the best scores at 512 and 1024 curves, while remaining comparable at 256 curves.
It is noteworthy that the baseline methods tend to show higher SSIM scores at some budgets.
This trade-off in SSIM can be plausibly explained as a consequence of our \emph{geometric regularization}, which prioritizes the professional usability of the resulting vector artwork.
SSIM evaluates on a local-window basis, which can favor locally detailed reconstructions even when they introduce less coherent vector structures.
The baseline methods allocate more capacity to local texture and edge fluctuations, whereas our hierarchical schedule tends to preserve larger coherent fills, which is more consistent with LPIPS at medium and high budgets.
This SSIM trade-off is consistent with our design goal of favoring well-stratified vector layers over aggressive local texture matching.

\subsubsection{Accelerated Optimization.}
As shown in the convergence plot in Figure~\ref{fig:figure_one_benchmark}, our method improves the speed-fidelity trade-off by enhancing \emph{both} reconstruction quality and optimization speed.
Our \emph{Vector Scaffolding} framework ensures granularity-wise scale separation and temporal Z-ordering, which stabilizes the optimization process and reduces the required gradient updates by allowing extremely high learning rates.
Consequently, we consistently accelerate the total optimization time by $2.5\times$ compared to the fastest baseline, B\'ezier Splatting~\cite{liu2025bezier}, while achieving the best PSNR scores.
We emphasize that this $2.5\times$ figure is measured in \emph{wall-clock time}; the per-step computational overhead of our method is in fact $1.25\times$ smaller than B\'ezier Splatting, since Interior Gradient Aggregation does not introduce a separate calculation stage and the rapid-inflation schedule further reduces the number of active primitives during the early phase.

\begin{figure}[t]
\centering

\begin{subfigure}[b]{0.155\linewidth}
\centering
\includegraphics[width=\linewidth]{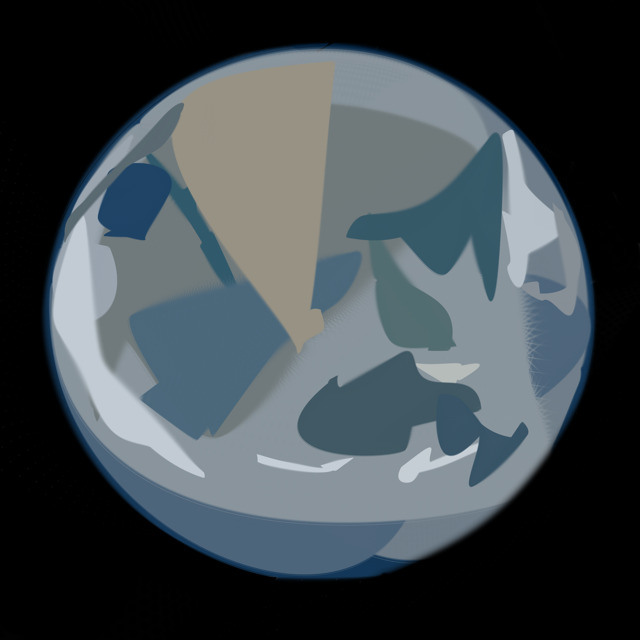}
\caption*{32 curves}
\end{subfigure}
\hfill
\begin{subfigure}[b]{0.155\linewidth}
\centering
\includegraphics[width=\linewidth]{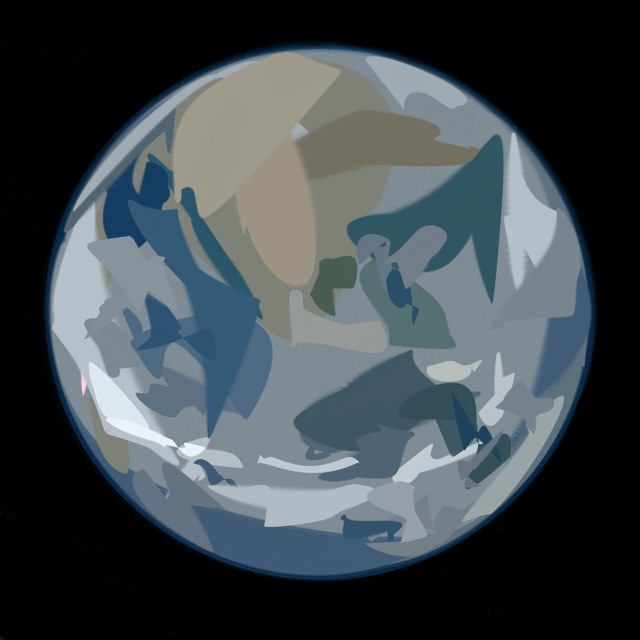}
\caption*{64 curves}
\end{subfigure}
\hfill
\begin{subfigure}[b]{0.155\linewidth}
\centering
\includegraphics[width=\linewidth]{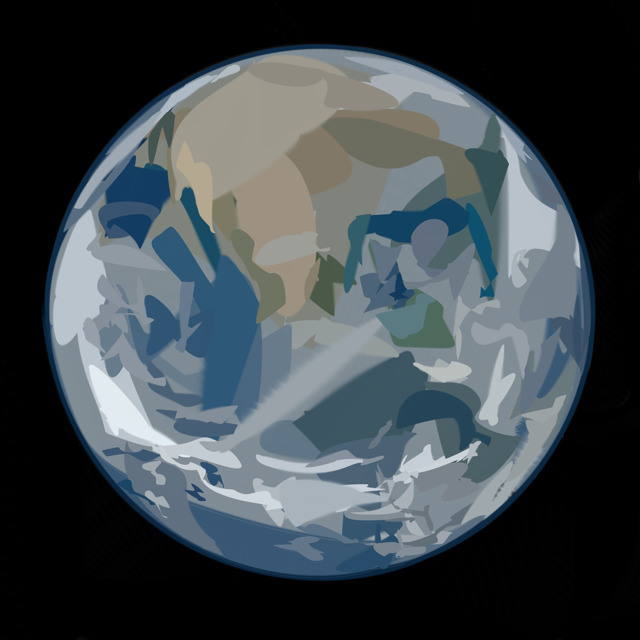}
\caption*{128 curves}
\end{subfigure}
\hfill
\begin{subfigure}[b]{0.155\linewidth}
\centering
\includegraphics[width=\linewidth]{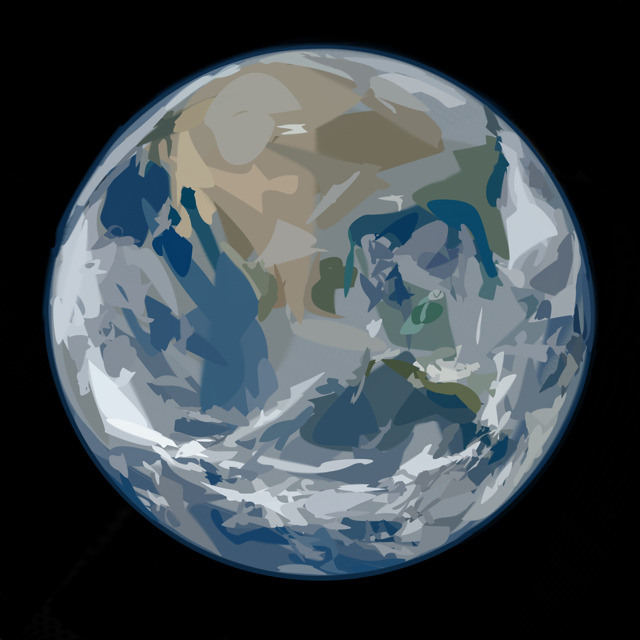}
\caption*{256 curves}
\end{subfigure}
\hfill
\begin{subfigure}[b]{0.155\linewidth}
\centering
\includegraphics[width=\linewidth]{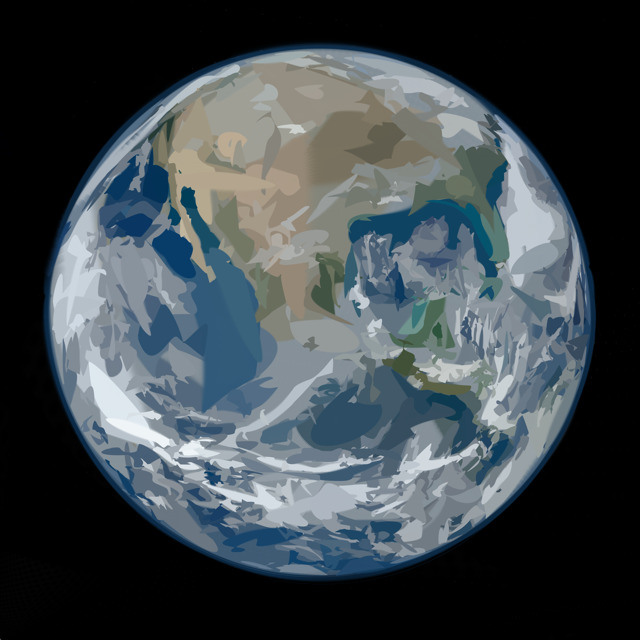}
\caption*{512 curves}
\end{subfigure}
\hfill
\begin{subfigure}[b]{0.155\linewidth}
\centering
\includegraphics[width=\linewidth]{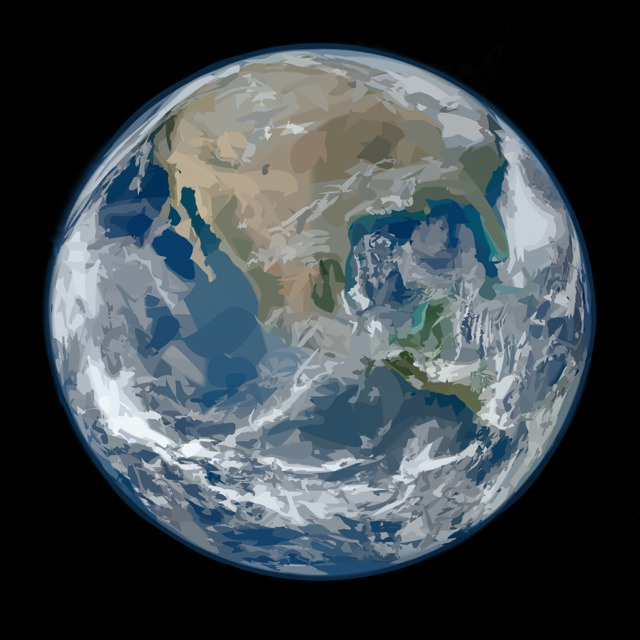}
\caption*{1024 curves}
\end{subfigure}

\begin{subfigure}[b]{0.155\linewidth}
\centering
\includegraphics[width=\linewidth]{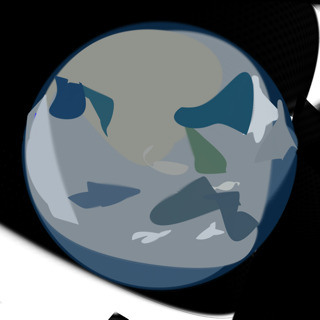}
\caption*{LoD 0}
\end{subfigure}
\hfill
\begin{subfigure}[b]{0.155\linewidth}
\centering
\includegraphics[width=\linewidth]{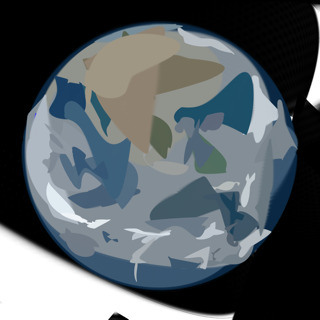}
\caption*{LoD 1}
\end{subfigure}
\hfill
\begin{subfigure}[b]{0.155\linewidth}
\centering
\includegraphics[width=\linewidth]{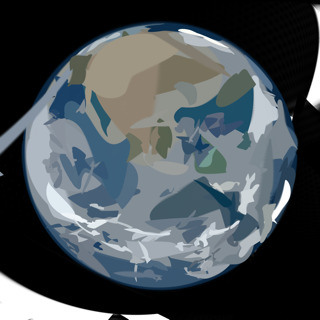}
\caption*{LoD 2}
\end{subfigure}
\hfill
\begin{subfigure}[b]{0.155\linewidth}
\centering
\includegraphics[width=\linewidth]{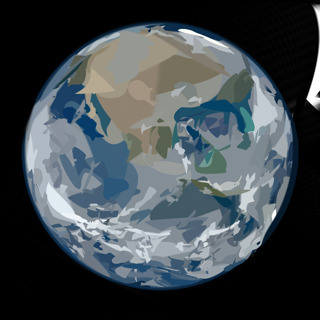}
\caption*{LoD 3}
\end{subfigure}
\hfill
\begin{subfigure}[b]{0.155\linewidth}
\centering
\includegraphics[width=\linewidth]{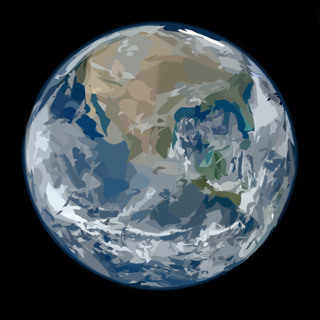}
\caption*{LoD 4}
\end{subfigure}
\hfill
\begin{subfigure}[b]{0.155\linewidth}
\centering
\includegraphics[width=\linewidth]{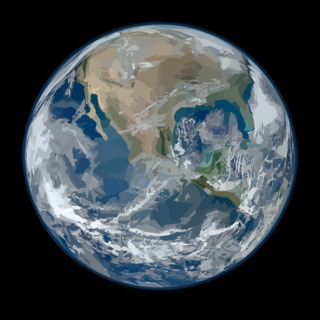}
\caption*{LoD 5}
\end{subfigure}

\caption{\textbf{LoD Control Demonstration.}
We fit our Vector Scaffolding to a super high-resolution image of the Earth ($8000 \times 8000$)~\cite{nasa2012blue}.
The first row shows the training dynamics at different curve counts, while the second row shows the level-of-detail (LoD) separation after fitting 1024 curves.}
\label{fig:earth_plots}
\end{figure}

\begin{figure}[t]
\centering
\begin{subfigure}[b]{0.32\linewidth}
\centering
\includegraphics[width=\linewidth]{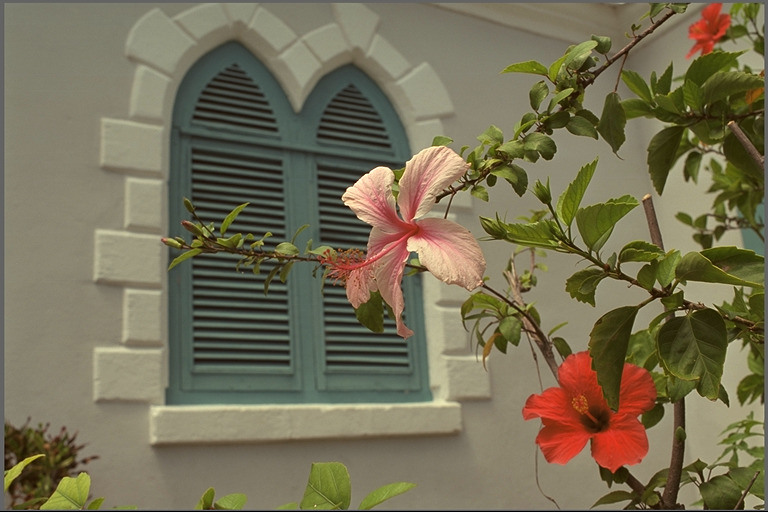}
\caption*{\parbox{\linewidth}{\centering (a) Ground truth\\\quad Kodim 07}}
\end{subfigure}
\hfill
\begin{subfigure}[b]{0.32\linewidth}
\centering
\includegraphics[width=\linewidth]{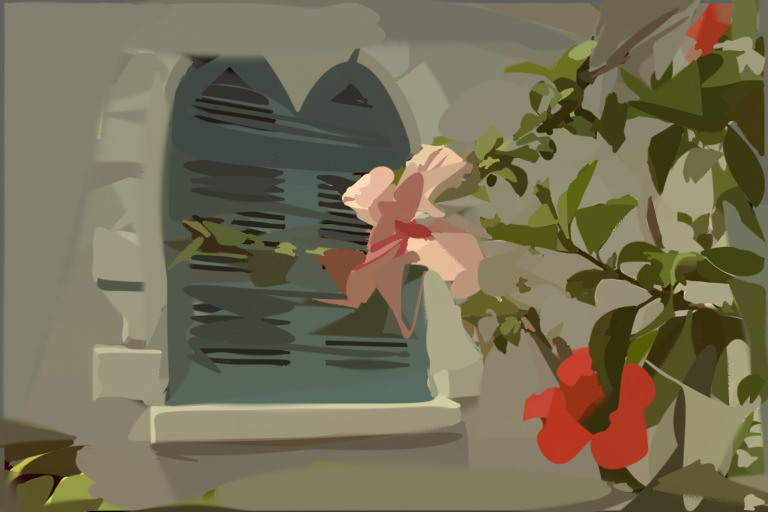}
\caption*{\parbox{\linewidth}{\centering (b) Without interior gradients\\\quad 23.4553 dB}}
\end{subfigure}
\hfill
\begin{subfigure}[b]{0.32\linewidth}
\centering
\includegraphics[width=\linewidth]{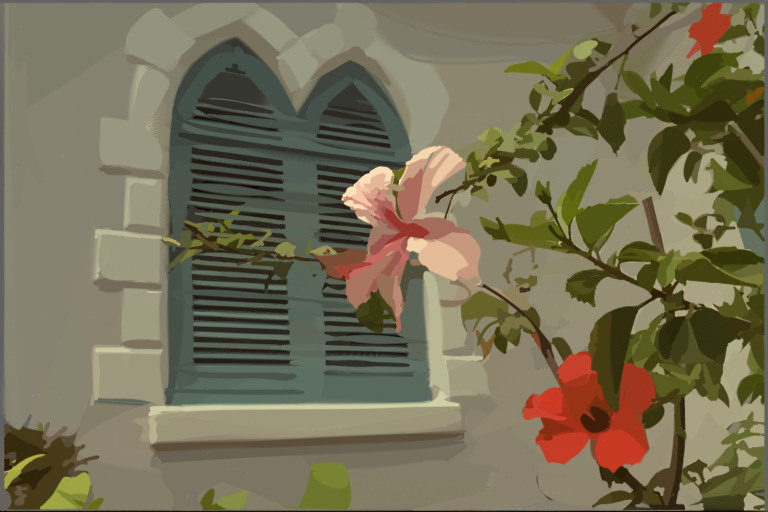}
\caption*{\parbox{\linewidth}{\centering (c) With interior gradients\\\quad \textbf{28.0802 dB}}}
\end{subfigure}
\caption{
\textbf{Effect of Interior Gradients.} (a) Ground truth. (b) Without interior gradients, the base curves lose their internal anchors, causing optimization drift and poor convergence. (c) With interior gradients, our method maintains structural integrity while capturing photometric information.}
\label{fig:ablation_decoupling}
\end{figure}

\begin{figure}[t]
\centering
\begin{subfigure}[b]{0.24\linewidth}
\centering
\includegraphics[width=\linewidth]{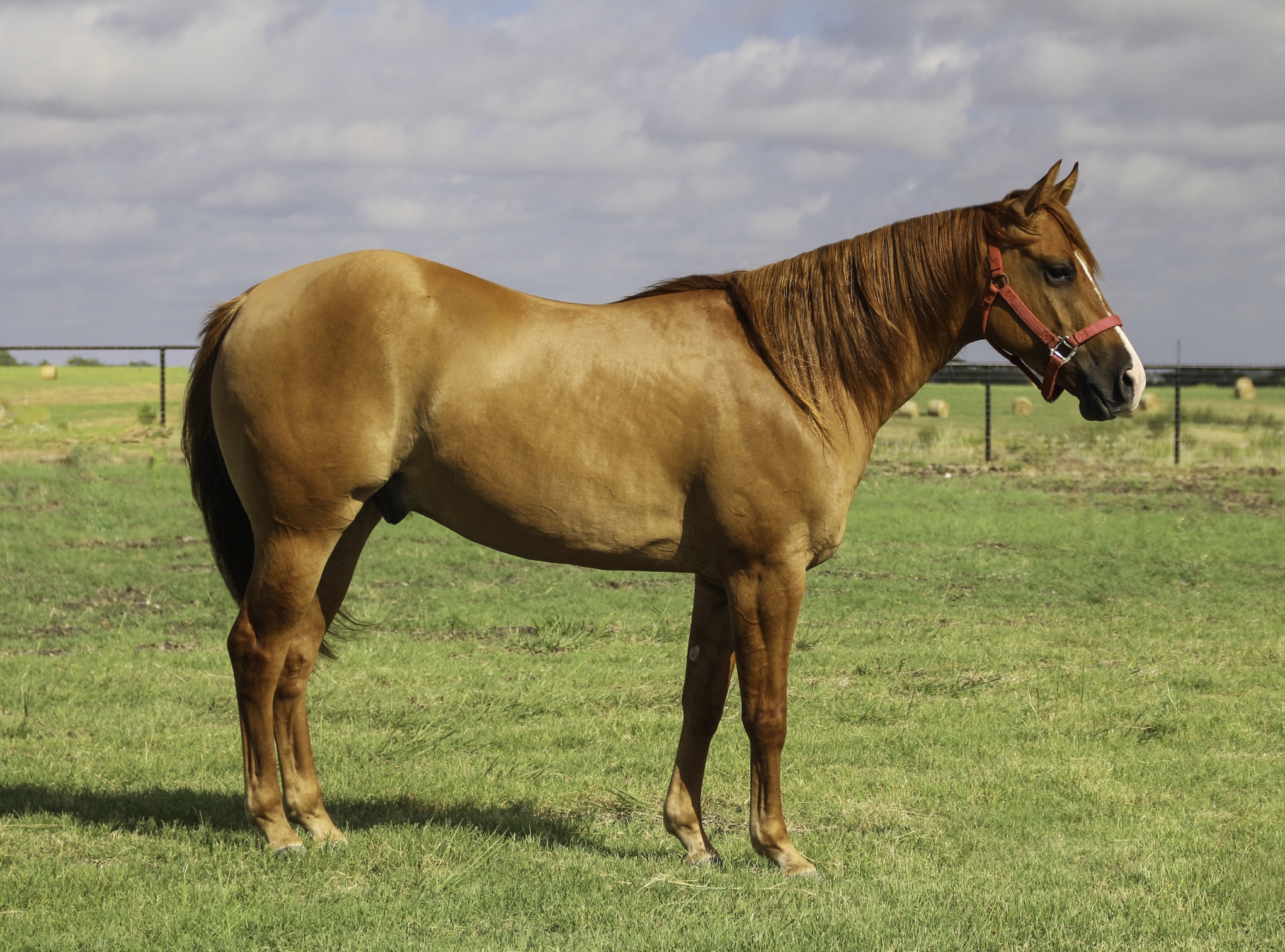}
\caption*{\parbox{\linewidth}{\centering (a) Ground truth\\\quad DIV2K 0383}}
\end{subfigure}
\hfill
\begin{subfigure}[b]{0.24\linewidth}
\centering
\includegraphics[width=\linewidth]{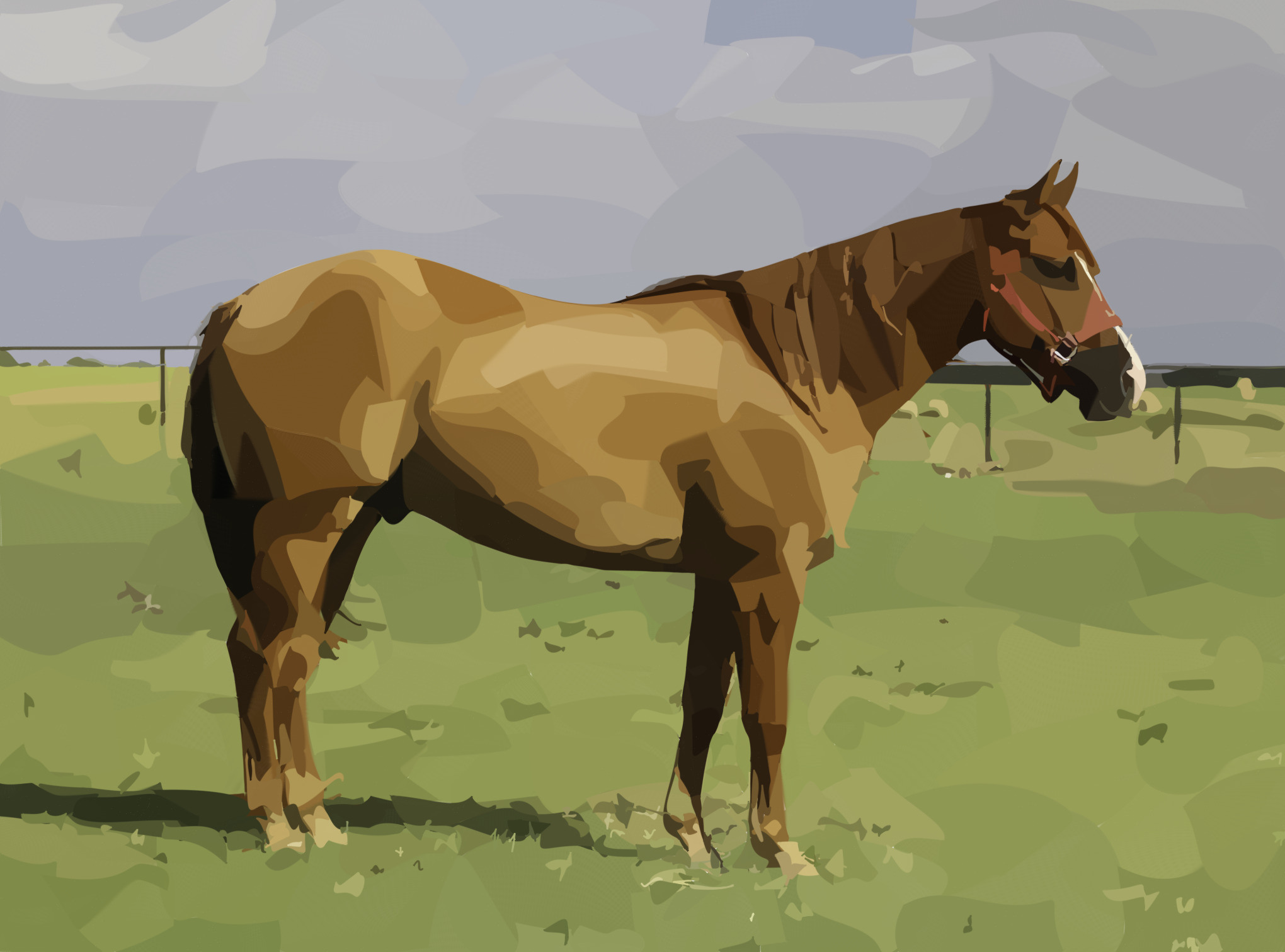}
\caption*{\parbox{\linewidth}{\centering (b) Baseline, $\text{lr} \times 1$\\\quad 24.4731 dB}}
\end{subfigure}
\hfill
\begin{subfigure}[b]{0.24\linewidth}
\centering
\includegraphics[width=\linewidth]{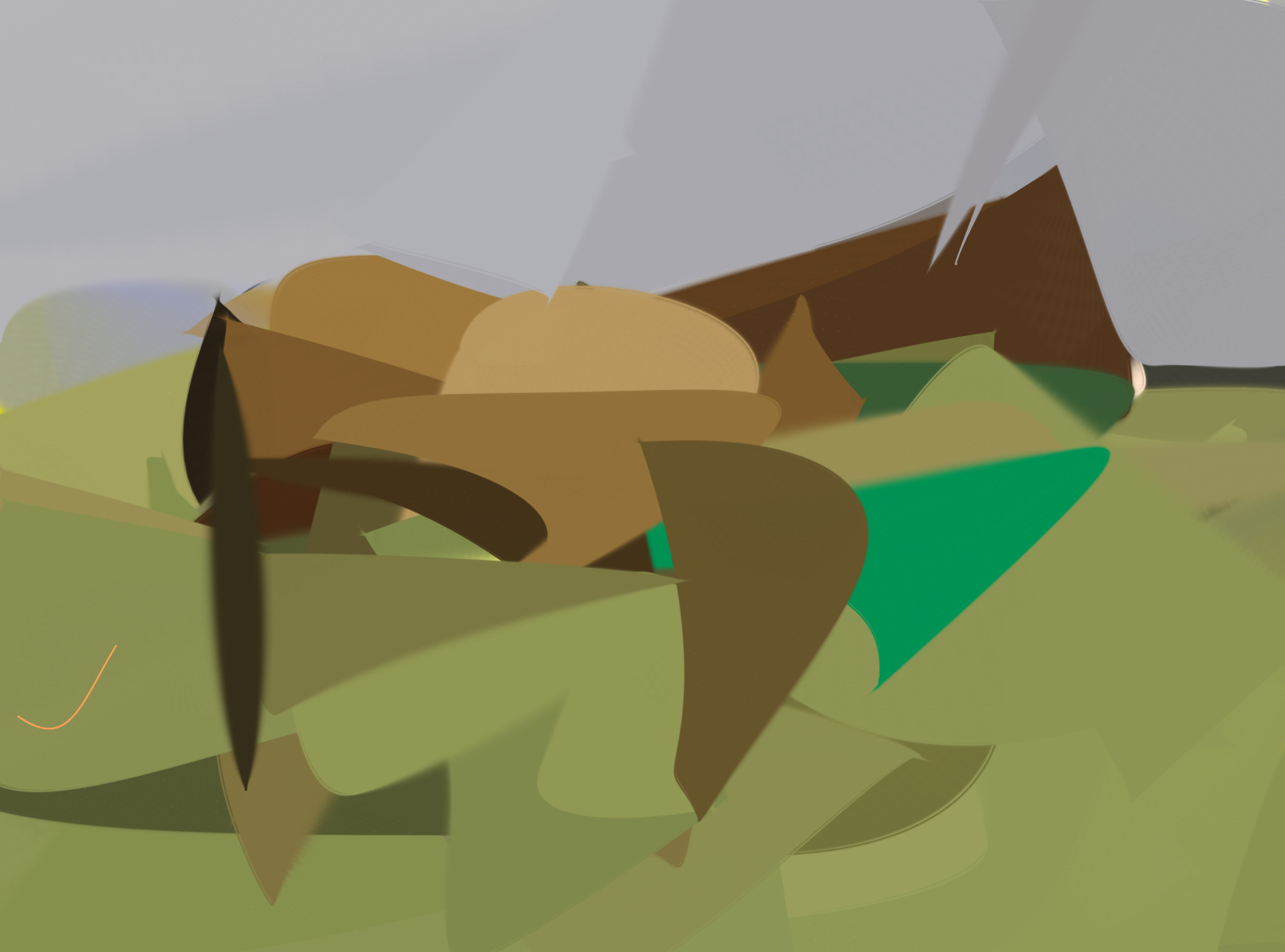}
\caption*{\parbox{\linewidth}{\centering (c) Baseline, $\text{lr} \times 50$\\\quad 19.1911 dB}}
\end{subfigure}
\hfill
\begin{subfigure}[b]{0.24\linewidth}
\centering
\includegraphics[width=\linewidth]{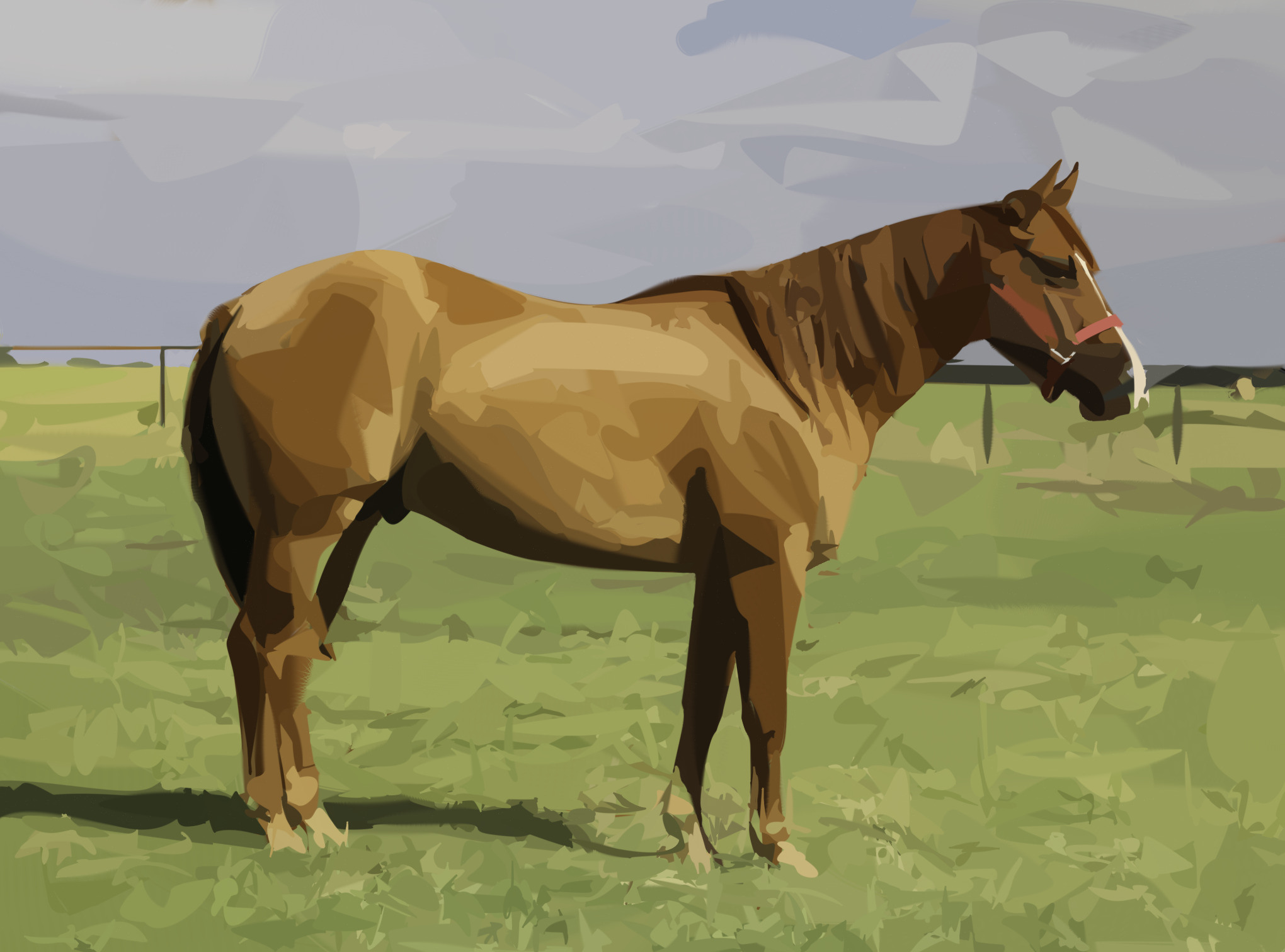}
\caption*{\parbox{\linewidth}{\centering (d) \textbf{Ours}, $\text{lr} \times 50$\\\quad \textbf{24.6808 dB}}}
\end{subfigure}
\caption{
\textbf{Hierarchical Scaffolding vs. Flat Optimization.} We ablate the \emph{Progressive Stratification} and temporal Z-ordering by reverting to a flat optimization strategy where all $N$ curves are initialized simultaneously and updated without spatial constraints. Flat optimization fails under our extreme learning rate.
This demonstrates the efficacy of our frequency-separation for fast and stable convergence.}
\label{fig:ablation_flat}
\end{figure}

\subsection{Structural Hierarchy and Arbitrary Level-of-Detail}
\label{subsec:lod_and_structure}

Since our structured densification scheduling via \emph{Progressive Stratification} and temporal Z-ordering enables monotonically increasing the number of curves, the vector representation can be densified sequentially from base structures to finest details.
Therefore, our Vector Scaffolding naturally supports flexible level-of-detail (LoD) rendering.
This is partially demonstrated in Tables~\ref{tab:kodak_quantitative} and~\ref{tab:div2k_quantitative}.
We further apply our method to a 64-megapixel high-resolution image to demonstrate flexible LoD control.
The results are shown in Figure~\ref{fig:earth_plots}.
By fixing all optimization hyperparameters such as the learning rate, the schedule, and the densification logic while varying only the number of densification steps, we can continuously increase the number of curves to achieve better reconstruction quality.
This mimics the typical artistic workflow, from simple base sketches to procedural polishing, providing designers with visually meaningful and easily editable layers, as shown in Figure~\ref{fig:figure_one} (left).

\subsection{Ablation Studies}
\label{subsec:ablation}

To validate our key ideas, we perform ablation studies isolating each component.
More ablation studies are provided in Appendix~\ref{appx:ablation}.

\subsubsection{Efficacy of Interior Gradient.}
We ablate the interior gradient in Figure~\ref{fig:ablation_decoupling}.
Previous methods~\cite{liu2025bezier} neglect the positional gradient from the interior of the curve, which leads to topology collapse.
Without interior gradients, the base curves lose their internal anchors, causing optimization drift and significantly degrading the PSNR.
We show that interior gradients and boundary gradients jointly contribute to the PSNR and preserve structural integrity.

\subsubsection{Hierarchical Scaffolding vs. Flat Optimization.}
We ablate \emph{Progressive Stratification} and temporal Z-ordering by reverting to a flat optimization strategy where all $N$ curves are initialized simultaneously and updated without spatial constraints.
The results are shown in Figure~\ref{fig:ablation_flat}.
Under our aggressive learning rate, this flat optimization completely fails, resulting in extreme gradient oscillations and diverging losses.
This suggests that our frequency-separation is an effective technique for fast and stable convergence.

\section{Conclusion}
\label{sec:5_concl}
In this work, we presented \textbf{Vector Scaffolding}, a hierarchical optimization framework for differentiable image vectorization.
Previous methods focused on improving the rendering backend; however, we found that overall performance including reconstruction quality (PSNR), perceptual quality (LPIPS), optimization speed, and editability is limited by the topological structure of the vector representation.
Our main strategy is to stabilize the optimization dynamics by matching the hierarchical structure of the optimization process to that of natural and articulated images.
This structural resonance enables a high learning rate regime without causing instability, allowing us to focus directly on topology construction.
Without altering the underlying renderer, our framework achieves a $2.5\times$ speedup in wall clock time and substantial PSNR gains under matched curve budgets, demonstrating that optimization structure is a major determinant of vectorization quality.

\section*{Acknowledgements}
This work was supported in part by the IITPgrants [No. RS-2021-II211343, Artificial Intelligence Graduate School Program (Seoul National University), No. RS-2025-02303870, No.2022-0-00156] funded by the Korea government (MSIT).

\bibliographystyle{splncs04}
\bibliography{main}

@String(CVPR  = {IEEE Conf. Comput. Vis. Pattern Recog.})

@String(ICCV  = {Int. Conf. Comput. Vis.})

@String(ECCV  = {Eur. Conf. Comput. Vis.})

@String(ICLR  = {Int. Conf. Learn. Represent.})

@String(CVPRW = {IEEE Conf. Comput. Vis. Pattern Recog. Worksh.})

@String(AAAI  = {AAAI})

@inproceedings{agustsson2017ntire,
  title={{NTIRE} 2017 Challenge on Single Image Super-Resolution: Dataset and Study},
  author={Agustsson, Eirikur and Timofte, Radu},
  booktitle={Proceedings of the IEEE Conference on Computer Vision and Pattern Recognition Workshops (CVPRW)},
  pages={1122--1131},
  year={2017},
  doi={10.1109/CVPRW.2017.150}
}

@inproceedings{cao2023svgformer,
  title={{SVGformer}: Representation Learning for Continuous Vector Graphics Using Transformers},
  author={Cao, Defu and Wang, Zhaowen and Echevarria, Jose and Liu, Yan},
  booktitle={Proceedings of the IEEE/CVF Conference on Computer Vision and Pattern Recognition (CVPR)},
  pages={10093--10102},
  year={2023}
}

@inproceedings{chen2024highfidelity,
  title={Towards High-fidelity Artistic Image Vectorization via Texture-Encapsulated Shape Parameterization},
  author={Chen, Ye and Ni, Bingbing and Liu, Jinfan and Huang, Xiaoyang and Chen, Xuanhong},
  booktitle={Proceedings of the IEEE/CVF Conference on Computer Vision and Pattern Recognition (CVPR)},
  pages={15877--15886},
  year={2024},
  doi={10.1109/CVPR52733.2024.01503}
}

@article{du2023lineargradient,
  title={Image Vectorization and Editing via Linear Gradient Layer Decomposition},
  author={Du, Zheng-Jun and Kang, Liang-Fu and Tan, Jianchao and Gingold, Yotam and Xu, Kun},
  journal={ACM Transactions on Graphics},
  volume={42},
  number={4},
  articleno={97},
  pages={1--13},
  year={2023},
  doi={10.1145/3592128}
}

@inproceedings{guo2025tetsphere,
  title={{TetSphere} Splatting: Representing High-Quality Geometry with Lagrangian Volumetric Meshes},
  author={Guo, Minghao and Wang, Bohan and He, Kaiming and Matusik, Wojciech},
  booktitle={International Conference on Learning Representations (ICLR)},
  year={2025}
}

@inproceedings{guedon2024sugar,
  title={{SuGaR}: Surface-Aligned Gaussian Splatting for Efficient {3D} Mesh Reconstruction and High-Quality Mesh Rendering},
  author={Gu{\'e}don, Antoine and Lepetit, Vincent},
  booktitle={Proceedings of the IEEE/CVF Conference on Computer Vision and Pattern Recognition (CVPR)},
  pages={5354--5363},
  year={2024},
  doi={10.1109/CVPR52733.2024.00512}
}

@article{hirschorn2024optimize,
  title={Optimize \& Reduce: A Top-Down Approach for Image Vectorization},
  author={Hirschorn, Or and Jevnisek, Amir and Avidan, Shai},
  journal={Proceedings of the AAAI Conference on Artificial Intelligence},
  volume={38},
  number={3},
  pages={2148--2156},
  year={2024},
  doi={10.1609/aaai.v38i3.27987}
}

@inproceedings{ho2020ddpm,
  title={Denoising Diffusion Probabilistic Models},
  author={Ho, Jonathan and Jain, Ajay N. and Abbeel, Pieter},
  booktitle={Advances in Neural Information Processing Systems},
  volume={33},
  pages={6840--6851},
  year={2020}
}

@inproceedings{huang2024twodgs,
  title={{2D} Gaussian Splatting for Geometrically Accurate Radiance Fields},
  author={Huang, Binbin and Yu, Zehao and Chen, Anpei and Geiger, Andreas and Gao, Shenghua},
  booktitle={ACM SIGGRAPH 2024 Conference Papers},
  articleno={32},
  pages={1--11},
  year={2024},
  publisher={Association for Computing Machinery},
  doi={10.1145/3641519.3657428}
}

@inproceedings{jain2023vectorfusion,
  title={{VectorFusion}: Text-to-{SVG} by Abstracting Pixel-Based Diffusion Models},
  author={Jain, Ajay and Xie, Amber and Abbeel, Pieter},
  booktitle={Proceedings of the IEEE/CVF Conference on Computer Vision and Pattern Recognition (CVPR)},
  pages={1911--1920},
  year={2023},
  doi={10.1109/CVPR52729.2023.00190}
}

@article{kerbl2023gaussian,
  title={{3D} Gaussian Splatting for Real-Time Radiance Field Rendering},
  author={Kerbl, Bernhard and Kopanas, Georgios and Leimk{\"u}hler, Thomas and Drettakis, George},
  journal={ACM Transactions on Graphics},
  volume={42},
  number={4},
  articleno={139},
  pages={1--14},
  year={2023},
  doi={10.1145/3592433}
}

@misc{kodak1999lossless,
  title={{Kodak} Lossless True Color Image Suite},
  author={{Eastman Kodak Company}},
  year={1999},
  howpublished={Dataset},
  url={https://r0k.us/graphics/kodak/},
  note={Accessed: 2026-05-12}
}

@article{li2020diffvg,
  title={Differentiable Vector Graphics Rasterization for Editing and Learning},
  author={Li, Tzu-Mao and Luk{\'a}{\v c}, Michal and Gharbi, Micha{\"e}l and Ragan-Kelley, Jonathan},
  journal={ACM Transactions on Graphics},
  volume={39},
  number={6},
  articleno={193},
  pages={1--15},
  year={2020},
  doi={10.1145/3414685.3417871}
}

@inproceedings{liu2025bezier,
  title={B{\'e}zier Splatting for Fast and Differentiable Vector Graphics Rendering},
  author={Liu, Xi and Zhou, Chaoyi and Zhao, Nanxuan and Huang, Siyu},
  booktitle={Advances in Neural Information Processing Systems},
  year={2025}
}

@inproceedings{lopes2019learned,
  title={A Learned Representation for Scalable Vector Graphics},
  author={Lopes, Raphael Gontijo and Ha, David and Eck, Douglas and Shlens, Jonathon},
  booktitle={Proceedings of the IEEE/CVF International Conference on Computer Vision (ICCV)},
  pages={7930--7939},
  year={2019},
  doi={10.1109/ICCV.2019.00802}
}

@inproceedings{luiten2024dynamic3dgaussians,
  title={Dynamic {3D} Gaussians: Tracking by Persistent Dynamic View Synthesis},
  author={Luiten, Jonathon Tyler and Kopanas, Georgios and Leibe, Bastian and Ramanan, Deva},
  booktitle={2024 International Conference on 3D Vision (3DV)},
  pages={800--809},
  year={2024},
  doi={10.1109/3DV62453.2024.00044}
}

@inproceedings{ma2022live,
  title={Towards Layer-Wise Image Vectorization},
  author={Ma, Xu and Zhou, Yuqian and Xu, Xingqian and Sun, Bin and Filev, Valerii and Orlov, Nikita and Fu, Yun and Shi, Humphrey},
  booktitle={Proceedings of the IEEE/CVF Conference on Computer Vision and Pattern Recognition (CVPR)},
  pages={16314--16323},
  year={2022},
  doi={10.1109/CVPR52688.2022.01583}
}

@article{mueller2022instant,
  title={Instant Neural Graphics Primitives with a Multiresolution Hash Encoding},
  author={M{\"u}ller, Thomas and Evans, Alex and Schied, Christoph and Keller, Alexander},
  journal={ACM Transactions on Graphics},
  volume={41},
  number={4},
  articleno={102},
  pages={1--15},
  year={2022},
  doi={10.1145/3528223.3530127}
}

@misc{nasa2012blue,
  title={Most Amazing High Definition Image of Earth -- Blue Marble 2012},
  author={{NASA Goddard Photo and Video}},
  year={2012},
  howpublished={Flickr image, NASA Goddard Space Flight Center},
  url={https://www.flickr.com/photos/gsfc/6760135001},
  note={Public domain (NASA media usage guidelines). Accessed: 2026-05-12.}
}

@inproceedings{reddy2021im2vec,
  title={{Im2Vec}: Synthesizing Vector Graphics Without Vector Supervision},
  author={Reddy, Pradyumna and Gharbi, Michael and Luk{\'a}{\v c}, Michal and Mitra, Niloy J.},
  booktitle={Proceedings of the IEEE/CVF Conference on Computer Vision and Pattern Recognition (CVPR)},
  pages={7342--7351},
  year={2021},
  doi={10.1109/CVPR46437.2021.00726}
}

@inproceedings{sitzmann2020siren,
  title={Implicit Neural Representations with Periodic Activation Functions},
  author={Sitzmann, Vincent and Martel, Julien N. P. and Bergman, Alexander W. and Lindell, David B. and Wetzstein, Gordon},
  booktitle={Advances in Neural Information Processing Systems},
  volume={33},
  pages={7462--7473},
  year={2020}
}

@inproceedings{tang2024dreamgaussian,
  title={{DreamGaussian}: Generative Gaussian Splatting for Efficient {3D} Content Creation},
  author={Tang, Jiaxiang and Ren, Jiawei and Zhou, Hang and Liu, Ziwei and Zeng, Gang},
  booktitle={International Conference on Learning Representations (ICLR)},
  year={2024}
}

@inproceedings{wang2025livss,
  title={Layered Image Vectorization via Semantic Simplification},
  author={Wang, Zhenyu and Huang, Jianxi and Sun, Zhida and Gong, Yuanhao and Cohen-Or, Daniel and Lu, Min},
  booktitle={Proceedings of the IEEE/CVF Conference on Computer Vision and Pattern Recognition (CVPR)},
  pages={7728--7738},
  year={2025}
}

@inproceedings{wu2024fourdfs,
  title={{4D} Gaussian Splatting for Real-Time Dynamic Scene Rendering},
  author={Wu, Guanjun and Yi, Taoran and Fang, Jiemin and Xie, Lingxi and Zhang, Xiaopeng and Wei, Wei and Liu, Wenyu and Tian, Qi and Wang, Xinggang},
  booktitle={Proceedings of the IEEE/CVF Conference on Computer Vision and Pattern Recognition (CVPR)},
  pages={20310--20320},
  year={2024},
  doi={10.1109/CVPR52733.2024.01920}
}

@article{xie2024adan,
  title={{Adan}: Adaptive Nesterov Momentum Algorithm for Faster Optimizing Deep Models},
  author={Xie, Xingyu and Zhou, Pan and Li, Huan and Lin, Zhouchen and Yan, Shuicheng},
  journal={IEEE Transactions on Pattern Analysis and Machine Intelligence},
  volume={46},
  number={12},
  pages={9508--9520},
  year={2024},
  doi={10.1109/TPAMI.2024.3423382}
}

@inproceedings{xing2023diffsketcher,
  title={{DiffSketcher}: Text Guided Vector Sketch Synthesis through Latent Diffusion Models},
  author={Xing, XiMing and Wang, Chuang and Zhou, Haitao and Zhang, Jing and Yu, Qian and Xu, Dong},
  booktitle={Advances in Neural Information Processing Systems},
  volume={36},
  pages={15869--15889},
  year={2023}
}

@inproceedings{xing2024svgdreamer,
  title={{SVGDreamer}: Text Guided {SVG} Generation with Diffusion Model},
  author={Xing, Ximing and Zhou, Haitao and Wang, Chuang and Zhang, Jing and Xu, Dong and Yu, Qian},
  booktitle={Proceedings of the IEEE/CVF Conference on Computer Vision and Pattern Recognition (CVPR)},
  pages={4546--4555},
  year={2024},
  doi={10.1109/CVPR52733.2024.00435}
}

@inproceedings{yi2024gaussiandreamer,
  title={{GaussianDreamer}: Fast Generation from Text to {3D} Gaussians by Bridging {2D} and {3D} Diffusion Models},
  author={Yi, Taoran and Fang, Jiemin and Wang, Junjie and Wu, Guanjun and Xie, Lingxi and Zhang, Xiaopeng and Liu, Wenyu and Tian, Qi and Wang, Xinggang},
  booktitle={Proceedings of the IEEE/CVF Conference on Computer Vision and Pattern Recognition (CVPR)},
  pages={6796--6807},
  year={2024},
  doi={10.1109/CVPR52733.2024.00649}
}

@article{zhang2024texttovector,
  title={Text-to-Vector Generation with Neural Path Representation},
  author={Zhang, Peiying and Zhao, Nanxuan and Liao, Jing},
  journal={ACM Transactions on Graphics},
  volume={43},
  number={4},
  articleno={36},
  pages={1--13},
  year={2024},
  doi={10.1145/3658204}
}

@inproceedings{zhang2018lpips,
  title={The Unreasonable Effectiveness of Deep Features as a Perceptual Metric},
  author={Zhang, Richard and Isola, Phillip and Efros, Alexei A. and Shechtman, Eli and Wang, Oliver},
  booktitle={Proceedings of the IEEE Conference on Computer Vision and Pattern Recognition (CVPR)},
  pages={586--595},
  year={2018},
  doi={10.1109/CVPR.2018.00068}
}

@inproceedings{zhang2024gaussianimage,
  title={{GaussianImage}: 1000 {FPS} Image Representation and Compression by {2D} Gaussian Splatting},
  author={Zhang, Xinjie and Ge, Xingtong and Xu, Tongda and He, Dailan and Wang, Yan and Qin, Hongwei and Lu, Guo and Geng, Jing and Zhang, Jun},
  booktitle={Computer Vision -- ECCV 2024},
  series={Lecture Notes in Computer Science},
  volume={15067},
  pages={327--345},
  year={2024},
  publisher={Springer},
  doi={10.1007/978-3-031-72673-6_18}
}

@inproceedings{zhang2025imagegs,
  title={{Image-GS}: Content-Adaptive Image Representation via {2D} Gaussians},
  author={Zhang, Yunxiang and Li, Bingxuan and Kuznetsov, Alexandr and Jindal, Akshay and Diolatzis, Stavros and Chen, Kenneth and Sochenov, Anton and Kaplanyan, Anton and Sun, Qi},
  booktitle={ACM SIGGRAPH 2025 Conference Papers},
  articleno={102},
  pages={1--11},
  year={2025},
  publisher={Association for Computing Machinery},
  doi={10.1145/3721238.3730596}
}

@inproceedings{zwicker2001ewa,
  title={{EWA} Volume Splatting},
  author={Zwicker, Matthias and Pfister, Hanspeter and van Baar, Jeroen and Gross, Markus},
  booktitle={Proceedings Visualization, 2001. VIS '01},
  pages={29--36},
  year={2001},
  publisher={IEEE Computer Society},
  doi={10.5555/601671.601674}
}

\clearpage
\appendix
\renewcommand{\thesection}{\Alph{section}}
\setcounter{section}{0}

\begin{center}
\vspace*{0.5cm}
{\LARGE\bfseries Supplementary Material}\\[1.5em]
{\large Vector Scaffolding: Hierarchical Optimization for Fast and Stable Differentiable Image Vectorization}\\[1em]
{\normalsize Jaerin Lee, Kanggeon Lee, Kyoung Mu Lee}
\end{center}

\addcontentsline{toc}{section}{Supplementary Material}
\section{Algorithmic Implementation Details}
\label{appx:algorithm}

In the main paper, we introduced \emph{Progressive Stratification} and \emph{Rapid Inflation Scheduling} to orchestrate the multi-scale curve mixture.
To ensure full reproducibility, we detail the exact optimization loop in Algorithm~\ref{alg:scaffolding}.

The algorithm highlights our seamless integration of automated pruning and residual-guided densification. At every inflation step ($t \in \{100, 200, \dots \}$), we prune visually negligible curves ($\sigma(\alpha) \le 0.01$). To maintain the scheduled population growth, we compensate for these pruned curves by spawning an equivalent number of new curves, alongside the exponential inflation governed by $r_{\mathrm{num}}$. Newly spawned curves are strictly restricted by the monotonically decaying radius bound ($R_{k} = R_{k-1} \times r_{\mathrm{rad}}$) and are initialized on regions with high reconstruction error ($\mathcal{E}$) to encourage spatial separation between generations.

\begin{algorithm}[h]
\caption{Vector Scaffolding Optimization Loop}
\label{alg:scaffolding}
\begin{algorithmic}[1]
\Require Target Image $\mathcal{I}_{gt}$, Total inflation steps $K$, Step interval $\Delta T = 100$, Population ratio $r_{\mathrm{num}}$, Scale decay $r_{\mathrm{rad}} = 0.5$.
\State Initialize base curves $\mathcal{B}$ with base radius $R_{base}$
\State $R_{curr} \gets R_{base}$
\For{$t = 1$ \textbf{to} MaxIterations}
    \State Render $\mathcal{I}_{pred} \gets g(\mathcal{B})$
    \State Compute Loss $\mathcal{L}$ and backpropagate

    \If{$t \in \{ \Delta T, 2\Delta T, \dots, K\Delta T \}$}
        \State \Comment{\textit{1. Opacity-based Pruning}}
        \State $\mathcal{M}_{prune} \gets \{ \mathcal{B}_i \mid \sigma(\alpha_i) \le 0.01 \}$
        \State Remove curves in $\mathcal{M}_{prune}$ from $\mathcal{B}$
        \State $N_{removed} \gets |\mathcal{M}_{prune}|$

        \State \Comment{\textit{2. Progressive Stratification}}
        \State $R_{curr} \gets R_{curr} \times r_{\mathrm{rad}}$

        \State \Comment{\textit{3. Residual Error Calculation}}
        \State $\mathcal{E} \gets \|\mathcal{I}_{gt} - \mathcal{I}_{pred}\|_2^2 \ast \text{GaussianBlur}$

        \State \Comment{\textit{4. Rapid Inflation Scheduling}}
        \State $N_{target} \gets \text{len}(\mathcal{B}) \times r_{\mathrm{num}}$
        \State $N_{spawn} \gets N_{target} + N_{removed}$

        \State Sample $N_{spawn}$ coordinates from $\mathcal{E}$
        \State Spawn new curves at sampled coords with $R_{curr}$
        \State Append new curves to $\mathcal{B}$ with Z-ordering
    \EndIf
    \State Optimizer Step and Zero Grad
\EndFor
\end{algorithmic}
\end{algorithm}

\section{More Ablation Studies}
\label{appx:ablation}

\begin{figure}[h]
\centering
\setlength{\tabcolsep}{1pt}
\renewcommand{\arraystretch}{0.4}
\begin{tabular}{ccc}
    \includegraphics[width=0.32\linewidth]{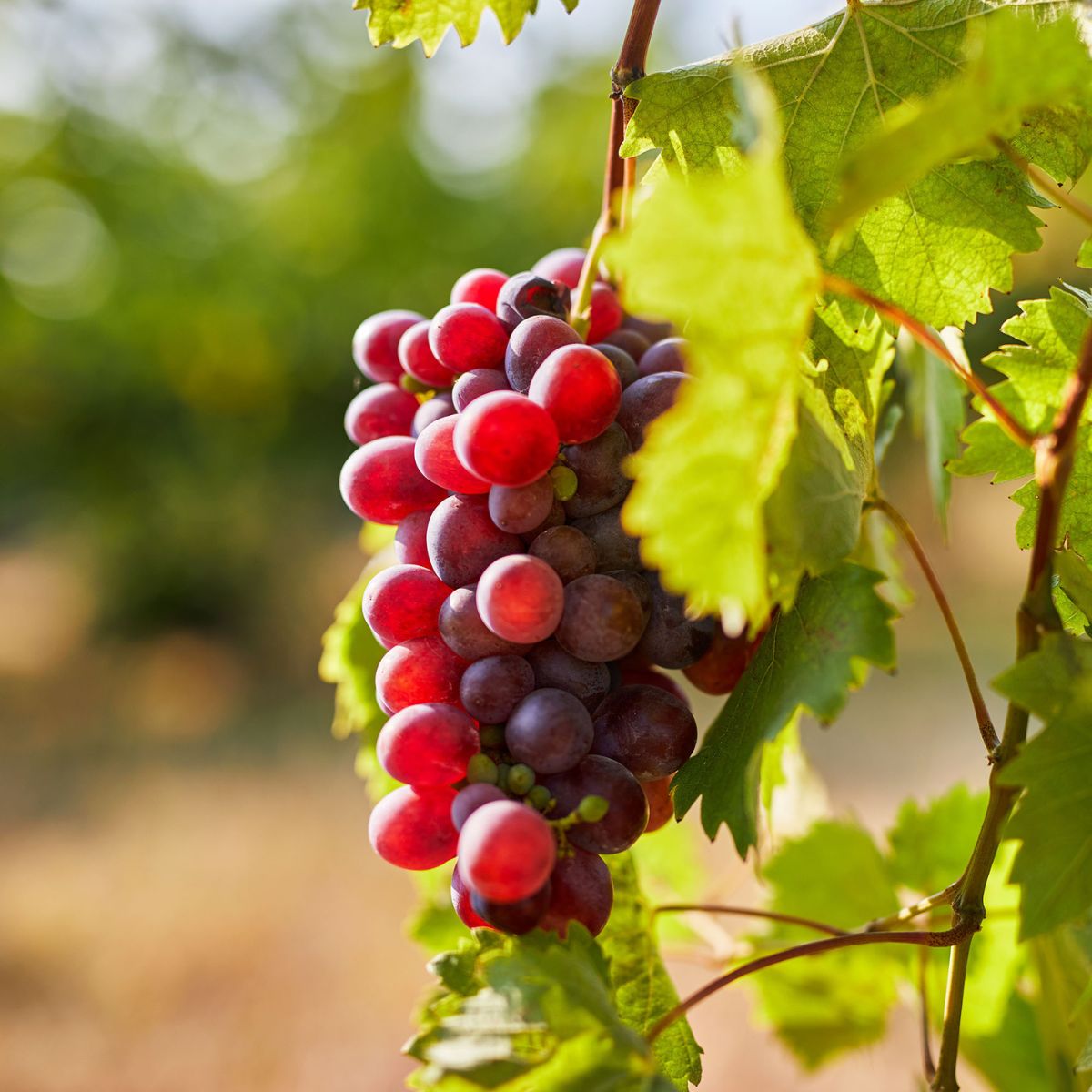} &
    \includegraphics[width=0.32\linewidth]{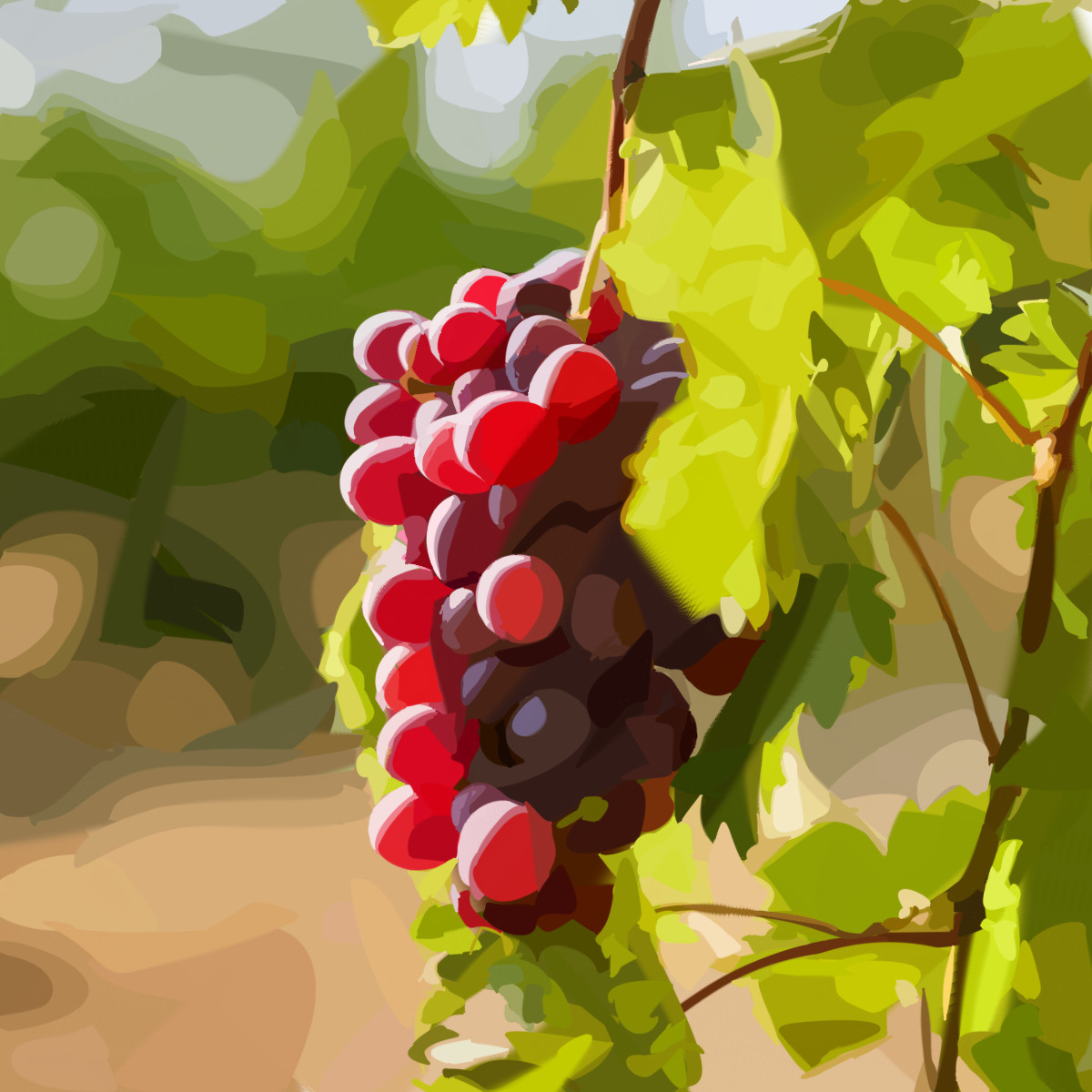} &
    \includegraphics[width=0.32\linewidth]{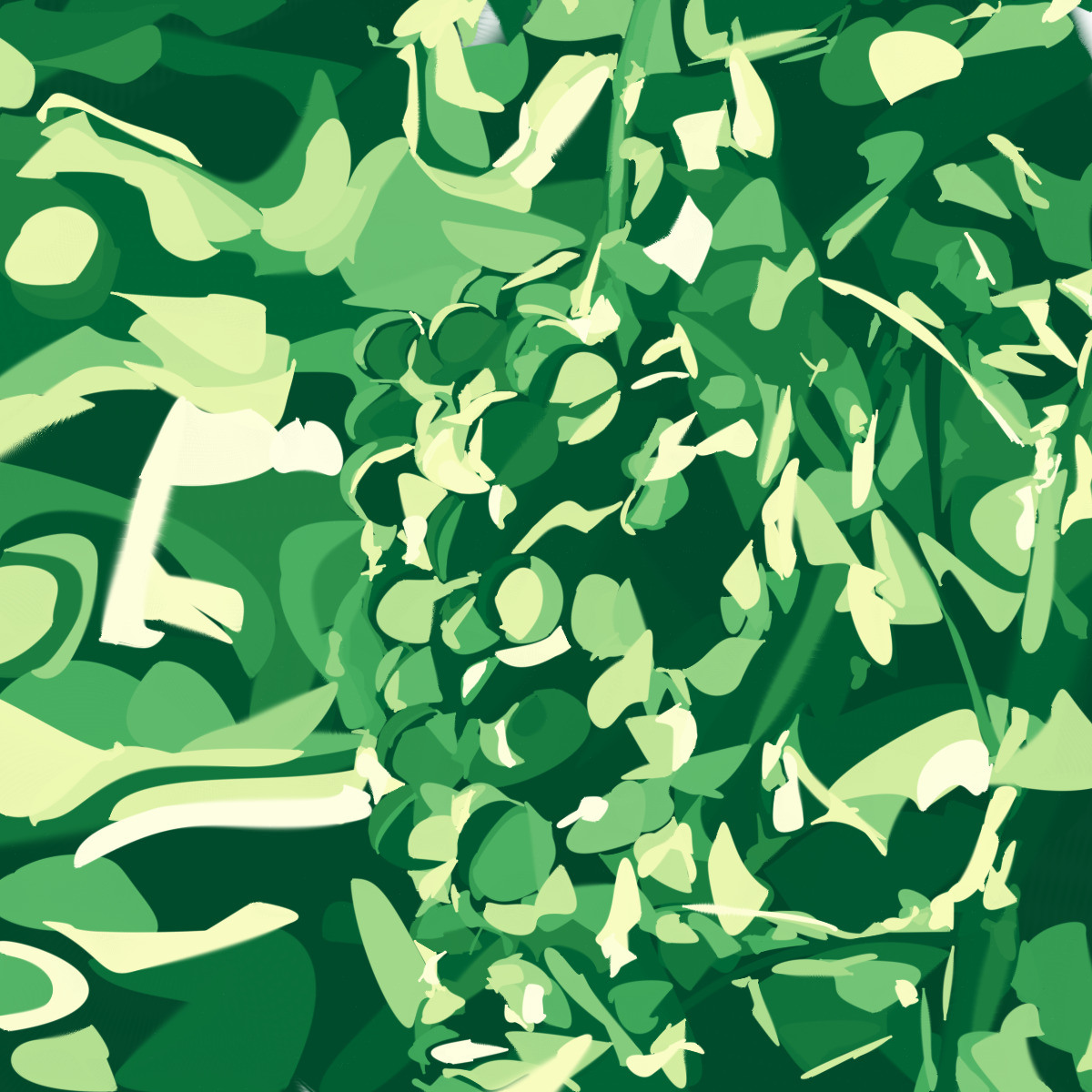} \\
    \small Original & \small Vectorized & \small Layer ordering \\
\end{tabular}
\caption{\textbf{Layered Primitive Visualization.}
The deterministic temporal $z$-ordering induced by \emph{Progressive Stratification} naturally aligns the optimization-induced layer index with the underlying scale hierarchy, so newer fine-scale curves sit on top of coarser base curves without dynamic re-sorting.}
\label{fig:primitive_visualization}
\end{figure}

\subsubsection{Layered Primitive Structure.}
Figure~\ref{fig:primitive_visualization} visualizes the curves of a vectorized image, colored by their generation index.
Because the temporal $z$-order is fixed by construction in \emph{Progressive Stratification}, the resulting primitive stack carries an interpretable LoD-aligned structure rather than the chaotic bounding-box-sorted ordering used by previous splatting-based methods~\cite{liu2025bezier}.

\begin{figure}[t]
\centering
\setlength{\tabcolsep}{1pt}
\renewcommand{\arraystretch}{0.4}
\begin{tabular}{ccc}
    \includegraphics[width=0.32\linewidth]{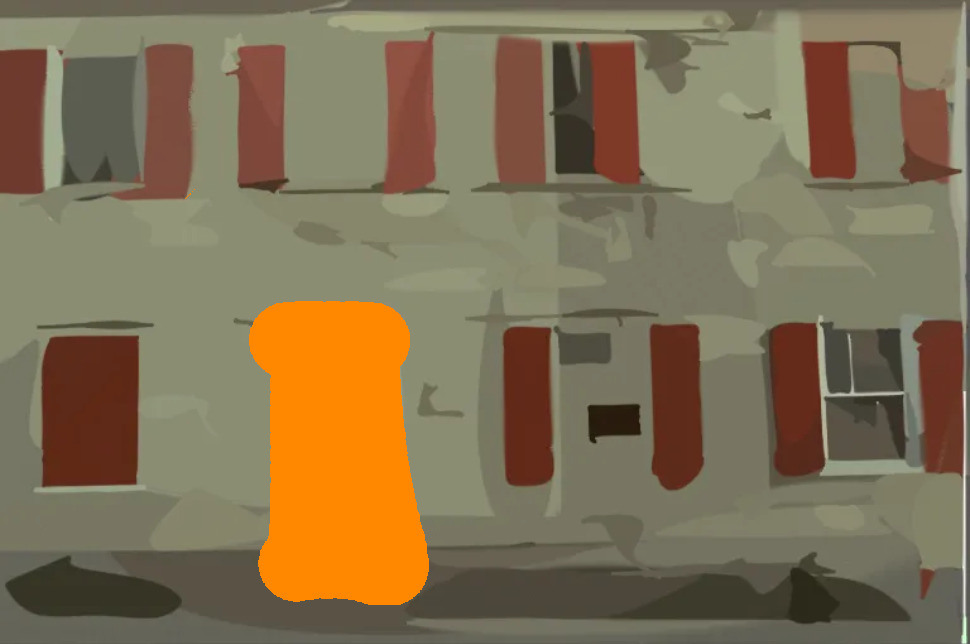} &
    \includegraphics[width=0.32\linewidth]{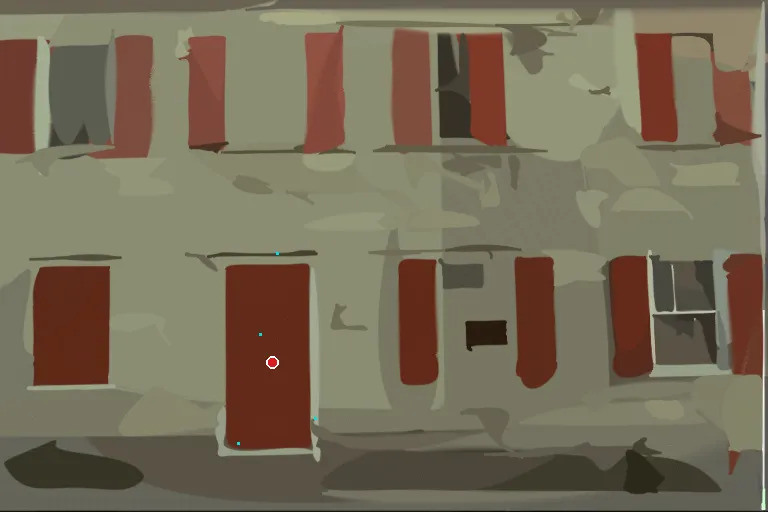} &
    \includegraphics[width=0.32\linewidth]{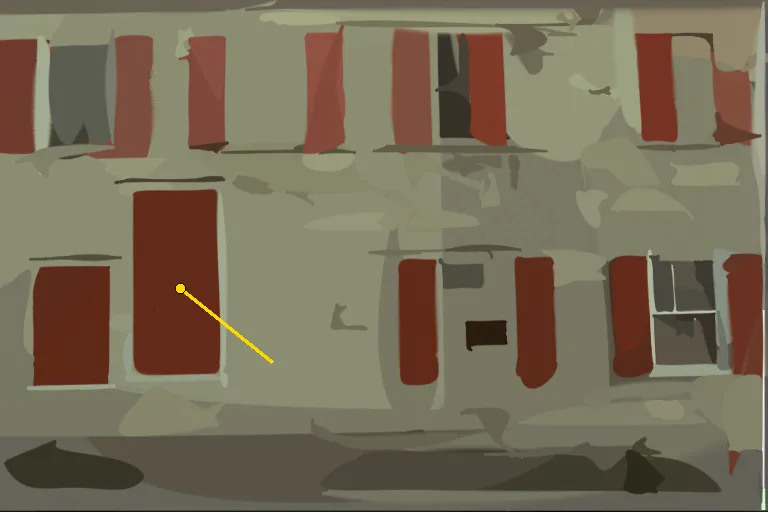} \\
    \small Brush tool & \small Selection & \small After edit \\
\end{tabular}
\caption{\textbf{Editability Demonstration.}
Output of our framework imported into a vector-editing demo built upon our pipeline.
The hierarchical scaffold yields path primitives organized by level-of-detail, enabling straightforward selection and local edits at the vector level.
We claim improved \emph{local} editability rather than a full semantic-editability solution.}
\label{fig:editability}
\end{figure}

\subsubsection{Editability.}
To complement the quantitative metrics, we provide a qualitative demonstration of vector-level editability in Figure~\ref{fig:editability}.
Because our framework produces a LoD-aligned hierarchy of closed B\'ezier curves (Fig.~\ref{fig:primitive_visualization}), users can directly select and modify primitives at a chosen scale.
Edits remain localized to the intended region, supporting our claim that hierarchical scaffolding yields improved local editability over flat polygon-soup outputs.

\begin{table}[t]
\centering
\setlength{\tabcolsep}{3pt}
\renewcommand{\arraystretch}{1.0}
\caption{\textbf{Hyperparameter sensitivity analysis on a 4-image Kodak subset (mean PSNR, dB).} \textbf{Bold*}: paper default; $\Delta$: max$-$min range. Performance is stable across a wide range of values around the defaults.}
\label{tab:hyperparam}
\begin{tabular}{@{}lrrrrrr@{}}
\toprule
\multicolumn{7}{@{}l}{\textit{(a) LR multiplier} $\lambda$} \\
\cmidrule(l){1-7}
value      & $\times 1$    & $\times 5$  & $\times 10$   & $\boldsymbol{\times 50^*}$  & $\times 100$   &  $\Delta$ \\
@\,1\,k    & 23.48 & 25.79 & 26.16 & \textbf{26.95} & 26.91 &  3.47 \\
@\,5\,k    & 25.83 & 27.07 & 27.38 & \textbf{27.95} & 27.84 &  2.12 \\
\midrule
\multicolumn{7}{@{}l}{\textit{(b) Radius decay} $r_{\mathrm{rad}}$} \\
\cmidrule(l){1-7}
value      &   0.1   & 0.25  & $\boldsymbol{0.5^*}$  & 0.75  & 1.0\,(off) &   $\Delta$ \\
@\,1\,k    & 26.40 & 26.68 & \textbf{26.96} & 26.93 & 24.33      & 2.63 \\
@\,5\,k    & 27.80 & 27.92 & \textbf{27.94} & 27.85 & 25.30      & 2.64 \\
\midrule
\multicolumn{7}{@{}l}{\textit{(c) Inflation gap} $\Delta T$} \\
\cmidrule(l){1-7}
value      & 25    & 50    & $\boldsymbol{100^*}$  & 200   & 400 &  $\Delta$ \\
@\,1\,k    & 27.27 & 27.16 & \textbf{26.97} & 23.81 & 21.39 & 5.87 \\
@\,5\,k    & 27.85 & 27.93 & \textbf{27.93} & 27.97 & 27.95 & 0.19 \\
\bottomrule
\end{tabular}%
\end{table}

\subsubsection{Hyperparameter Sensitivity.}
We perform a one-at-a-time sweep around the paper defaults to verify that our optimization is robust to hyperparameter choices.
Table~\ref{tab:hyperparam} reports the PSNR sensitivity to (a) the global learning-rate multiplier $\lambda$, (b) the radius decay ratio $r_{\mathrm{rad}}$, and (c) the inflation gap $\Delta T$ on a 4-image Kodak subset.
Across all three knobs, the PSNR at $5\,\mathrm{k}$ iterations varies by at most $\sim\!2.6\,\mathrm{dB}$, with the paper default consistently lying at or near the optimum.
The largest improvement margin (LR $\Delta\!=\!3.47\,\mathrm{dB}$ at $1\,\mathrm{k}$ iterations) confirms that the $\times 50$ learning rate is the most impactful design decision, while performance remains stable across a wide range of $r_{\mathrm{rad}}$ and $\Delta T$.

\section{Visualization of Optimization Videos}
\label{sec:video_guide}
To better demonstrate the topological advantages of our proposed method, we capture intermediate renders during the training process.
We compare the learning dynamics between the previous state-of-the-art, B\'ezier Splatting~\cite{liu2025bezier}, and our Vector Scaffolding side by side across various challenging scenes, with synchronized iteration counts for direct comparison.
Note that our method has a slightly more efficient ($\sim25\%$) iteration loop than the baseline; therefore, running our full algorithm for $5\,\mathrm{k}$ iterations is more than $2.5\times$ faster than running the baseline algorithm for $10\,\mathrm{k}$ iterations.
This speedup is visualized in Figure~\ref{fig:figure_one_benchmark} in the main text.

\begin{figure}[t]
\centering
\setlength{\tabcolsep}{1pt}
\renewcommand{\arraystretch}{0.4}
\begin{tabular}{ccccc}
    \includegraphics[width=0.19\linewidth]{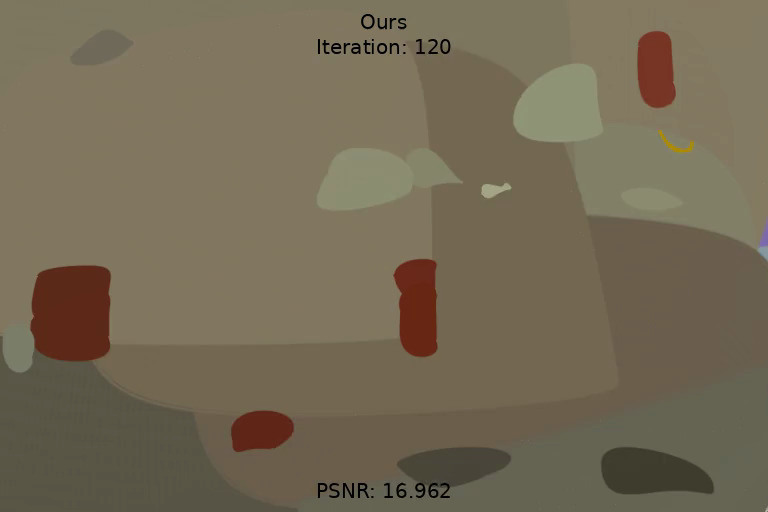} &
    \includegraphics[width=0.19\linewidth]{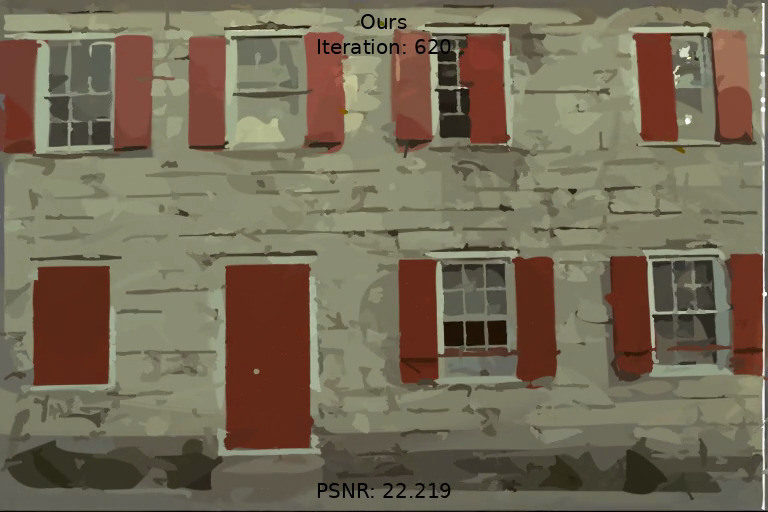} &
    \includegraphics[width=0.19\linewidth]{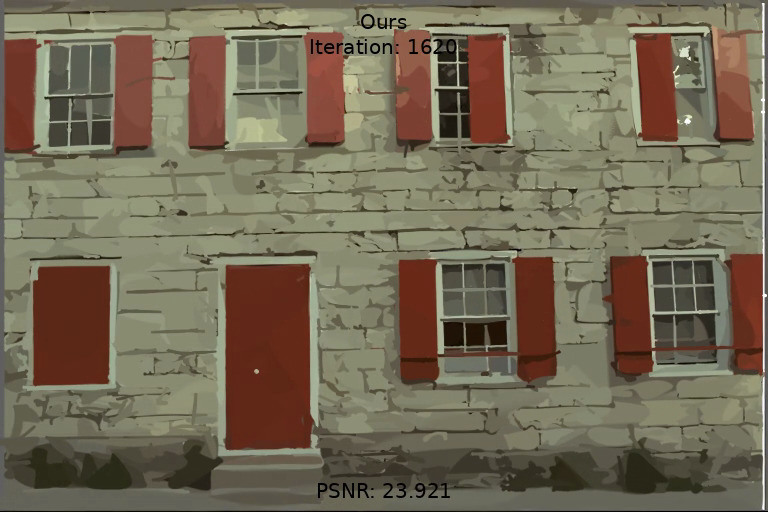} &
    \includegraphics[width=0.19\linewidth]{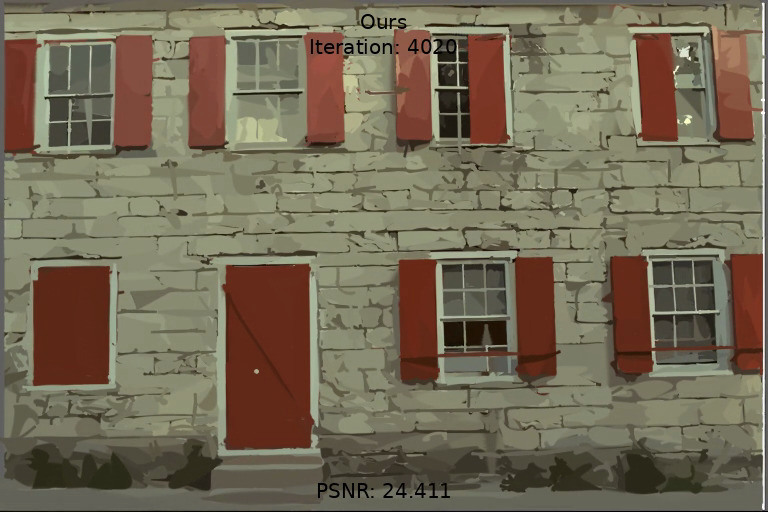} &
    \includegraphics[width=0.19\linewidth]{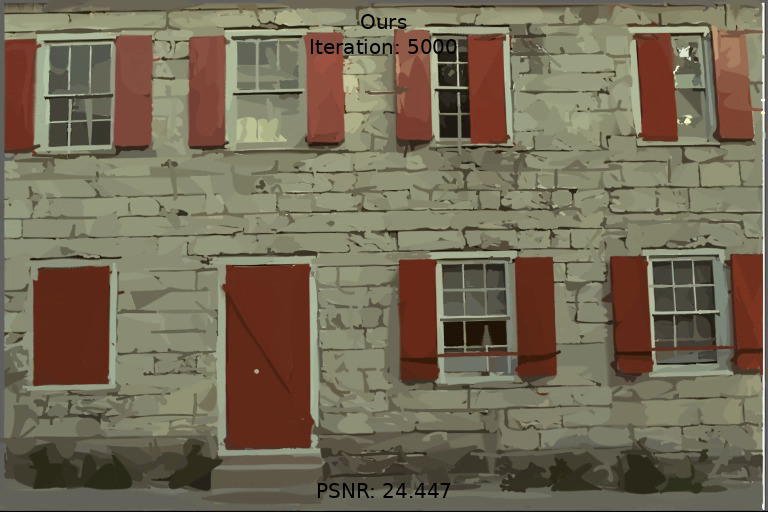} \\
    \includegraphics[width=0.19\linewidth]{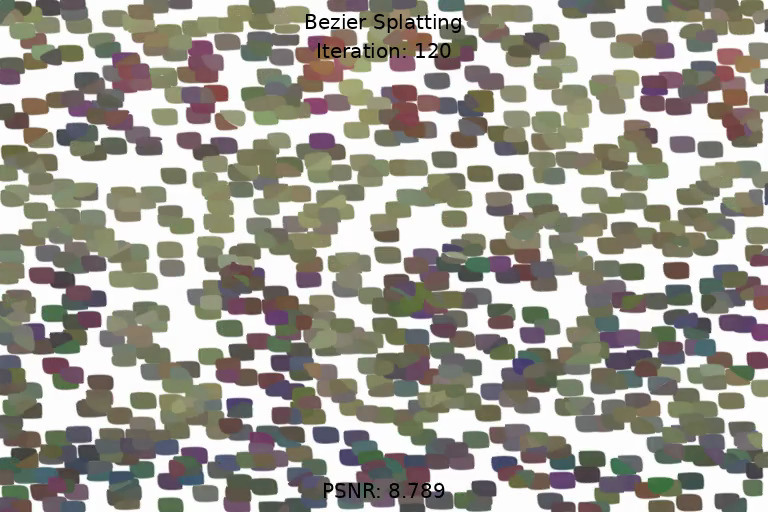} &
    \includegraphics[width=0.19\linewidth]{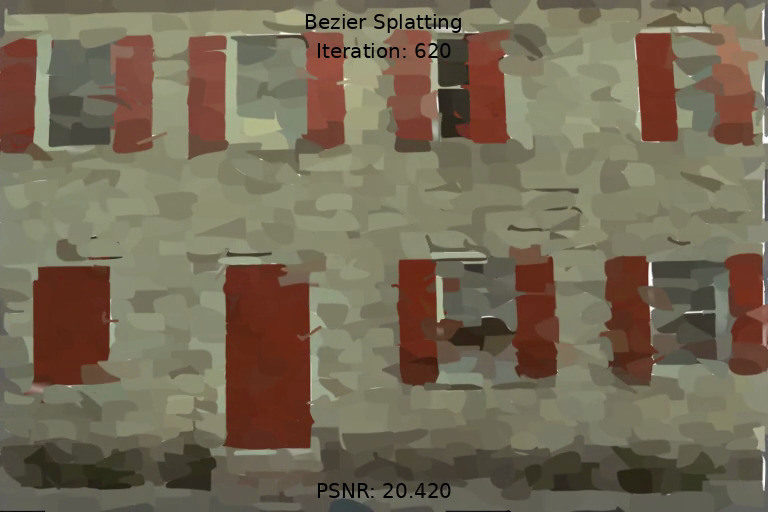} &
    \includegraphics[width=0.19\linewidth]{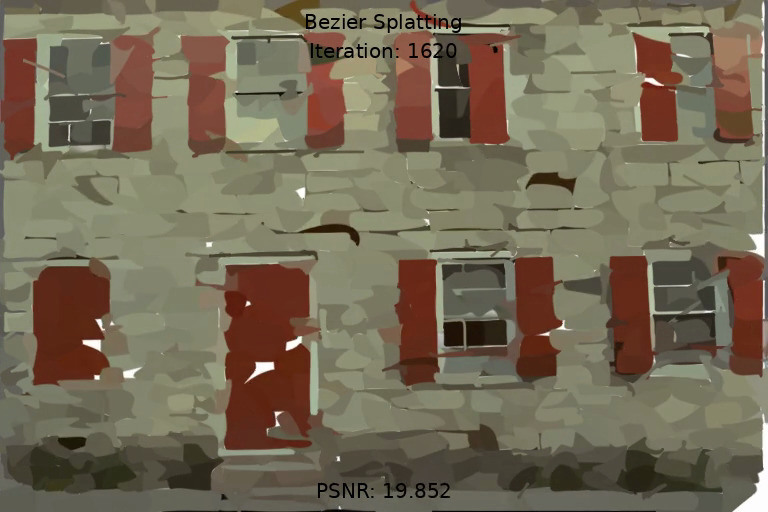} &
    \includegraphics[width=0.19\linewidth]{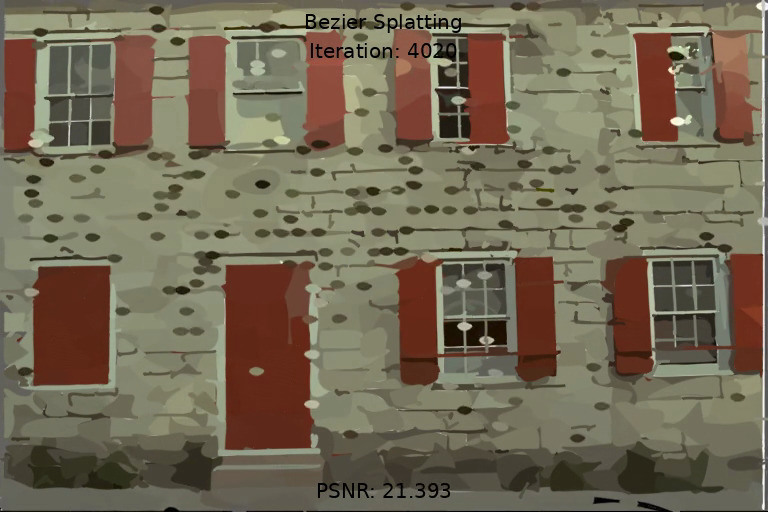} &
    \includegraphics[width=0.19\linewidth]{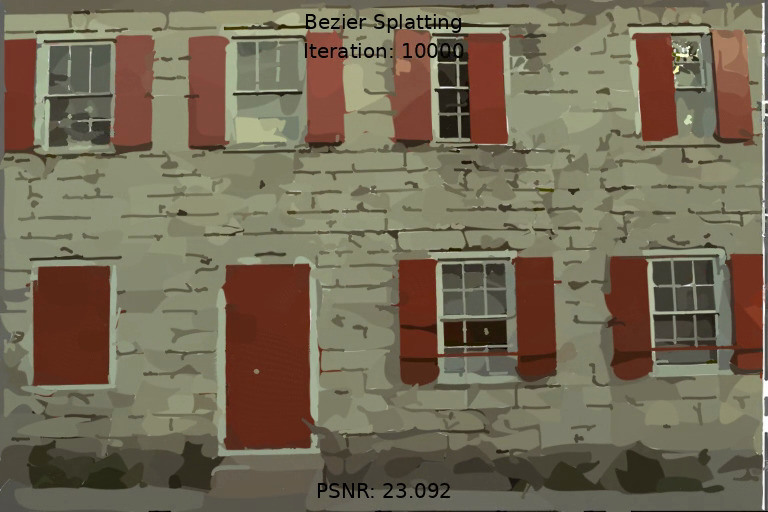} \\
\end{tabular}
\caption{\textbf{Optimization trajectory on Kodak \texttt{kodim01}.}
Top: ours; bottom: B\'ezier Splatting. Columns are matched iterations
($\sim$100, 600, 1600, 4000, 9980). Our method anchors smooth roof/wall regions early, whereas the baseline scatters narrow strokes that never coalesce.}
\label{fig:video_kodim01}
\end{figure}

\begin{figure}[t]
\centering
\setlength{\tabcolsep}{1pt}
\renewcommand{\arraystretch}{0.4}
\begin{tabular}{ccccc}
    \includegraphics[width=0.19\linewidth]{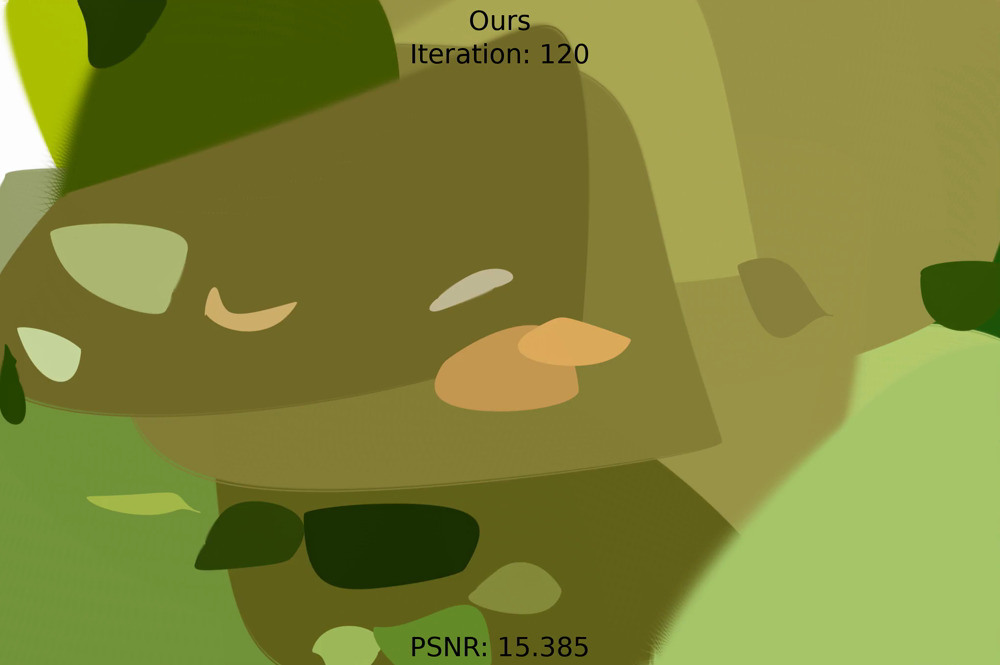} &
    \includegraphics[width=0.19\linewidth]{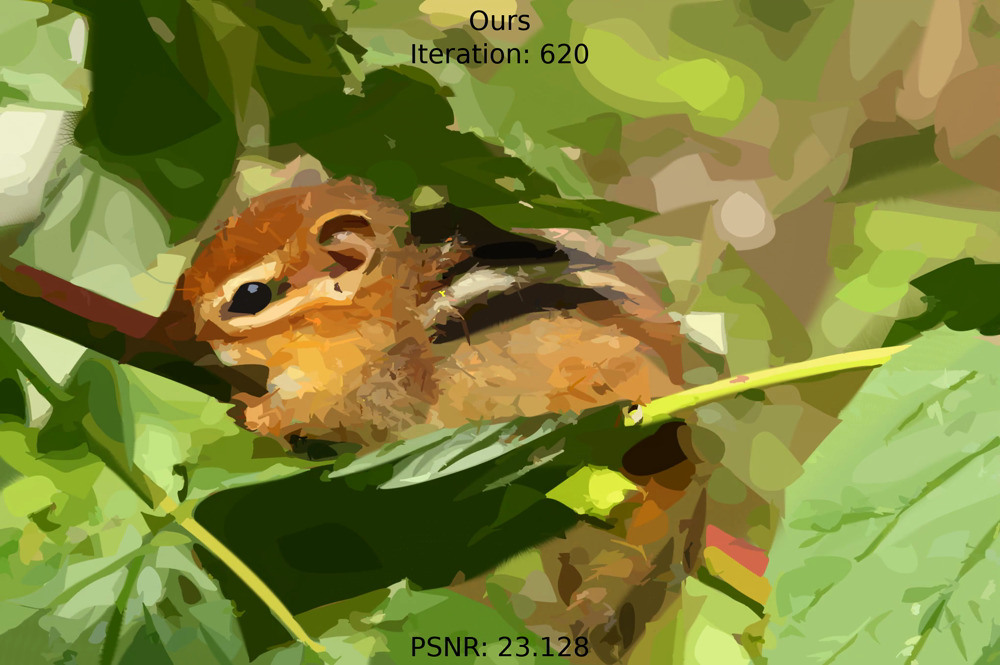} &
    \includegraphics[width=0.19\linewidth]{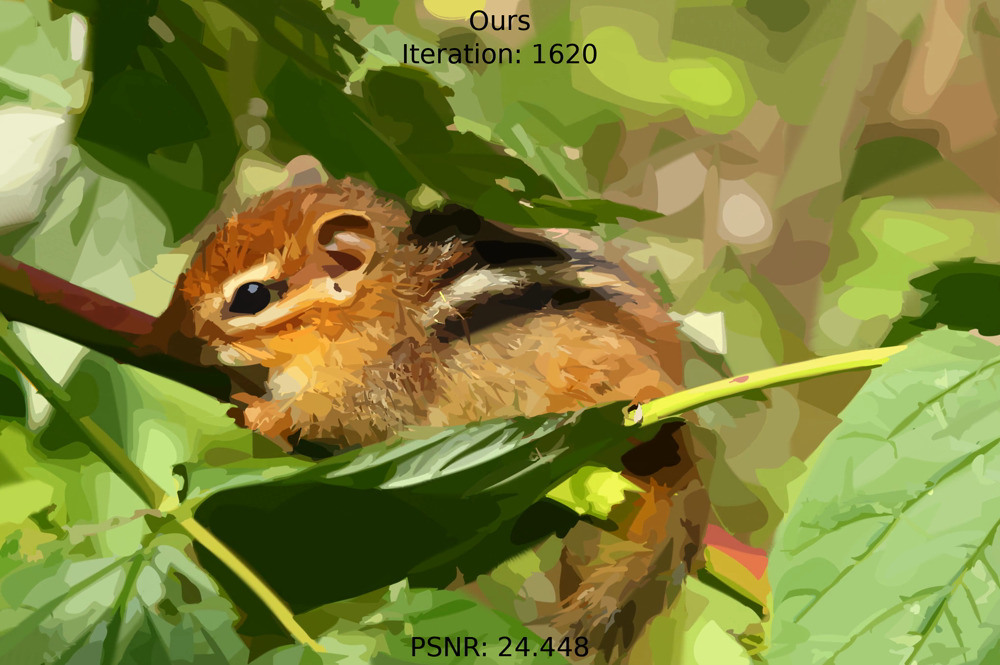} &
    \includegraphics[width=0.19\linewidth]{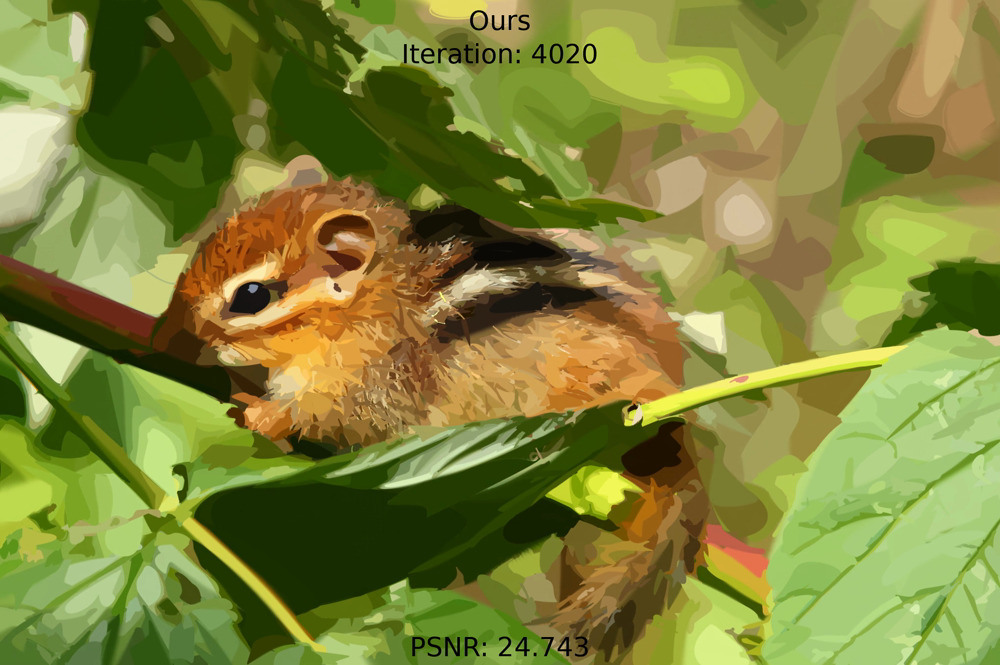} &
    \includegraphics[width=0.19\linewidth]{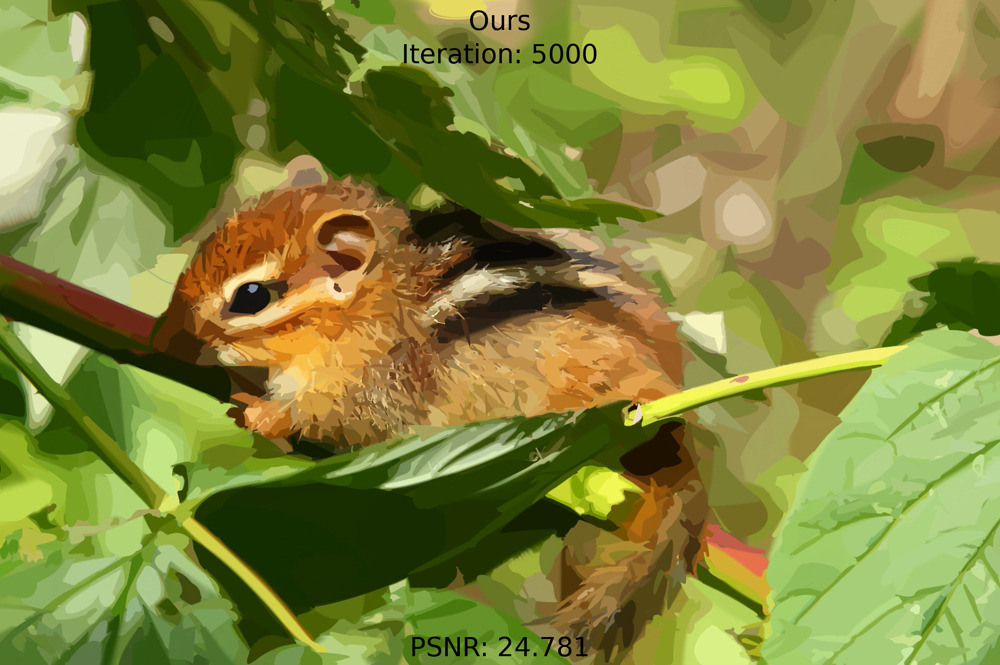} \\
    \includegraphics[width=0.19\linewidth]{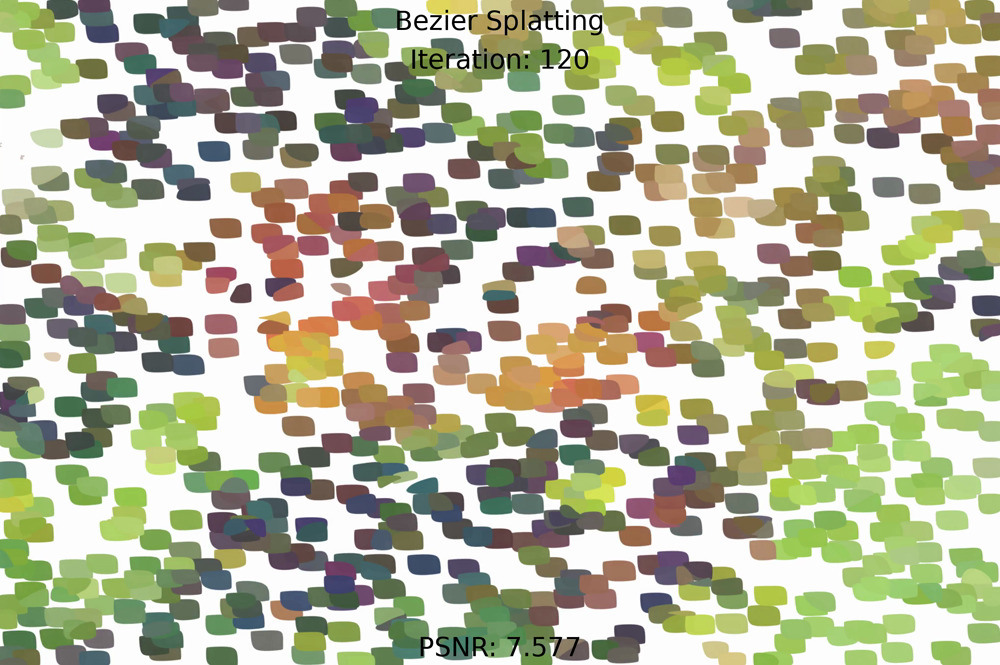} &
    \includegraphics[width=0.19\linewidth]{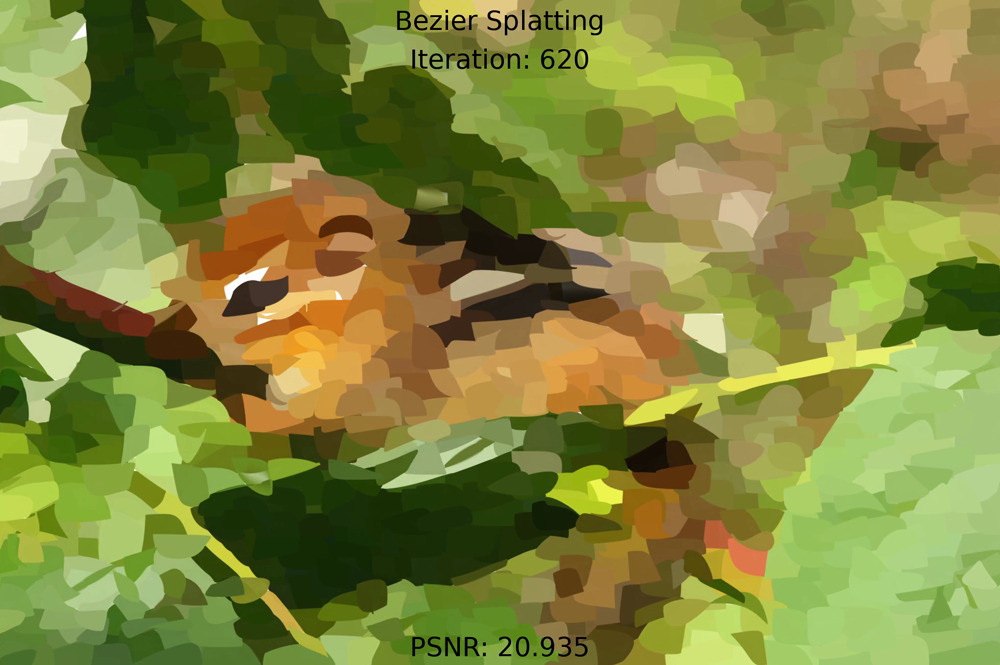} &
    \includegraphics[width=0.19\linewidth]{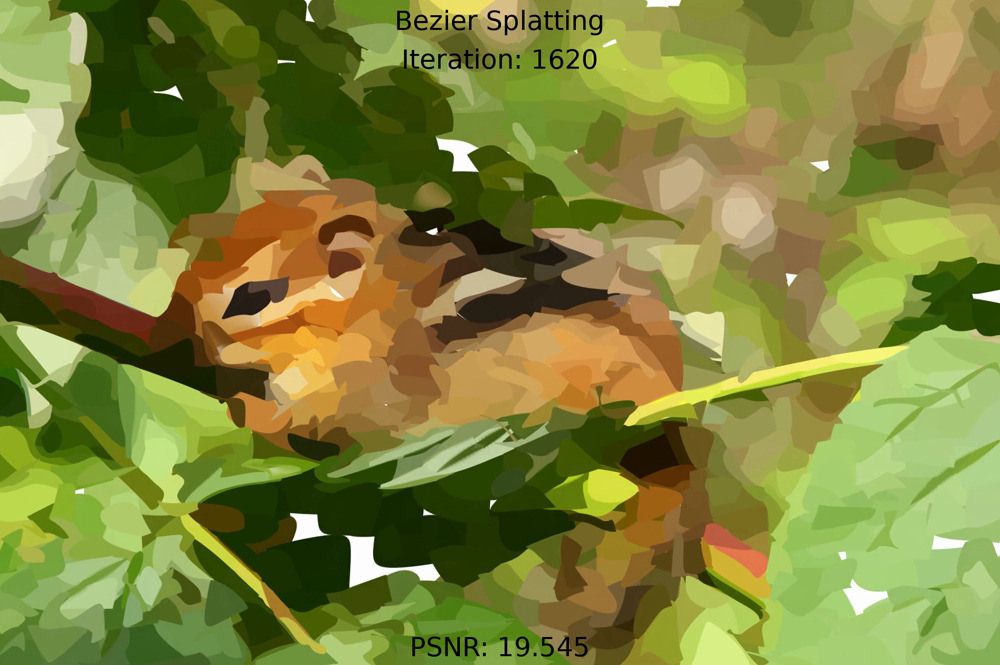} &
    \includegraphics[width=0.19\linewidth]{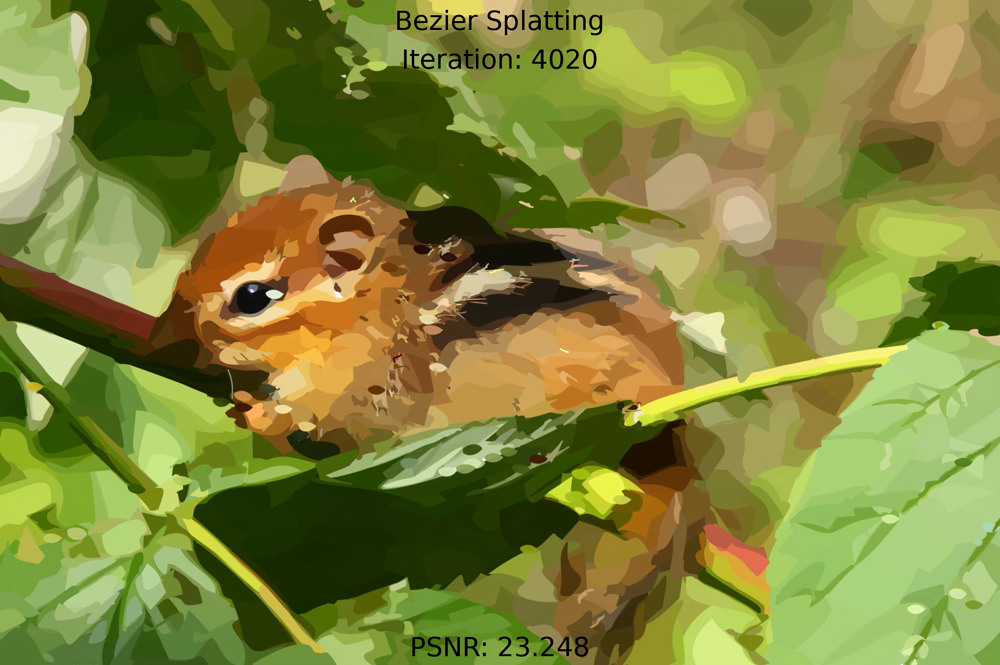} &
    \includegraphics[width=0.19\linewidth]{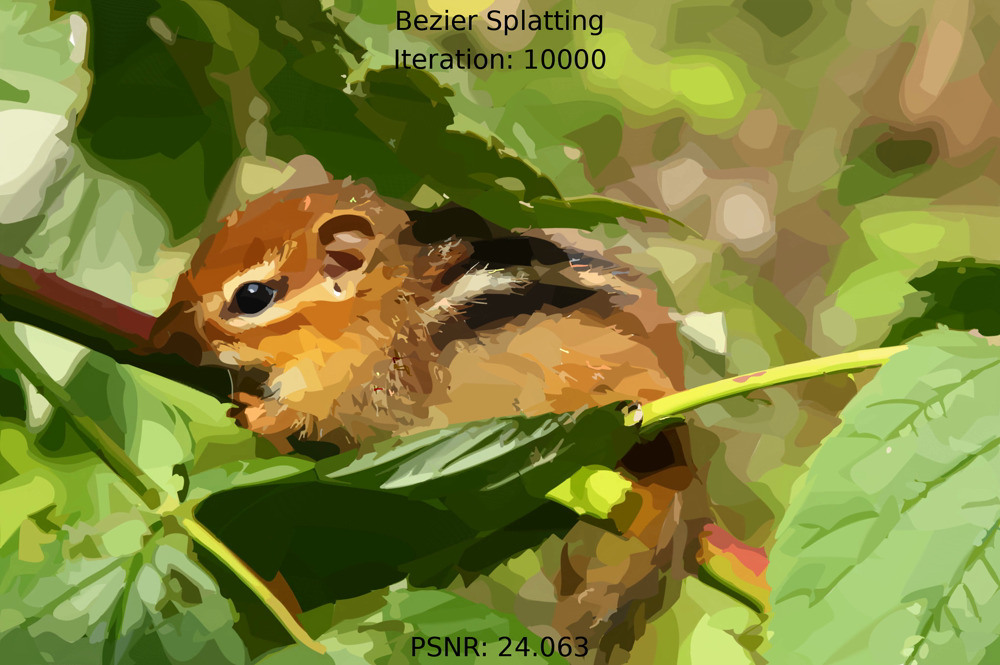} \\
\end{tabular}
\caption{\textbf{Optimization trajectory on DIV2K \texttt{0294}.}
Top: ours; bottom: B\'ezier Splatting. Background foliage and fur texture form coherently in our run, while the baseline keeps redistributing strokes near the subject without locking the surrounding context.}
\label{fig:video_div2k_0294}
\end{figure}

To this end, Figures~\ref{fig:video_kodim01}--\ref{fig:video_div2k_0294} present intermediate frames extracted from four representative videos at five synchronized iteration checkpoints ($\sim$100, 600, 1600, 4000, and $\geq$5000 iterations).
The PSNR and iteration count are overlaid on each frame for reference.
Across all examples, our method (top row) establishes coherent macroscopic structure within the first few hundred iterations, while B\'ezier Splatting (bottom row) exhibits a noisy ``polygon-soup'' distribution that slowly resolves through aggressive local edits.
This visually supports our claim: the speedup originates from a better optimization trajectory enabled by progressive stratification, not merely from per-step efficiency.

\begin{figure}[t]
\centering
\setlength{\tabcolsep}{1pt}
\renewcommand{\arraystretch}{0.4}
\begin{tabular}{ccccc}
    \includegraphics[width=0.19\linewidth]{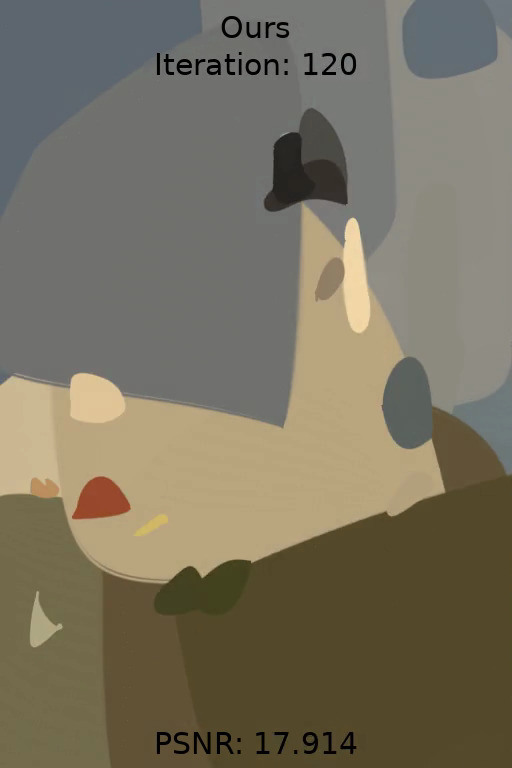} &
    \includegraphics[width=0.19\linewidth]{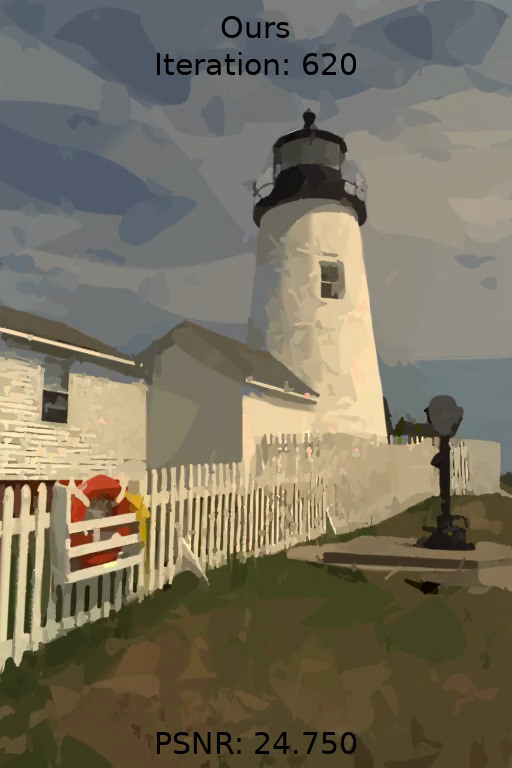} &
    \includegraphics[width=0.19\linewidth]{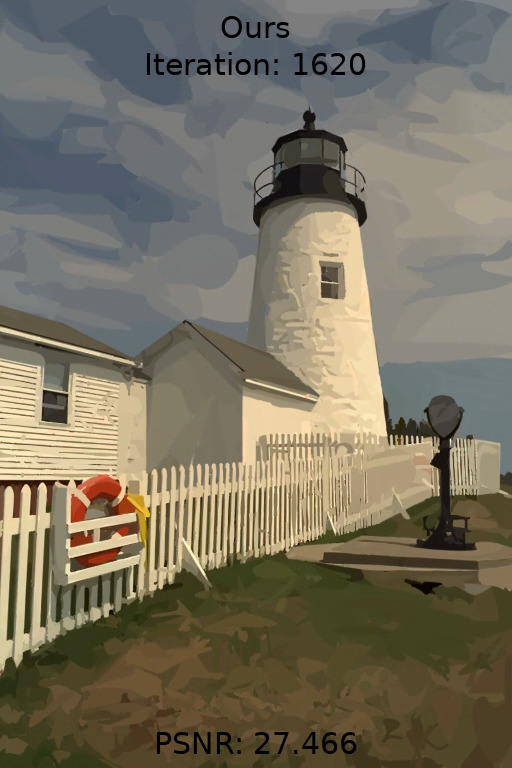} &
    \includegraphics[width=0.19\linewidth]{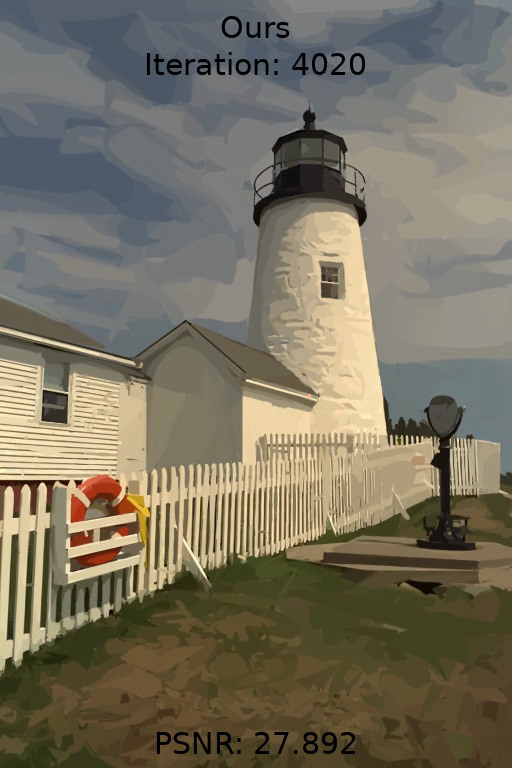} &
    \includegraphics[width=0.19\linewidth]{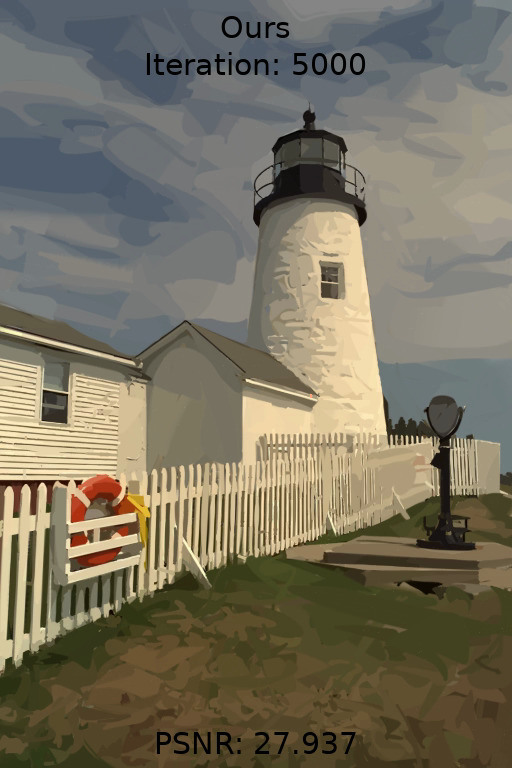} \\
    \includegraphics[width=0.19\linewidth]{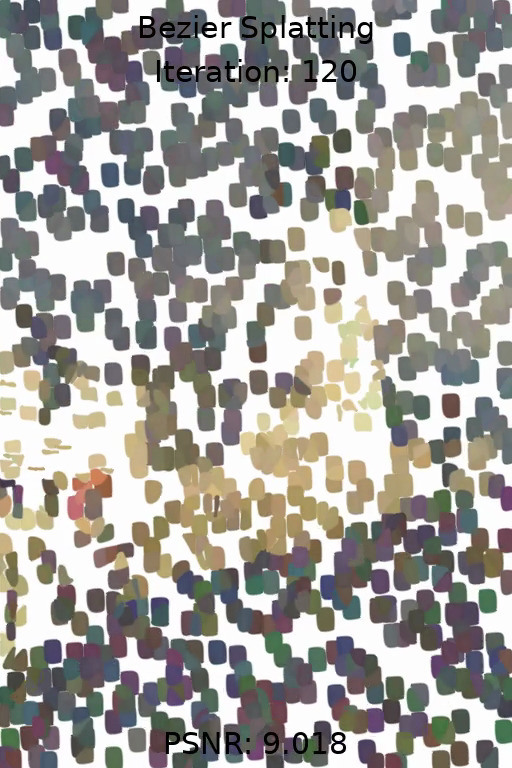} &
    \includegraphics[width=0.19\linewidth]{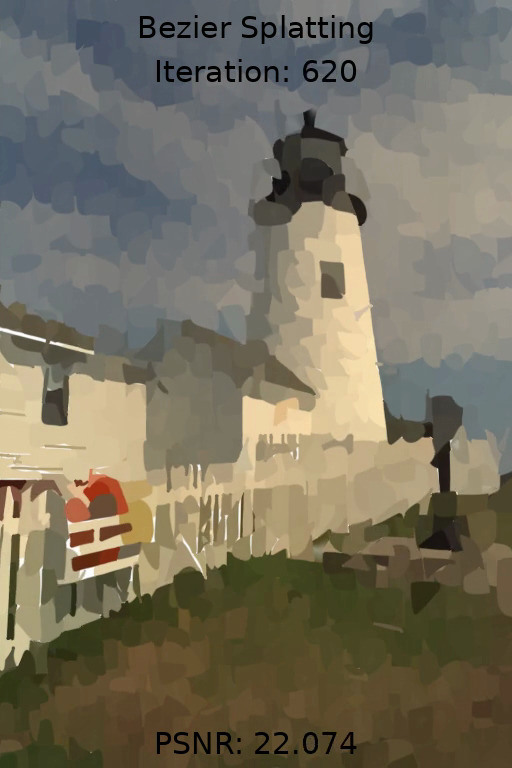} &
    \includegraphics[width=0.19\linewidth]{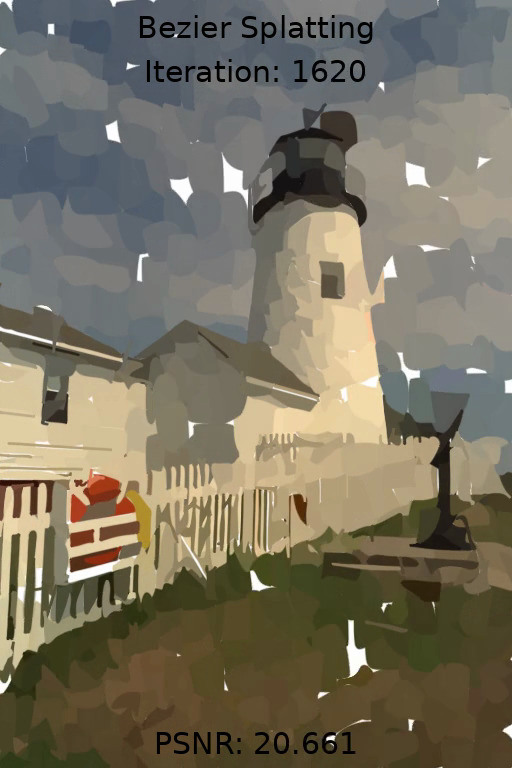} &
    \includegraphics[width=0.19\linewidth]{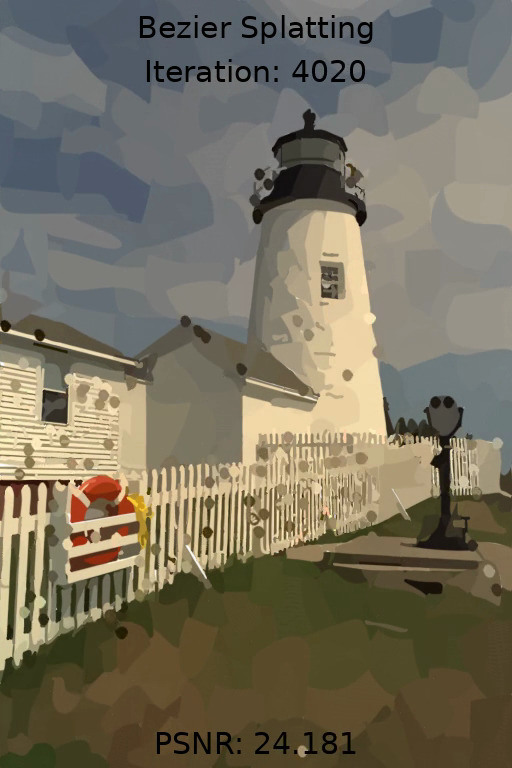} &
    \includegraphics[width=0.19\linewidth]{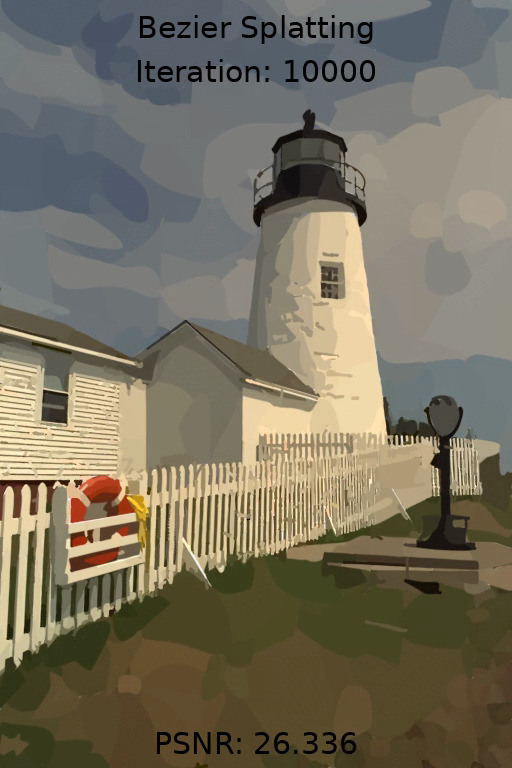} \\
\end{tabular}
\caption{\textbf{Optimization trajectory on Kodak \texttt{kodim19} (portrait).}
Top: ours; bottom: B\'ezier Splatting at matched iterations. Our hierarchical refinement quickly converges to clean silhouettes, while the baseline keeps scattered fragments around the boundaries throughout training.}
\label{fig:video_kodim19}
\end{figure}

\begin{figure}[t]
\centering
\setlength{\tabcolsep}{1pt}
\renewcommand{\arraystretch}{0.4}
\begin{tabular}{ccccc}
    \includegraphics[width=0.19\linewidth]{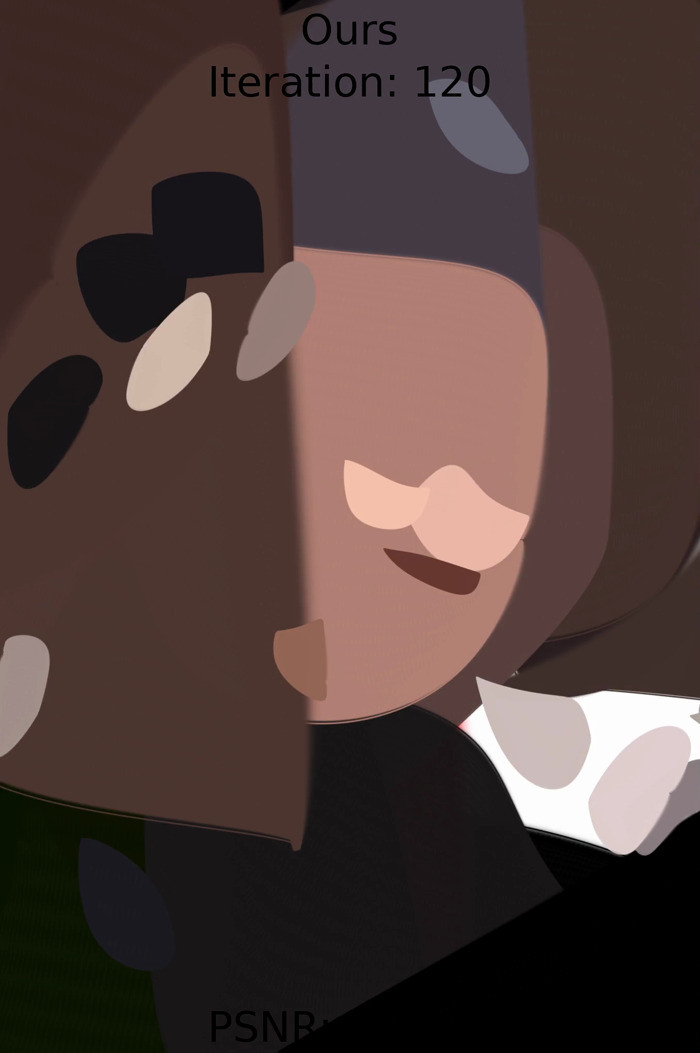} &
    \includegraphics[width=0.19\linewidth]{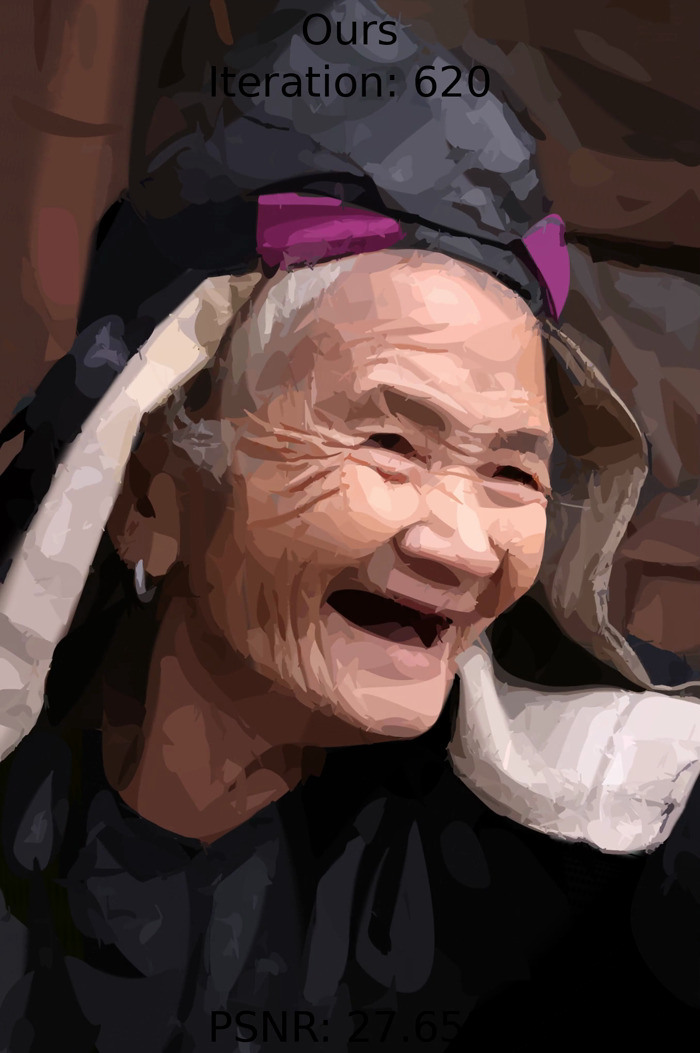} &
    \includegraphics[width=0.19\linewidth]{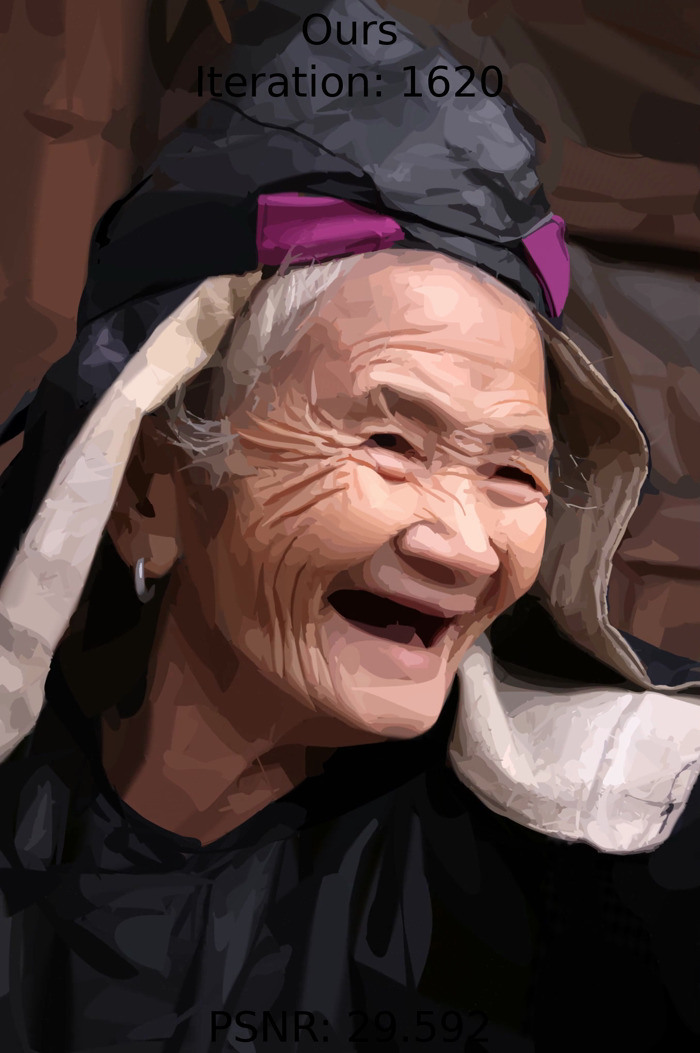} &
    \includegraphics[width=0.19\linewidth]{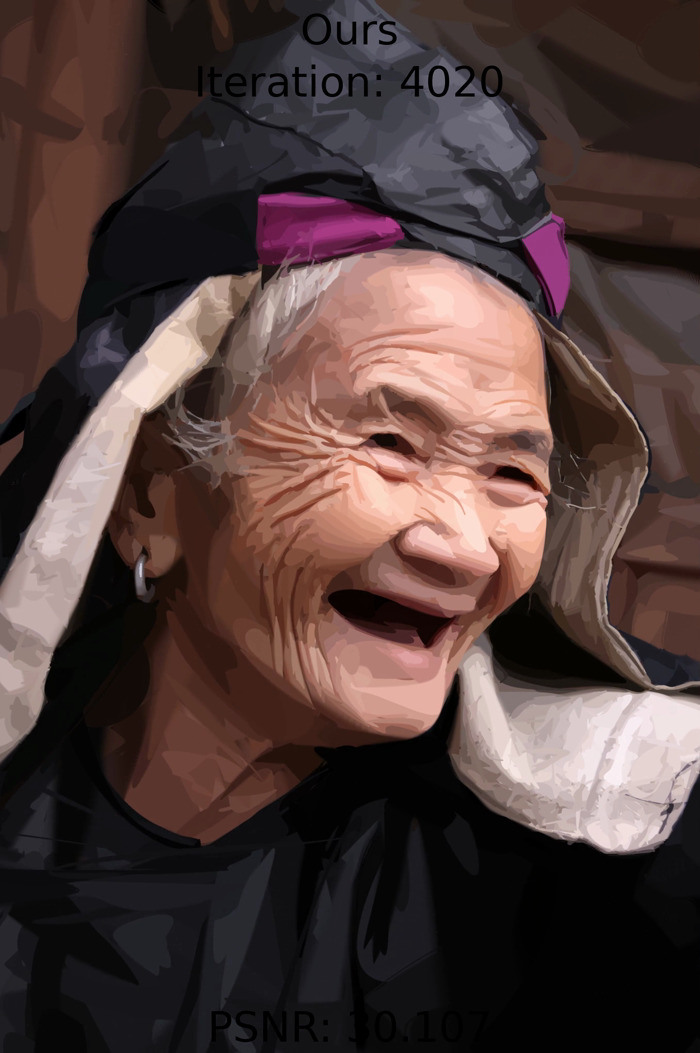} &
    \includegraphics[width=0.19\linewidth]{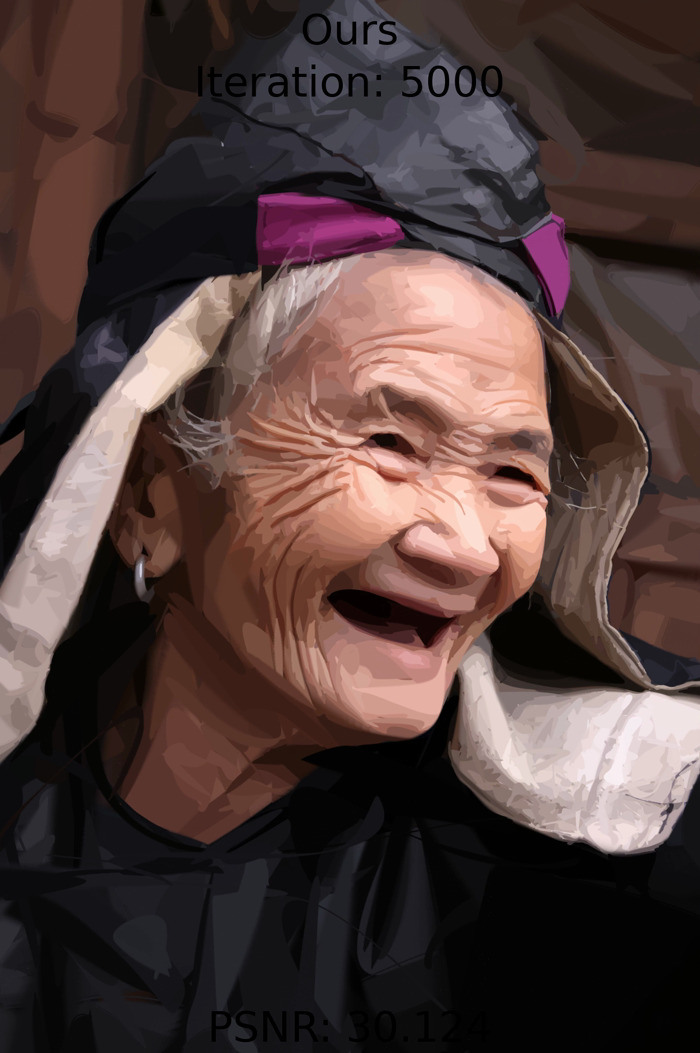} \\
    \includegraphics[width=0.19\linewidth]{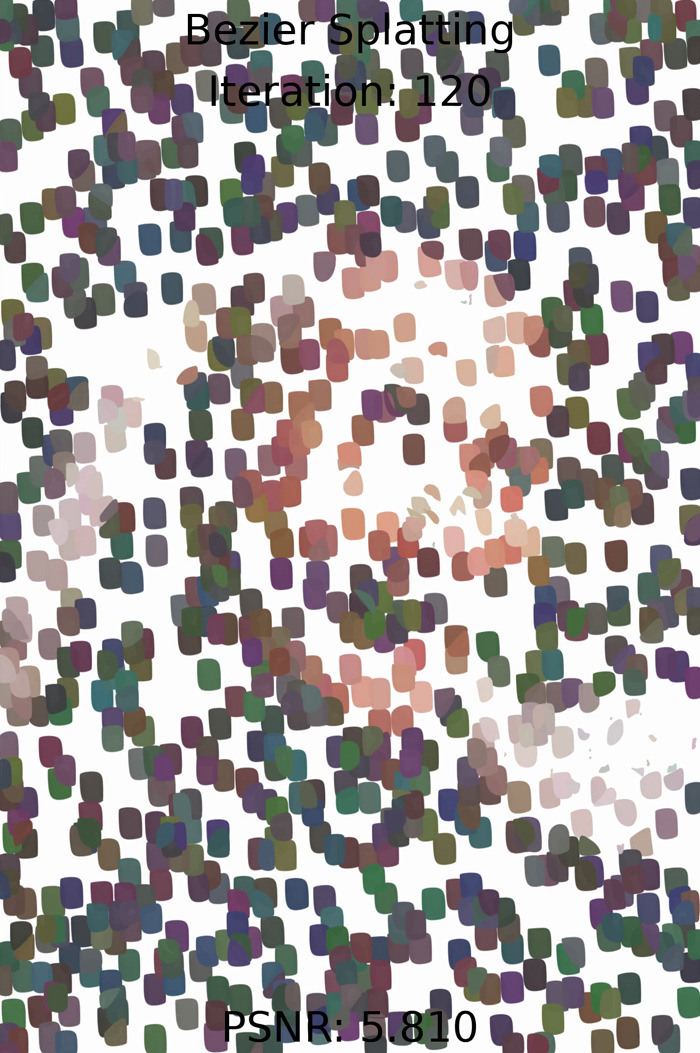} &
    \includegraphics[width=0.19\linewidth]{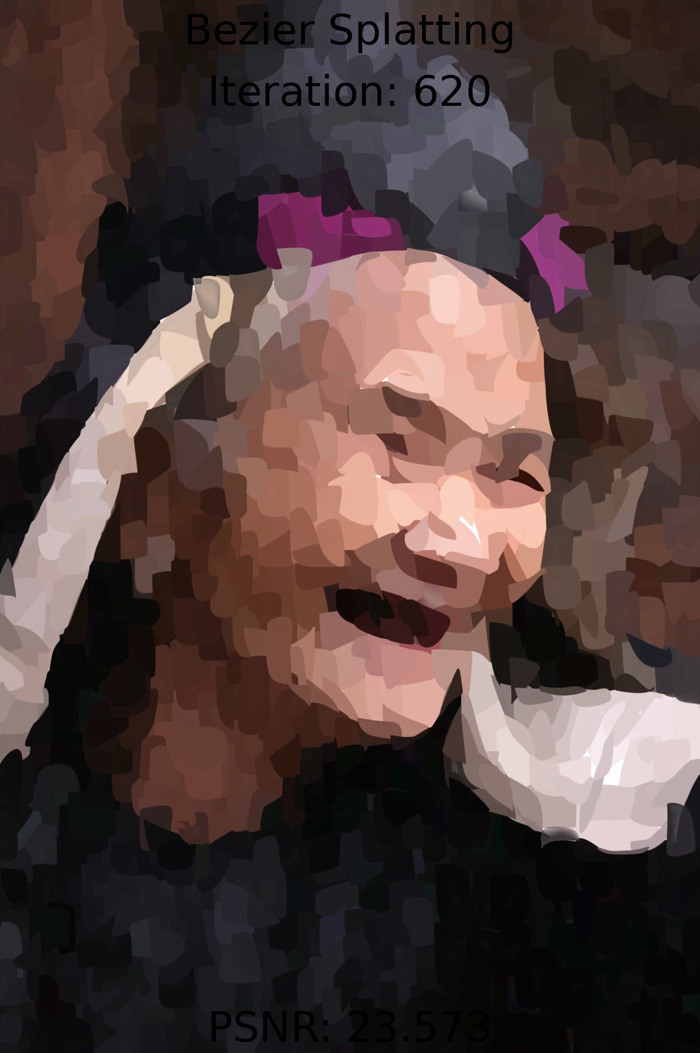} &
    \includegraphics[width=0.19\linewidth]{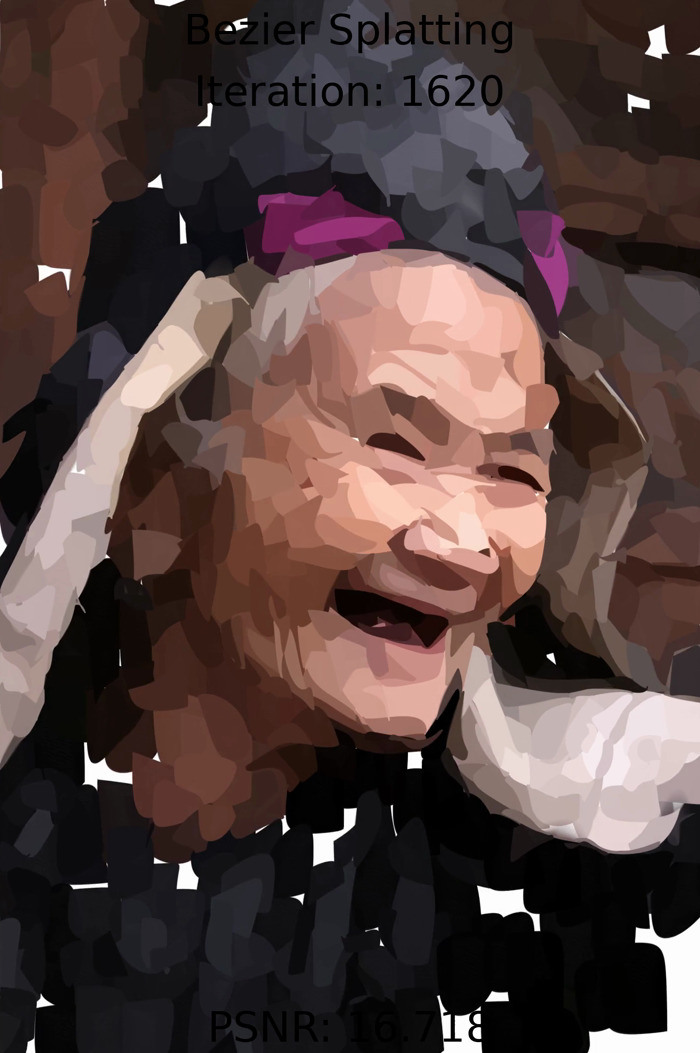} &
    \includegraphics[width=0.19\linewidth]{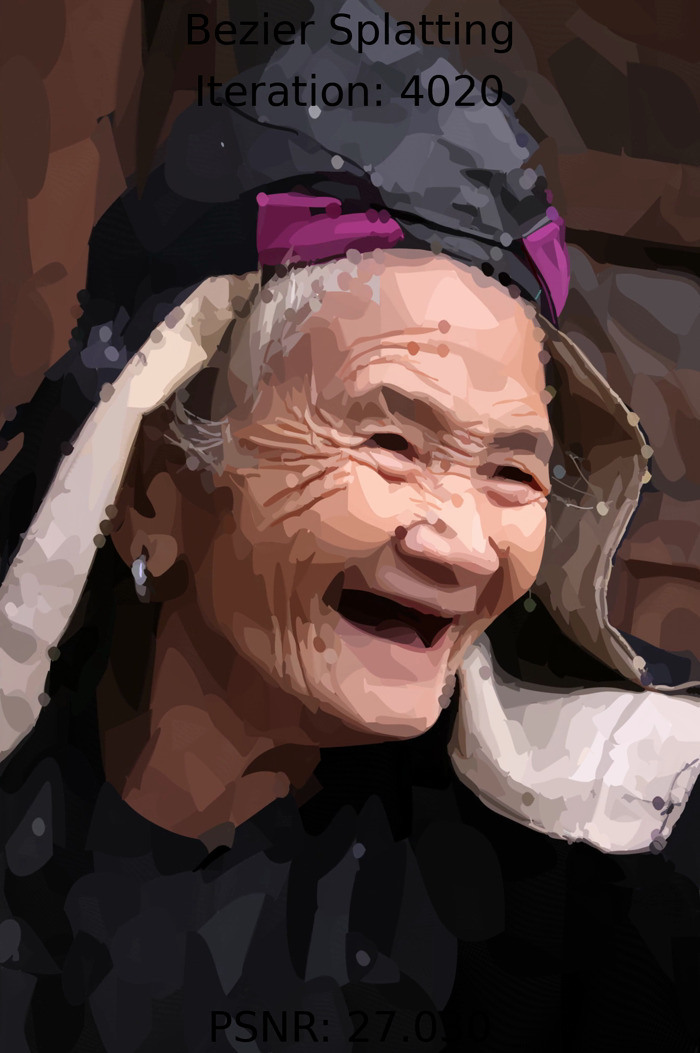} &
    \includegraphics[width=0.19\linewidth]{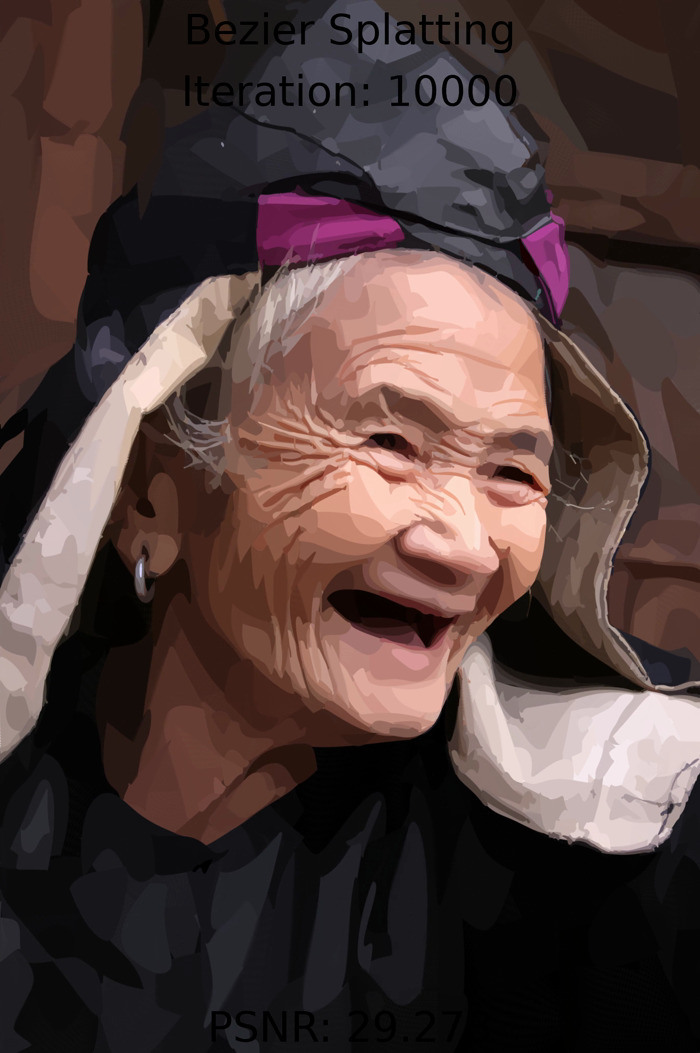} \\
\end{tabular}
\caption{\textbf{Optimization trajectory on DIV2K \texttt{0112}.}
Top: ours; bottom: B\'ezier Splatting. The portrait scene benefits the most from progressive stratification --- skin tones and fabric shading are recovered smoothly in our method, while the baseline distributes high-frequency noise across the face throughout training.}
\label{fig:video_div2k_0112}
\end{figure}

\end{document}